\documentclass{article}

\usepackage[utf8]{inputenc}
\usepackage{setspace}
\usepackage{graphicx}
\graphicspath{{./figures/}}
\usepackage{subcaption}
\usepackage{amsmath}
\usepackage{amssymb}
\usepackage{booktabs}
\usepackage{hyperref}
\usepackage{algorithm}
\usepackage{algpseudocode}
\usepackage[margin=1in]{geometry} 

\begin{document}

\title{Adaptive Mixing of Policies from Searching and\\ Policies from Learning}

\author{Gavin B.\ Rens\\
Computer Science Division, Stellenbosch University, Stellenbosch, South Africa\\
gavinrens@sun.ac.za}


\maketitle

\abstract{
{\bf Background:} 
    Distillation of training targets generated thru search/planning has proven useful in reinforcement learning, but search can take exceedingly long
    
    {\bf Objectives:}
    Rather than perform search to the same depth every time (typically at a fixed period of steps), reduce the search depth proportionally to the quality of the policy network priors.
    
    {\bf Methods:}
    We describe Flexer, an architecture that, for each step, mixes the policy from a neural network and the policy from Monte Carlo tree search. The mixing factor favors the MCTS policy as the policy imitation error of the network and the environment models' variance increases. 
    
    {\bf Results:}
    Flexer outperforms a version of AlphaZero (and DQN and ADP) for some experiments on three toy symbolic problems.
}
\section{Introduction}
Many reinforcement learning (RL) problems require some planning, or could at least benefit from planning. Planning requires environment models. If they are unknown, they must be learned or provided. In this work, we assume they will be learned from environment interactions. Planning is typically computationally and temporally expensive. It would thus be beneficial to store the results of planning to be resued later without having to replan for the same (similar) situations.
A common approach is to employ an artificial neural network (ANN) to learn and represent a policy (aka a policy network) and to employ Monte Carlo tree search (MCTS) to perform look-ahead planning \cite{atb17,alphazero,muzero20}.  
We propose an RL architecture that mixes the action recommendation of a policy network and MCTS planner at every step, weighting the MCTS recommendation more heavily inversely proportional to the quality of the policy network (PN) and to the variance of the environment models over time.
%
We call this architecture \textit{Flexer} for ease of reference.

When the PN quality is high, the advice of the planner will weigh less. It is a waste of resources to do extensive planning in such cases. One can also moderate the amount of planning with the quality of the models it depends on: if the transition and reward function have not yet been well-learned, then less planning is desirable, because the plans will be less accurate. Less planning allows the agent to spend more time gaining experience (to learn the environment models).

Flexer is similar to AlphaZero \cite{alphazero} and MuZero \cite{muzero20} in that there is interaction between a PN and a planner. There are two major differences though: 1) Flexer mixes policies and 2) Flexer maintains a global (state) value function which, in a sense, acts like an intermediary information structure between the PN and planner. The PN training data come from the value function, not directly from the MCTS tree root as is the case with AlphaZero and MuZero.
Moreover, in Flexer, the final value of each node (for a complete MCTS run) is used to improve (boost) the value function estimates.


We design an architecture, called AZLike, based on AlphaZero, to fit the context of a single agent in a stochastic environment where the agent has (almost) no access to the environment. The agent does not know the reward model nor the transition model a priori, but has to learn them to use them in planning. 
We then define Flexer, based on AZLike but with the policy trade-off (mixing) and budget control mechanics added.

Several variants of AZLike and Flexer are tested on three small problems. 
 Double Deep Q-learning Network (DQN) \cite{vgs16} and Adaptive Dynamic Programming (ADP) \cite{sb18} are also applied to the problems to get further perspectives on the AZLike/Flexer model-based approach and to better assess Flexer. The results are mixed but show that there is potential for the Flexer approach to reduce wasted computation on planning.

\begin{figure}[h]
  \centering
  \includegraphics[width=0.24\textwidth]{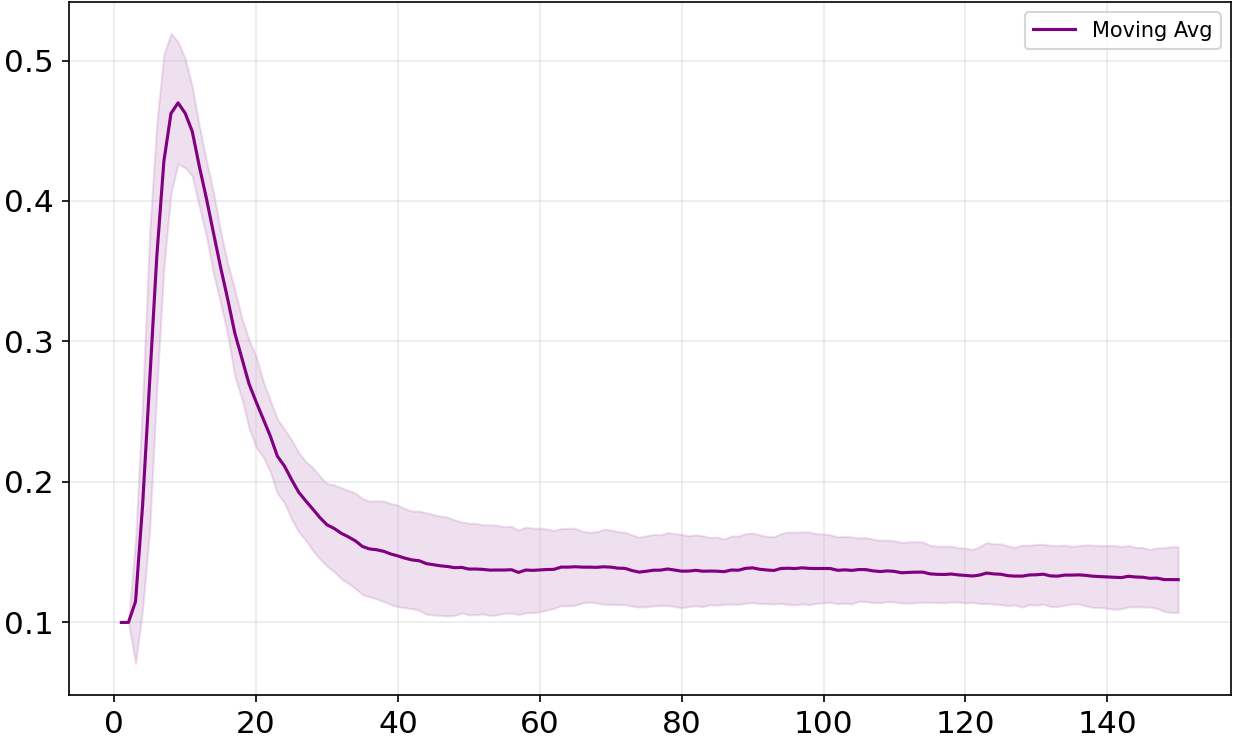}
  \hfill
  \includegraphics[width=0.24\textwidth]{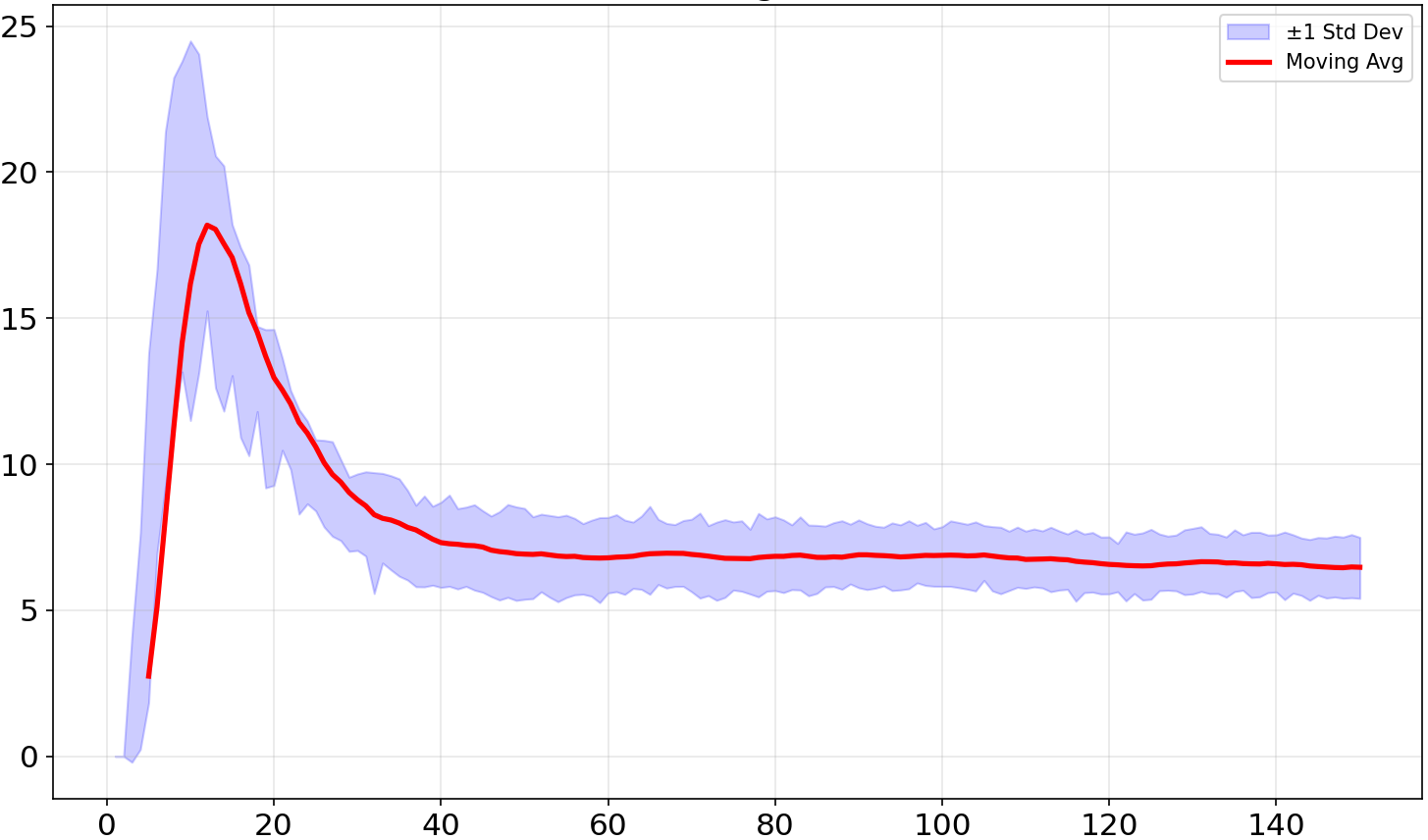}
  \hfill
  \includegraphics[width=0.24\textwidth]{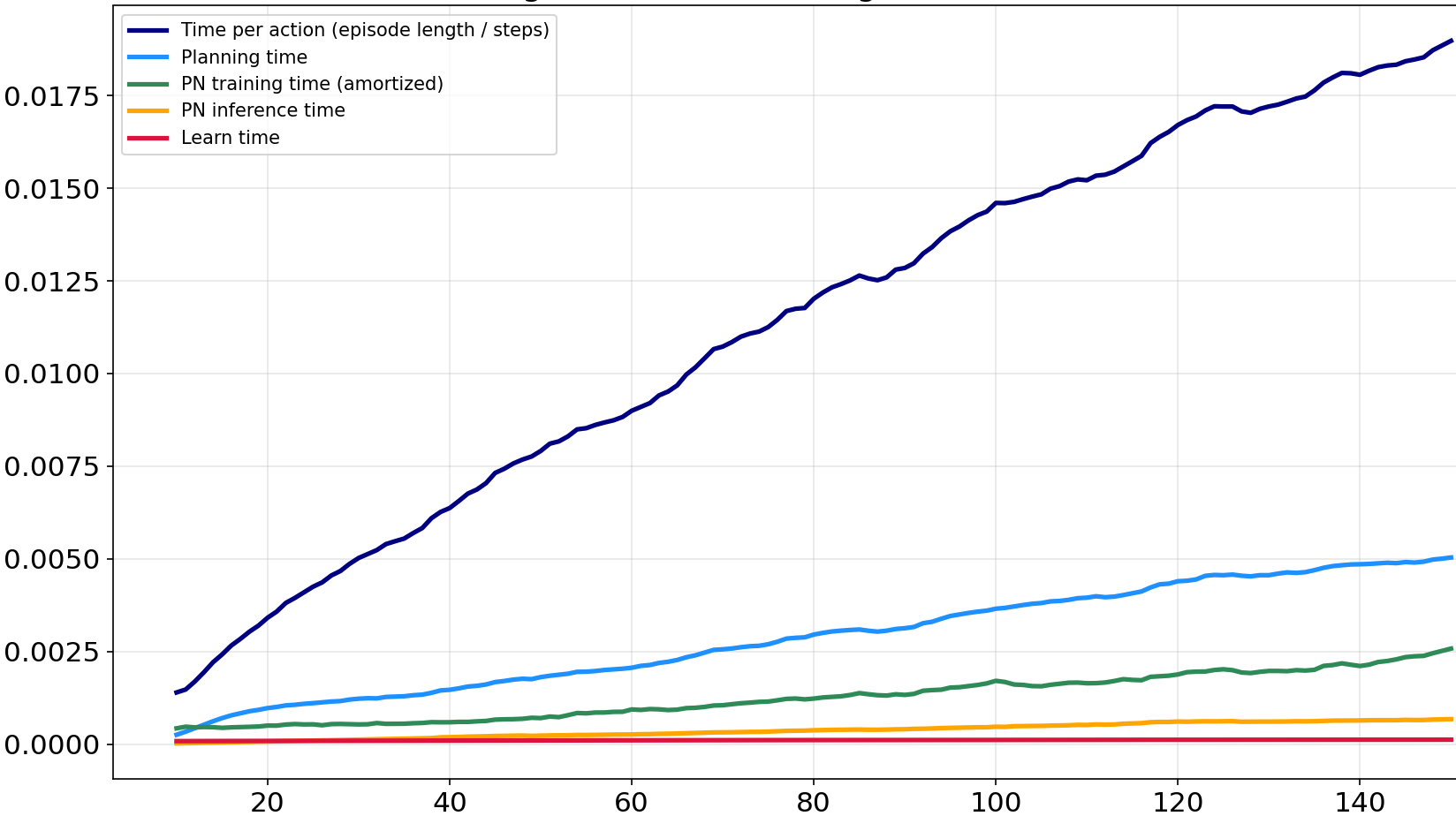}
  \hfill
  \includegraphics[width=0.24\textwidth]{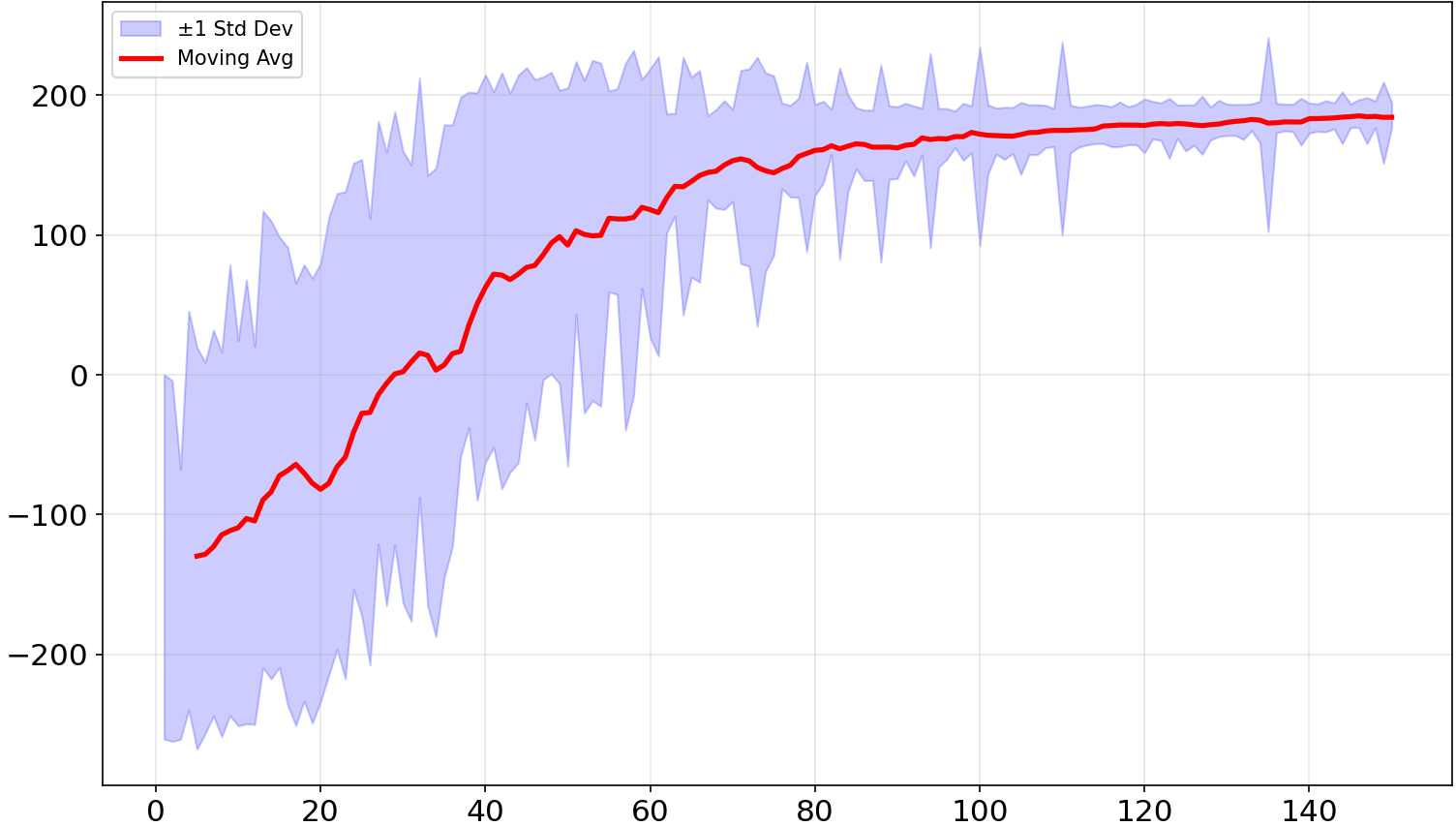}
  \caption{Change in mixing factor and its influence (left to right): mixing factor ($\mu$), number of nodes in MCTS tree. running times, return. Results are averages over 30 problem instances.}
  \label{fig:plot-of-mu-etc}
\end{figure}

Figure \ref{fig:plot-of-mu-etc} shows how the mixing factor changes as training progresses over 150 episodes (on the easy BlocksWorld problem; see \S~\ref{sec:environs}), and how it influences the amount of compute spent on planning.

The contribution is the description and evaluation of a model-based RL architecture with planning and adaptive dynamical policy mixing.
Specifically, we:
\begin{itemize}
\itemsep=0pt
    \item introduce a mixing factor $\mu$ inversely proportional to PN quality and environment model quality,
    \item mix MCTS and PN policies, weighted according to $\mu$,
    \item set softmax temperature for MCTS action distributions according to $\mu$ and
    \item set the exploration constant that guides the MCTS selection phase according to policy quality, and
    \item boost the value function from interior MCTS tree nodes.
\end{itemize}
We investigate
\begin{enumerate}
\itemsep=0pt
    \item the effect of maintaining a value function as intermediate data element between a planner and a PN,
    \item the effect of mixing planner policies with network policies with an adaptive mixing factor,
    \item whether boosting the value function from interior MCTS tree nodes is beneficial in Flexer and
    \item how these effects change with environment complexity/size.
\end{enumerate}
on three toy environments.
Flexer is also compared with a comparable version of AlphaZero and the standard version of MuZero.


The rest of this article is organized as follows.
The next section reviews the necessary concepts and the works most related to ours.
Section~\ref{sec:our-architect} presents our proposal; the Flexer architecture.
Section~\ref{sec:eval} shows how we evaluated Flexer and the results, and Section~\ref{sec:Discussion} discusses the results.
Conclusions are discussed in Section~\ref{sec:conclusions}.

\section{Basic Concepts and Related Works}\label{sec:related-work}

Reinforcement learning (RL) techniques \cite{sb18} are usually based on the assumption that the system or environment is modeled as a Markov decision process (MDP) \cite{p94}. An MDP is a tuple $\langle S,A,T,R,s_0\rangle$, where
\begin{itemize}
    \itemsep=0pt
    \item $S$ is the set of states of the system,
    \item $A$ is the set of actions that can change the system,
    \item $T$ is the transition function, such that $T(s,a,s')$ is the probability of going from state $s$ to state $s'$ via action $a$,
    \item $R$ is the reward function, such that $R(s,a,s')$ is any real number that reflects the utility of going from state $s$ to state $s'$ via action $a$, and
    \item $s_0$ is the initial state of the system.
\end{itemize}

Given an MDP, one can search from $s_0$ forward, by selecting actions from $A$, making transitions according to $T$ and assigning values to states according to $R$. One then chooses the `best' simulated action in $s_0$ as the one to execute in the real system/environment. Monte Carlo Tree Search (MCTS) \cite{mcts_survey12} is such a search algorithm with special properties: a tree of nodes is grown by adding one node per iteration over several iterations. Every iteration has four stages: (1) \textit{select} a path from the root node, (2) \textit{expand} a node $n$ at the end of the path that has not yet been fully expanded (i.e. generate child node node $n'$ of $n$, where $n$ does not have branches out of it for every action in $A$), (3) perform a \textit{rollout} from the newly generated node (i.e. simulate a trajectory from $n'$) and (4) \textit{backpropagate} the value of the trajectory up thru the path till the root. 

Every node in the tree represents a state and every branch for an action $a$ out of a node representing $s$ stores a state-action value $\hat{Q}(s,a)$ estimating the total value for performing $a$ in $s$.  These `q-values' are updated during backpropagation.

In the select stage, standard MCTS often uses the upper confidence bound on trees (UCT) \cite{ks06} to select which branch (action $a^*$) to follow down:
\[
a^* = \arg\max_{a\in A}  \Bigg( \hat{Q}(s,a) + C \sqrt{\frac{\ln N(s)}{N(s,a)}}\Bigg),
\]
where
\begin{itemize}
\itemsep=0pt
    \item $  s  $ is the state represented by the node where the next branch must be selected,
    \item $  \hat{Q}(s,a)  $ is the average (empirical) reward observed from taking action $  a  $ in state $  s  $,
    \item $  N(s)  $ is the number of times state $  s  $ has been visited,
    \item $  N(s,a)  $ is the number of times action $  a  $ has been taken from state $  s  $ and
    \item $  C  $ is exploration constant (commonly $  \sqrt{2}  $ for theoretical guarantees, or tuned empirically).
\end{itemize}

Our architecture builds on the foundations of AlphaZero \cite{alphazero}, which was developed for two-player deterministic board games. It maintains a neural network representing the policy and value function (aka Policy Value Network or PVN for short), and is trained from data extracted from an MCTS planner.
Actions are selected from the distribution based on the visit counts of the children of the root of the plan-tree. The selection phase of the MCTS is guided by the policy head of the PVN according to the predictor upper confidence bound for trees (PUCT) \cite{r11}:
\begin{equation}
a^* = \arg\max_{a\in A} \Bigg( \hat{Q}(s,a) + C\cdot \, P(s,a) \, \frac{\sqrt{N(s)}}{1 + N(s,a)} \Bigg),
\end{equation}
where $  P(s,a)  $ is the prior probability of action $  a  $ from the learned policy of the PVN.
    
The planner uses known transition and reward models, that is, the planner has access to the real environment dynamics. 

The architecture by Anthony, Tian and Barber \cite{atb17}
which was published just before AlphaZero, is very similar.

\begin{figure}
    \centering
    \includegraphics[width=0.8\linewidth]{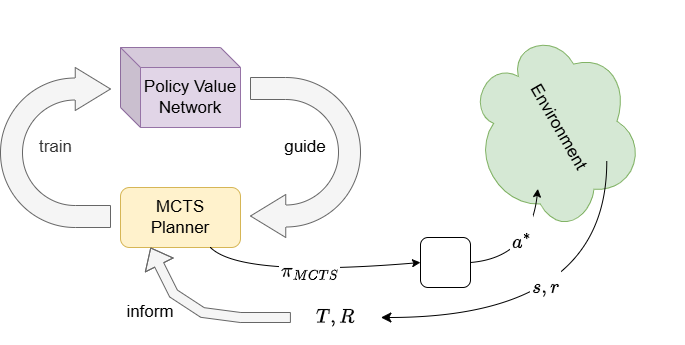}
    \caption{Conceptual representation of the AZLike architecture}
    \label{fig:azlike-arch}
\end{figure}

In our context, the agent has no access to real environment dynamics -- it must learn the transition and reward models, transitions might be stochastic, and we consider only single player problems. Hence, we implemented a baseline architecture based on AlphaZero, called AZLike. AZLike (like Flexer) is a single-player architecture that learns stochastic transition and reward models. See Figure~\ref{fig:azlike-arch}.

MuZero \cite{muzero20} is closer to Flexer in that it also learns environment models and has been applied to single-player problems, including fifty-seven Atari games. MuZero generalises AlphaZero to domains where no simulator is available.
It learns three separate neural networks — for state representation, environment dynamics, and prediction — so it can plan entirely from experience, with no simulator and no knowledge of the rules. MuZero was explicitly designed to work with standard reinforcement learning scenarios, including single-agent environments with intermediate rewards of arbitrary magnitude and with time discounting. AlphaZero, by contrast, was designed only for two-player games that could be won, drawn, or lost.

The representation network encodes a real observation into a latent vector. The dynamics network, given a latent state and action, predicts the next latent state and the immediate reward; this is the learned model, and MCTS planning happens entirely in latent space. The prediction network, given a latent state, outputs a policy (prior over actions) and a scalar value, exactly as in AlphaZero — the policy network.

At execution time, the root latent is computed via the representation net. Each simulation step expands a node by calling the dynamics net (no real environment step), then the prediction net. Action selection uses a modified PUCT rule: because state values in general RL domains are unbounded, MuZero normalises Q-values into $[0, 1]$ using the minimum and maximum values observed in the search tree up to that point before applying the PUCT formula, unlike AlphaZero which assumes values are bounded within $[0, 1]$ by construction. Unlike AlphaZero's flat transition buffer, full episodes are stored in a replay buffer, which is necessary for $K$-step unrolling during training. The parameters of all three networks are jointly trained end-to-end by backpropagation-through-time, meaning gradients flow back through the dynamics net from prediction heads at depths $1, 2, \ldots, K$. This trains the dynamics net to be a useful model for planning, not just a one-step predictor.

\cite{moaz24} focus on online model-free RL problems in non-Markovian (i.e. history based) decision processes where no simulator is available. In their approach, an agent
learns through interactions with an unknown environment without estimating the environment.
They define Monte Carlo tree learning (MCTL), which builds a tree based on agent trajectories. Flexer and MuZero do learn environment models, which are applied in MCTS.
They mention that it is generally known that a combination of models based on policy gradient (e.g. a PN) and search (e.g. MCTS) can have a positive effect, and they cite \cite{k14} in this regard.
%
At every step while learning, \textit{either the PN or MCTS policy} is selected and an action executed according to that policy. The policy selection is done with respect to a probability function $\lambda$, which may be a fixed value or parameterized (based on user-defined hyper-parameters). When training is complete, the $\lambda$-mixture of the two policies is the final learned policy.
This is different to Flexer in that actions in Flexer are based on the mixed policies at \textit{every step}.

\cite{zdwyxd26} propose a framework for flexible job shop scheduling where a pre-trained RL policy is enhanced by MCTS. In the enhancement phase, where MCTS is applied, the procedure incorporates an early-stopping mechanism based on consecutive non-improving iterative rounds to optimize efficiency. This provides a tunable quality-time trade-off, where an increased MCTS simulation budget consistently yields non-inferior solutions. Their budget-control approach is more sophisticated than Flexer's and involves a search space pruning stage. There is no mixing of PN and MCTS policies in their approach, though.

\section{The Proposed Architecture: Flexer}\label{sec:our-architect}

We shall consider four variants of Flexer (and their non-mixing AZLike counterparts).
Two variants will have rollout in the MCTS planner and two variants will not have rollout, but bootstrap newly generated node values directly from the value function, in AlphaZero style. Two variants will have tabular environment model representations, and two variants will have neural network environment model representations. There will thus be the following variants. Flexer-RT, Flexer-RN, Flexer-BT and Flexer-BN, where R = rollout, B = bootstrap, T = tabular and N = neural net. All variants have an NN-based policy representation and a tabular value function.

\subsection{Mixing Logic}\label{sec:mixing-factor}
At every step, when an agent must select an action to execute in state $s$, Flexer chooses the action with the highest value from a distribution $\pi(s)$ of values of actions, where $\pi(s)$ is the mixture of the action-distribution from the MCTS planner ($\pi_\mathit{MCTS}(s)$) and the action-distribution from the policy network (PN) ($\pi_\mathit{PN}(s)$). Formally,
\begin{equation}
\pi(s) \doteq \mu\cdot\pi_\mathit{MCTS}(s) + (1-\mu)\pi_\mathit{PN}(s),
\label{eq:mixed-policy}    
\end{equation}
where $\mu\in[0,1]$ is a mixing factor that controls the mixing ratio. The discussion about how to determine $\mu$ comes next.


Let $\kappa_\mathit{PN}\in[0,1]$ be the current quality of the PN and let $\kappa_\mathit{EM}\in[0,1]$ be the current quality of the environment model (transition model $T$ and reward model $R$).
Note that because the plan-tree structure and associated values directly depend on $T$ and $R$, results from planning depend on $\kappa_\mathit{EM}$. $\mu$ should tend to 1 as $\kappa_\mathit{PN}$ tends to 1, independent of the value of $\kappa_\mathit{EM}$. And $\mu$ should tend to 1 as $\kappa_\mathit{EM}$ tends to 1 and $\kappa_\mathit{PN}$ tends to 0. If neither the Pn nor the planner can be relied upon, then the agent should take a random action. These cases are summarized by the following table.
\vspace{3mm}

\begin{center}
    \begin{tabular}{|c|c|c|}
\hline
    $\kappa_\mathit{PN}$ & $\kappa_\mathit{EM}$ & Strategy \\
    \hline
    1 & 1 & Use PN output \\
    1 & 0 & Use PN output \\
    0 & 1 & Use MCTS result \\
    0 & 0 & Take random action \\
    \hline
\end{tabular}
\end{center}
\vspace{3mm}

Assuming that not both $\kappa_\mathit{PN}$ and $\kappa_\mathit{EM}$ are zero, defining $\mu$ to be $(1-\kappa_\mathit{PN})\times\kappa_\mathit{EM}$ would result in the strategies listed in the table. The algorithm will check for the 0/0 case before applying $\mu$.

Let $\psi_\mathit{PN}\in[0,1]$ be the (normalized) policy imitation error  \cite{atb17} of the PN (a measure of PN inference quality; more about this below). We define $\kappa_\mathit{PN}$ as $1-\psi_\mathit{PN}$.

Because there are variants of Flexer that represent environment models in tabular and in neural net forms, we need definitions of $\kappa_\mathit{EM}$ for each.
For tabular variants, Flexer keeps track of $T_{var}$, the variance of a history of transition probabilities and $R_{var}$, the variance of a history of rewards.
The exact definitions of these factors will be given shortly.
The MCTS policy should be relied on in proportion to the least of these two values: $1-T_{var}$ and $1-R_{var}$. Hence, $\kappa_\mathit{EM}$ is defined as $\min\{(1-T_{var}),(1-R_{var})\}$.
For NN variants, Flexer measures the quality of the environment model network as an exponential moving average of prediction accuracy measured after each training batch. Accuracy is the fraction of predicted next states that fall within a fixed tolerance of the true next states. It starts at zero when the network is untrained and rises toward one as the network's predictions become consistently close to the ground truth.

Above, we said that the algorithm will check for the 0/0 case before applying $\mu$. In practice, having $\kappa_\mathit{PN}=0=\kappa_\mathit{EM}$ never occurs. 
Suppose $x=(2-\kappa_\mathit{PN} -\kappa_\mathit{EM})/2$
combines PN quality and environment model quality into a single value in [0,1]. 
When both are at their worst ($\psi_\mathit{PN}=1, \kappa_\mathit{EM}=0$), $x=1$; when both are at their best ($\psi_\mathit{PN}=0, \kappa_\mathit{EM}=1$), $x=0$.
$x$ is then reshaped as 
$\mathit{rand\_act}=(e^{kx}-1)/(e^k-1)$; experiments in this work uses $k=10$. The function maps $[0,1]$ to $[0,1]$ but with an exponential shape: 
$\mathit{rand\_act}$ stays near 0 for small $x$ and rises steeply as $x$ approaches 1. This makes the agent explore frequently only when both quality signals are genuinely poor, avoiding unnecessary random actions when only one is slightly degraded.
If a uniform random draw falls below $\mathit{rand\_act}$, a random action is returned immediately, skipping MCTS entirely.



\subsection{Definitions of Mixing-Factor Factors}

\subsubsection{PN Inference Quality}\label{sec:PIE}

Right after every training step, we compute how badly the current policy network is imitating its training targets, and we maintain a smoothed (running-average) version of this measure of quality.
More precisely, we calculate the average KL divergence between two probability distributions over the recent training batch $B$: the target distribution and the current policy network’s distribution.

Let \( y_i \in \mathbb{R}^A \) be the target distribution for example \( i \) in the $B$, and let \( p_i \in \mathbb{R}^{|A|} \) be the current policy distribution for example \( i \), that is, \( p_i = \text{softmax}( \text{network}(s_i) ) \). Let
$$
\text{KL}_i = \sum_{a=1}^{|A|} y_{i,a} \log \left( \frac{y_{i,a}}{p_{i,a}} \right).
$$
Then
$$
\text{kl} = \frac{1}{|B|} \sum_{i=1}^{|B|} \text{KL}_i 
$$
is the average over the batch. This is the policy imitation error (PIE) \cite{atb17}.

Because raw KL divergence can be arbitrarily large (in theory), it is normalized by dividing by \( \log {|A|} \).
$$
\psi_{\text{cur}} = \frac{\text{kl}}{\log {|A|}}
$$
The normalized value is then used to smoothly update a persistent variable $\psi_{PN}$ using an exponential moving average. It is the long-term estimate of how far the policy network still is from matching the targets it is being trained on.
$$
\psi_{PN} \gets (1 - \alpha) \cdot \psi_{PN} + \alpha \cdot \psi_{\text{cur}}
$$
where \( \alpha \) is a smoothing factor (e.g. 0.05 to 0.2).

\subsubsection{Environment Model Uncertainty for the Tabular Case}

A transition buffer $B_\mathit{tran}$ contains a large history (e.g. 10000) of observed transitions: $(s, a, s', p)$, where $p$ is the probability of going from $s$ to $s'$ via $a$ estimated at the moment the tuple was added to the buffer. Similarly, observed rewards are stored in a reward buffer $B_\mathit{rew}$: $(s, a, s', r)$, where $r$ is the average reward experienced for $s$, $a$ and $s'$.

We measure the uncertainty of the transition and reward functions in the applicable, local state space by filtering the buffers by selecting only tuples that have first component $s$ that are in a set of states close to the agent's current state $s_{cur}$. We determine this set of local states $S_{mc}(s_{cur})$ by growing a tree (rooted at $s_{cur}$) of states in breadth-first fashion. To grow the tree, a child state $s'$ is generated from parent $s$ via action $a$ with probability $\hat{T}(s,a,s')$, that is, according to the current estimate of the transition function. The tree stops growing when the number of states it contains equals some hyper-parameter (e.g. 100).
Each state in $S_{mc}(s_{cur})$ thus defines a collection of values ($p$s or $r$s) for which we calculate the sample variance.
The final variance used is the mean of the variances of all the collections with respect to $s_{cur}$.
Mathematically, for the transition case:
Let $B_\mathit{tran}(s)$ be the set of tuples $(s,a,s',p)\in B_\mathit{tran}$ such that $s \in S_{mc}(s)$. 
Let $B_k(s)$ be a subset of $B_\mathit{tran}(s)$ such that for any two $(s_1,a_1,s'_1,p_1), (s_2,a_2,s'_2,p_2)\in B_k(s)$, $(s_1,a_1,s'_1)= (s_2,a_2,s'_2)$.
\[
var(T,s) = \frac{1}{K} \sum_{k=1}^{K} \operatorname{Var}\bigl(\{p \mid (s,a,s',p)\in B_k(s)\}\bigr),
\]
where \(K\) is the number of unique \((s,a,s')\) triples that appear in the filtered buffer $B_\mathit{tran}(s)$, and \(\operatorname{Var}\) is the usual sample variance (ddof=1).

Flexer maintains an exponentially smoothed variance tracker:
\begin{equation}
T_\text{var} \leftarrow T_\text{var} + \alpha \cdot (var(T,s) - T_\text{var}),
\label{eq:T-var}    
\end{equation}
where $\alpha$ is the step-size (e.g. 0.05).
The update for $R_\text{var}$ is similar.

In summary, Flexer continually measures how reliable the learned transition and reward models are in the local region the agent is currently thinking about, and it feeds that information into the mixing function $\mu(\cdot)$ via the exponential moving averages $T_\text{var}$ and $R_\text{var}$.

\subsubsection{Environment Model Uncertainty for the Neural Net Case}
\label{sec:EM-quality-NN}

Let $A_t$ be the state prediction accuracy at batch $t$, defined as the fraction of samples in the batch for which the predicted next state $\hat{s}'$ is within a fixed tolerance $\delta = 0.05$ of the true next state $s'$:
$$A_t = \frac{1}{B} \sum_{i=1}^{B} \mathbf{1}\left[\max_j \left|\hat{s}'_{i,j} - s'_{i,j}\right| < \delta\right],$$
where $B$ is the batch size and $\max_j |\hat{s}'_{i,j} - s'_{i,j}|$ denotes the element-wise maximum absolute deviation across the state encoding. 
$s'_{i,j}$ is the $j$-th element of the true next-state encoding vector for the $i$-th sample in the training batch. So $i$ indexes the sample and $j$ indexes the component within the state vector.

Then $q_t$ is the environment model quality after training batch $t$. Then:
$$q_t = (1 - \alpha) q_{t-1} + \alpha \cdot A_t,$$
where $\alpha$ is the smoothing factor (e.g.\ 0.3).
The quality $q_0 = 0$ at initialization, and $q_t \in [0, 1]$ for all $t$.

The policy net quality and environment model net quality are calculated in different ways They measure fundamentally different types of output, which demands different metrics.

The environment model produces a specific predicted value — a next-state vector — which can be compared directly against the observed ground truth. A simple element-wise tolerance check is the natural measure: the prediction is either close enough to the truth or it is not.

The policy network produces a probability distribution over actions. There is no single correct action to compare against; instead, the target is the MCTS visit-count distribution. The natural way to measure how far apart two distributions are is KL divergence, not a tolerance check (PIE)

Using model prediction error as a confidence measure appears in model-based RL work such as MBPO \cite{jfzl19} and PETS \cite{ccml18}, which use ensemble disagreement or prediction variance rather than accuracy.

\subsection{The State Value Function}

The state value function $V:S\mapsto \mathbb{R}$ is at the center of the Flexer architecture. It is updated at every agent step due to agent experience, and, optionally, from state values computed for nodes in the MCTS tree. It is used at leaf nodes of the MCTS tree, for training data for the PN and it is used in the MCTS rollout policy. The two ways to update the value function are discussed next. The ways it is used are discussed in the applicable sections below.

At every step, the agent observes the new state $s'$ and reward $r$.
The empirical transition model \(\hat{P}\) and reward model \(\hat{R}\), are then updated.

A single-step incremental value iteration update for the learned state-value function \(V(s)\) is performed. It approximates the Bellman optimality operator using \(\hat{P}\) and \(\hat{R}\).
Formally, after observing a transition, the value of the current state \(s\) is updated as
\[
V(s) \leftarrow V(s) + \alpha \Bigl( \max_{a \in \mathcal{A}} \hat{Q}(s,a) - V(s) \Bigr),
\]
where \(\alpha\) is the learning-rate hyper-parameter and the action-value estimate \(\hat{Q}(s,a)\) is defined piecewise from the learned model:
\[
\hat{Q}(s,a) =
\begin{cases}
r_{\text{step}} + \gamma V(s) & \text{if no transitions observed for action } a \text{ in state } s, \\
\displaystyle\sum_{s' \in \mathcal{S}} \hat{T}(s,a,s') \Bigl( \hat{R}(s,a,s') + \gamma V(s') \Bigr) & \text{otherwise}.
\end{cases}
\]

The empirical transition probabilities and rewards are obtained directly from the stored counts:
\[
\hat{T}(s,a,s') = \frac{N(s,a,s')}{\sum_{s''} N(s,a,s'')}, \qquad
\hat{R}(s,a,s') = \frac{\sum r(s,a,s')}{N(s,a,s')},
\]
where \(N(s,a,s')\) is the number of times transition \((s,a,s')\) has been observed and the reward sum $\sum r(s,a,s')$ is stored. If an action has never been tried from \(s\), the algorithm falls back to the prior (stay in \(s\), receive the constant step cost; e.g. \(r_{\text{step}} = -1\)).

The incremental form \(V(s) \leftarrow V(s) + \alpha (\cdot)\) is mathematically equivalent to the standard value-iteration update with learning rate \(\alpha\) and ensures smooth, online convergence even when the model is still being populated.
This fresh value is then added to the value buffer as discussed earlier.

\subsection{Value-function Boosting from Interior MCTS Nodes}\label{sec:boosting}

Some versions of Flexer boost the global value function (table) by blending values of all interior nodes of the MCTS tree directly after the planner is called. We consider two kinds of node values: One is based on visit-weighted q-values of children and the other is based on maximum q-values leading to the children.

Broadly, our method is as follows.
Recurse through all visited internal nodes of the search tree;
Compute the value target at each node from local Q-statistics (max-Q or visit-weighted Q);
Apply a depth-discounted learning rate ($\alpha$ / (depth+1)) to update a global value function.
Algorithm~\ref{alg:update_v_max_q} is the pseudocode for the maximum q-values version; the visit-weighted q-values version is symmetrical.

\begin{algorithm}
\caption{Update Value Function from MCTS Tree (Max-Q Style)}
\label{alg:update_v_max_q}
\begin{algorithmic}[1]
\Procedure{UpdateVFromTreeMaxQ}{root}
    \State Initialize $Queue \gets \{(root, 0)\}$
    \While{$Queue \neq \emptyset$}
        \State $(node, d) \gets Queue.\text{dequeue}()$
        \If{$node.N_s > 0$}
            \State $tried \gets \{a \mid node.N[a] > 0\}$
            \State $v \gets \max_{a \in tried} node.Q[a]$
            \State $\alpha_d \gets \alpha / (d + 1)$
            \State $V(node.state) \gets V(node.state) + \alpha_d (v - V(node.state))$
        \EndIf
        \ForAll{$child \in node.children$}
            \State $Queue.\text{enqueue}((child, d+1))$
        \EndFor
    \EndWhile
\EndProcedure
\end{algorithmic}
\end{algorithm}


The work of Efroni et al.\ \cite{edsm19} is the most direct theoretical match. They explicitly argue that backing up only at the root is non-contractive and propose value updates along the optimal tree path with depth-discounted ($\gamma^h$) rates. However, it is a tabular/theoretical treatment -- no neural network, no visit-count weighting, and updates only along the optimal path, not all visited nodes.

ReST-MCTS \cite{zzhydt24} derives quality-value targets for every node in a reasoning tree and trains a process reward model on all of them. The spirit of their approach matches ours, but it is in the LLM / chain-of-thought domain, not RL.

Willemsen et al.\ \cite{wbk22} introduced a family of AlphaZero value targets culminating in \textit{AlphaZero with Greedy Backups} (A0GB). Rather than using only the terminal game outcome as the value training target, A0GB uses Q-values from internal tree nodes — specifically the leaf of a greedy path through the tree — as the training target. They showed A0GB can find the optimal policy in tabular domains where the original AlphaZero target fails, and achieves faster training on Connect-Four and Breakthrough.
Their approach focuses on the value-head training targets, not directly the policy-head.

We could not find any published work that use all internal nodes and applying a depth-discounted learning rate.


\subsection{Policy Network Training}\label{sec:PN-training}

The policy network (PN) is trained via MSE regression.
Let $f_\theta(s)$ be the output of the PN for input $s$ and $\mathbf{y}_{\text{greedy}}$ be a hard one-hot target distribution over actions using one-step look-ahead with the value function \(V(s)\).
So the training targets are the greedy policy induced by the current learned value function $V(s)$ and the empirical models $\hat{P}, \hat{R}$.
These (state, target) pairs are stored in an experience buffer $B_{PN}$.

Let $\mathcal{M}(s,a)$ denote the learned model based on $\hat{P}$ and $\hat{R}$. $\mathbf{y}_{\text{greedy}}$ is computed as follows.


$$\hat{s}', r \;\sim\; \mathcal{M}(s, a)$$

$$Q(s,a) = r + \gamma\, V(\hat{s}') \qquad \forall\, a \in A$$

$$a^* = \arg\max_{a \in A}\; Q(s,a)$$

$$A^* = \bigl\{\, a \in A \;\big|\; Q(s,a) = Q(s,a^*) \,\bigr\}$$

\begin{equation}
    \mathbf{y}_{\text{greedy}} = \begin{cases} \dfrac{1}{|\mathcal{A}^*|} & \text{if } a \in A^* \\[6pt] 0 & \text{otherwise} \end{cases}
    \label{eq:y-greedy}
\end{equation}

When the buffer exceeds the minimum size, the training process samples batches and minimizes MSE loss between the network’s raw logits and the greedy, $V(s)$-based targets, plus a small entropy regularizer:
\[
\mathcal{L} = \| f_\theta(s) - \mathbf{y}_{\text{greedy}}  + \epsilon \cdot H(\text{softmax}(f_\theta(s))).\|^2
\]

Training runs for a fixed number of epochs per call. Calls to the trainer occur periodically; in this study, we train at every episode.
Flexer has an adaptive epochs approach: the number of epochs per call is based on current policy quality $\psi_{PN}$ as discussed in Section~\ref{sec:PIE}. This code snippet should illustrate the approach:
\begin{verbatim}
    base_epochs = 3
    extra_epochs = max(0, int(7 * self.pie))   # up to +7 when PIE approx. 1.0
    num_epochs = base_epochs + extra_epochs       # 3..10
\end{verbatim}


A probability distribution is calculated for each node with at least one child. Let $n$ be such a node.
Let $A_n$ be the set of actions that have been tried in $n$, $N(a)$ be the number of times the child reached by action $  a  $ was visited, and $N_{\text{total}} = \sum_{a \in A_n} N(a) $.

Then the target vector $\pi_{\text{target}} \in \mathbb{R}^{|A|}$ is:
$$\pi_{\text{target}}[a] = 
\begin{cases}
\frac{N(a)}{N_{\text{total}}} & \text{if action } a \text{ has a child (i.e. was expanded)} \\
0 & \text{otherwise}
\end{cases}$$

Finally, $(s_n,\pi_{\text{target}})$ is added to $B_{PN}$, where $s_n$ is the state represented by node $n$.

\subsection{Flexer's MCTS}
PUCT is used for the selection strategy to descend down the tree thru every fully expanded node.
As soon as a node $n$ that is not fully expanded is encountered, one of the untried actions $a$ is selected uniformly randomly. A new node $n'$ which is a child of $n$ is generated and a value is assigned to $n'$ by either (i) rollout initiated from $n'$ or (ii) bootstrapping from the value function w.r.t.\ $n'$. 

If the algorithm variant uses rollout, there are two versions, depending whether the algorithm a neural net or a tabular representation of environment models. If tabular, the rollout strategy is guided by one-step greedy look-ahead with a probability (e.g. 20\%) of a random action being chosen at each step.
That is, every non-random action is decided by $\hat{R}(s) +\gamma V(s')$, where $s'$ is sampled from $\hat{T}(s,a^*,\cdot)$ and $a^*$ is the one-step greedy action from $s$. 

If the environ-models are neural nets, then performing multiple inference calls to access $\hat{R}(s)$ and $\hat{T}(s,a^*,\cdot)$ are too expensive. So these variants maintain a tabular representation alongside the neural net representation. When in a tree node for state $s$, a random successor $s'$ of $(s,a^*)$ is chosen from a set of experienced successors for $(s,a^*)$. If this set is empty, the $s'$ is assigned $s$. 
 
Rollout depth is the maximum tree depth (a given parameter) minus the depth of $n'$. The last node $n_\mathit{last}$ generated gets value $V(s_\mathit{last})$ (where $n_\mathit{last}$ represents $s_\mathit{last}$).
Also, if the maximum tree depth is reached before rollout is called, that node gets $V(s_\mathit{last})$. 

The results of a call to MCTS for an agent step must be a sum-to-one distribution of non-negative values ($\pi_\mathit{MCTS}(s)$). This is required for mixing with the results from the PN output. 

The computation of $\pi_\mathit{MCTS}(s)$ is based on softmax. The question is, what should the temperature term $\tau$ be? In this work, we set $\tau = \max\{0.01, 0.2 \mu\}$, where $\mu$ is the mixing factor (see Eq.~\ref{eq:mixed-policy}).
We argue that $\tau$ should decrease as $\mu$ \textit{decreases}, because less planning budget is more likely to result in more uniform visit counts, and keeping this uncertainty is important. Put another way, $\tau$ should increases as $\mu$ \textit{increases}, because MCTS will have more budget when $\mu$ is larger, and more planning results in sharper MCTS root visit count distribution, which should be sharpened.
AZLike does not maintain a $\mu$ variable; it thus uses $\tau=0.2$, which is effectively like always assuming $\mu=0.5$.

Another parameter that is modulated by $\mu$ is the planning budget. This is done by weighting the maximum number of MCTS iterations by $\mu$. Recall from Section~\ref{sec:mixing-factor} that as $\mu$ tends towards 1, planning should dominate, that is, MCTS should run with the maximum possible (user defined) number of iterations.
AZLike does not maintain a $\mu$ variable; it thus uses always uses half the max number of iterations used in Flexer (as if $\mu$ is always 0.5).

\subsection{A Flexer Algorithm}

\begin{figure}
    \centering
    \includegraphics[width=1.0\linewidth]{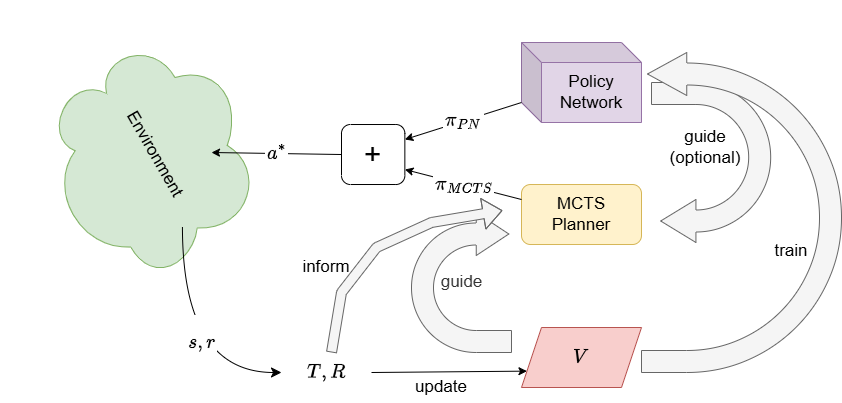}
    \caption{Conceptual representation of the Flexer architecture}
    \label{fig:Flexer-arch}
\end{figure}

Figure \ref{fig:Flexer-arch} shows the conceptual representation of the Flexer architecture. Note the addition of the value function (compared to AZLike architecture; Fig.~\ref{fig:azlike-arch}) and the function's interaction with the policy network and MCTS planner.

For context, the training loop is presented in Algorithm~\ref{alg:Flexer-train}. To keep the evaluation as simpe as possible, $\epsilon$-greedy is used for all but ADP, which has its own exploration strategy. Note that $\epsilon$ decays from 1 to 0 from the first to the last episode. This is likely suboptimal exploration, but investigating exploration strategies for Flexer is not an aim of this study.

\begin{algorithm}[t]
\caption{Flexer Training Loop}
\label{alg:Flexer-train}
\begin{algorithmic}[1]
\Require number of episodes $N_{ep}$, $\epsilon_0 = 1.0$
\For{$ep = 1,2, \ldots, N_{ep}$}
    \State $s \gets$ \textsc{InitializeEnviron}$()$
    \Repeat
        \State $\mathit{frac}\gets  \min(1.0, ep / \max(1, N_{ep}))$
        \State $\epsilon \gets \epsilon_0 \times (1.0 - \mathit{frac})$
        \If{$\text{rand}() < \epsilon$}
            \State $a \gets$ uniform random action in $A$
        \Else
            \State $a \gets \textsc{GetMixedAction}(s)$
        \EndIf
        \State $(s', r, \text{done}) \gets$ \textsc{Step}$(a)$
        \State Update transition and reward buffers with $(s, a, r, s')$
        \State Add $(s,\mathbf{y}_{\text{greedy}})$ to experience buffer $B_\mathit{PN}$, where $\mathbf{y}_{\text{greedy}}$ is defined in Eq.~\ref{eq:y-greedy}
        \State $s \gets s'$
    \Until{done}
    \If{$ep > 0$ and $|B_\mathit{PN}| \geq B_{\min}$}
        \State Train policy network $\pi_\mathit{PN}$ on $B_\mathit{PN}$
    \EndIf
\EndFor
\end{algorithmic}
\end{algorithm}

Most of the main contributions can be seen in Algorithm~\ref{alg:get-mixed-action}. This is the version for Flexer-RT and Flexer-BT. For Flexer-RN and Flexer-BN, the only difference is that environment model (EM) quality ($\kappa_\mathit{EM}$) is based on the NN representing the EM, not on the variance of transition and reward function values. For Flexer-RN and Flexer-BN, $\kappa_\mathit{EM}$ is computed and maintained in a EM training process that runs once per episode (see \S~\ref{sec:EM-quality-NN}).

A link to a repository for all code used in this study will be provided if this article is accepted for publication.

\begin{algorithm}[t]
\caption{\textsc{GetMixedAction}}
\label{alg:get-mixed-action}
\begin{algorithmic}[1]
\Require current state $s$, policy network $\pi_\mathit{PN}$, MCTS iterations $I$, MCTS depth $D$;\\
maintained scalars $\psi_\mathit{PN}$, $\kappa_\mathit{EM}$, $T_{\mathrm{var}}$, $R_{\mathrm{var}}$.
\State $\psi_\mathit{PN}$ is computed and maintained in a PN training process that runs once per episode; see \S~\ref{sec:PIE}
\State Compute $\pi_\mathit{PN}(s): \mathbf{p}^{\text{pn}} \in \Delta^{|\mathcal{A}|}$; shift and renormalize so $\mathbf{p}^{\text{pn}} \geq 0$, $\|\mathbf{p}^{\text{pn}}\|_1 = 1$
\State $\mathcal{S}_{\text{tree}} \gets$ reachable states from $s$ under learned model
\State $T_\text{var} \leftarrow T_\text{var} + \alpha \cdot (var(T,s,\mathcal{S}_{\text{tree}}) - T_\text{var})$ \Comment{see \eqref{eq:T-var}}
\State $R_\text{var} \leftarrow R_\text{var} + \alpha \cdot (var(R,s,\mathcal{S}_{\text{tree}}) - R_\text{var})$ \Comment{$\alpha=0.05$, e.g.}
\State $\kappa_\mathit{EM}\gets\min\!\bigl(1 - T_{\mathrm{var}},\; 1 - R_{\mathrm{var}}\bigr)$
\State $x\gets(\psi_\mathit{PN} + 1 -\kappa_\mathit{EM})/2$ \Comment{equivalently, $x\gets(2 -\kappa_\mathit{PN}-\kappa_\mathit{EM})/2$}
\State $k\gets10$
\State $\mathit{rand\_act} \gets \mathit{rand\_act}=(e^{kx}-1)/(e^k-1)$ \Comment{dealing with cases when $\kappa_\mathit{PN}\approx\kappa_\mathit{EM}\approx0$}
\If{$\mathit{rand\_act} > 0$ \textbf{ and } $\mathrm{rand}() < \mathit{rand\_act}$}
    \State \Return uniform random action from $A(s)$
\EndIf
\State $\mu \gets \psi_\mathit{PN} \cdot \kappa_\mathit{EM}$
\State $I_{\mathrm{eff}} \gets \lfloor I \cdot \mu \rfloor$
\State $\mathrm{root} \gets \mathrm{MCTS}(s,\, I_{\mathrm{eff}},\, D)$
\State $\tau \gets \max\!\bigl(0.01,\; 0.2\,\mu\bigr)$
\State $\pi_{\mathit{MCTS}\;a} \propto \exp\!\bigl(n_a / \tau\bigr)$ for each action $a$, where $n_a$ is the visit count of child $a$
\State $\pi(s) \doteq \mu\cdot\pi_\mathit{MCTS}(s) + (1-\mu)\pi_\mathit{PN}(s)$
\State \Return $\arg\max_a\; \pi(s)_a$
\end{algorithmic}
\end{algorithm}

\section{Evaluation}\label{sec:eval}

Experiments are performed on three environments, each with an easier and harder version. First, all variants of Flexer and AZLike are run on the easy versions of the environments. Second, the best performing variant of Flexer (in terms of success rate and number of steps until maximum return) is run on all harder versions of the environments, and the best performing variant of AZLike is run on all harder versions of the environments, with a third the number of training epochs and half the number of MCTS iterations (reasons provided later). Thirdly, the best Flexer variant is augmented with value-function boosting run on the three harder environments. Fourthly, the best Flexer variant is compared to Double DQN \cite{vgs16} and ADP \cite{sb18} all environment versions. The specific version of ADP we implement is based on prioritized sweeping \cite{ma93} which operates on a similar computation regime as the Dyna architecture \cite{s90}. Our version also employes R-max exploration \cite{bt02}.

For all versions of Flexer and AZLike, MCTS depth and iterations are fixed for a given problem (environment) instance. They are chosen to be as small as reasonably possible (to save experimentation time). Neural network training parameters were chosen via preliminary experiments; these parameters are constant and the same for all variants of Flexer and AZLike, and for all problems, except when a third of the number of epochs are used with AZLike. The reason we use fewer epochs is to keep the comparison more fair: more training epochs does not necessarily result in better return, and running-time suffers. Due to \texttt{num\_epochs = 3 + max(0, int(7 * self.pie))} for Flexer (see \S~\ref{sec:PN-training}), \texttt{num\_epochs} is typically $\sim 4.75$.

\begin{itemize}
    \item policy replay buffer size: 10000
    \item env-model replay buffer size: 40000
    \item neursl-net learning rates: 0.001
    \item number of training epochs for Flexer: 3 .. 10
    \item number of training epochs for AZLike: 3
    \item batch size: 64
    \item discount factor ($\gamma$): 0.95
\end{itemize}

Algorithm performance will be measured according to four quantities:
\begin{enumerate}
    \item Number of successful completions (as defined by the environment).
    \item Maximum (sustained) return.
    \item Number of steps per episode, where fewer steps means earlier success.
    \item Time per step (action).
\end{enumerate}

The maximum number of steps allowed per episode is 200, in all cases. All experiments on easy problems were run for thirty environment instances and on hard problems for sixty instances, and reported results are the averages for those runs.

Experiments were run on a machine with
Processor:	13th Gen Intel(R) Core(TM) i7-1365U (1.80 GHz) and
Installed RAM:	16.0 GB (15.4 GB usable).
Operating System: Microsoft Windows Enterprise 11.
IDE: Visual Studio Code.
Programming language and interpreter: Python 3.1.1.9.

\subsection{Environments for Evaluation}\label{sec:environs}
\begin{figure}[h]
  \centering
  \includegraphics[width=0.22\textwidth]{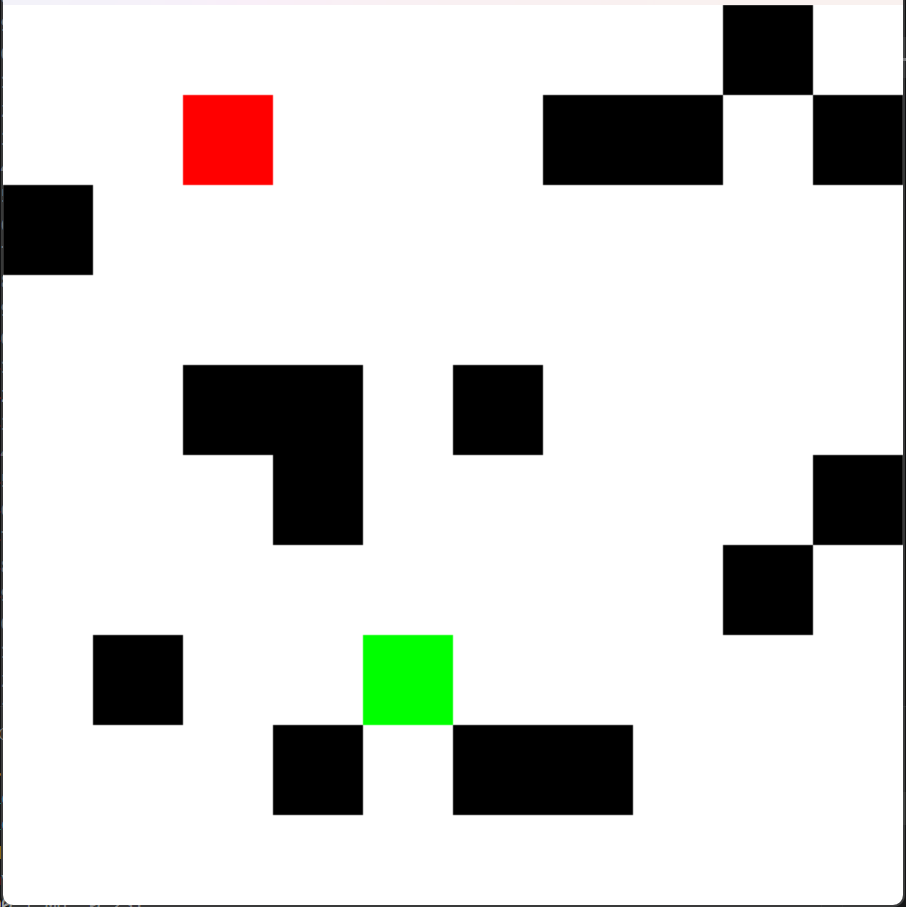}
   \hspace{15mm}
  \includegraphics[width=0.16\textwidth]{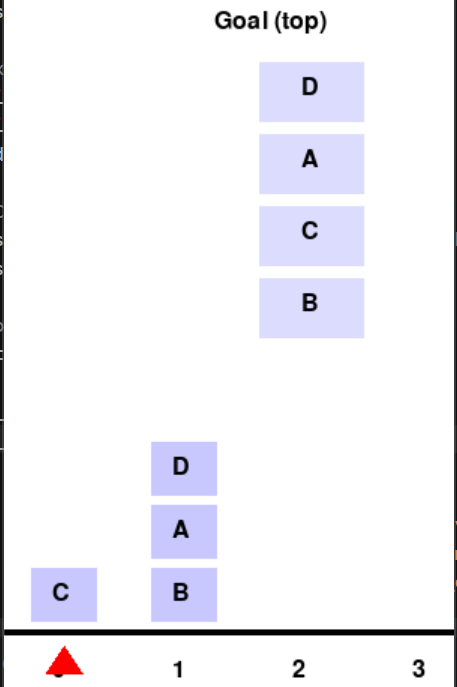}
   \hspace{15mm}
  \includegraphics[width=0.22\textwidth]{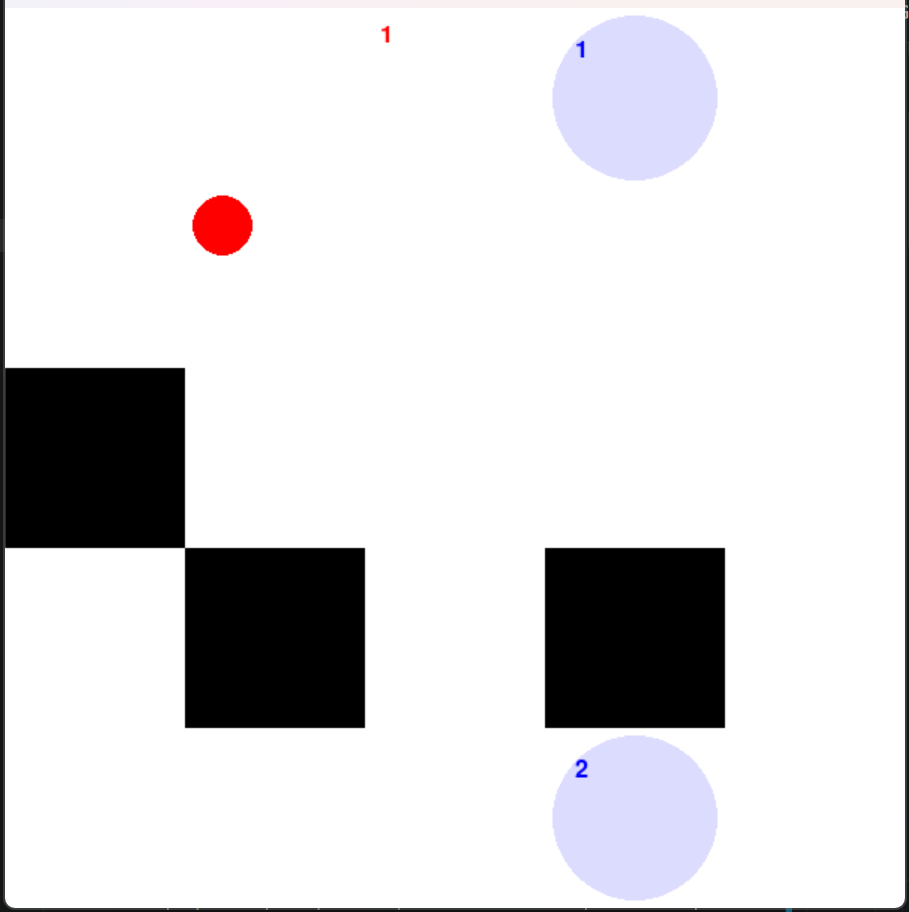}
  \caption{Environments (left to right): SimpleGrid[10,0.15], BlocksWorld[4,4], MovingNumbers[5,2]}
  \label{fig:environs}
\end{figure}

Evaluation was done on three environments: SimpleGrid, where the agent must reach a goal thru a maze, BlocksWorld, where blocks must be stacked into a particular ordering, and MovingNumbers, where the agent must move numbers to corresponding locations in a particular order. See Figure~\ref{fig:environs}. Details of each environment are given in the following subsections.

\subsubsection{SimpleGrid}

\textbf{Structure:}
SimpleGrid[N,O] is an N-by-N grid with fraction O of the cells occupied by unpenetrable obstacles. For each instance, the obstacles, one goal and the agent are placed at random locations.
\textbf{Movement:}
The agent can move in the four cardinal directions with a 10\% chance of moving perpendicular to the intended direction.
\textbf{Rewards:}
The agent gets 10 points for reaching the goal, else a small reward inversely proportional to the Manhattan distance to the goal.

\subsubsection{BlocksWorld}

\textbf{Structure:}
BlocksWorld[M,N] is the problem of stacking N of M blocks in a particular order (the goal). For each instance, the M blocks are placed randomly in one of M positions. Equivalently, there are M stacks of height zero to M. The goal is a stack of height N (in any of the M positions).
\textbf{Movement:}
The agent can move a gripper left and right and can grasp and drop a block. Drop slip: with 10\% probability the block lands one position left (5\%) or right (5\%) of the gripper.
\textbf{Rewards:}
If the resulting stack at the drop position matches the goal sequence from the bottom up (any prefix), +1 is added. This fires on every drop that extends the correct goal prefix.
When any stack matches the full goal sequence, +200 is added.
Every step costs 1 point.
Illegal actions (e.g. moving left when already at position 0, picking when holding something): the action is silently ignored but the -1 step cost is still incurred.

\subsubsection{MovingNumbers}

\textbf{Structure:}
MovingNumbers[N,O,T] is an N-by-N grid with fraction O of the cells occupied by unpenetrable obstacles and there are T tasks.
Tasks are ...
For each instance, the obstacles, one goal and the agent are placed at random locations.
\textbf{Movement:}
The agent can move in the four cardinal directions with a 10\% chance of moving perpendicular to the intended direction.
\textbf{Rewards:}
The agent gets 10 points for reaching the goal, else a small reward inversely proportional to the Manhattan distance to the goal.

\subsection{Flexer and AZLike on Easy Environs}

In variants using rollout, rollout depth is maximally 10. Hence, in variants using bootstrapping, MCTS tree depth is 10 more than rollout variants. For all experiments in this subsection, maximum MCTS tree depth is 15 for algorithms using rollout and 25 for bootstrapping. Maximum MCTS iterations is 50 for Flexer and 25 for AZLike.

\begin{figure*}[htb!]
    \centering
    \begin{subfigure}{0.4\textwidth}
        \includegraphics[width=\textwidth]{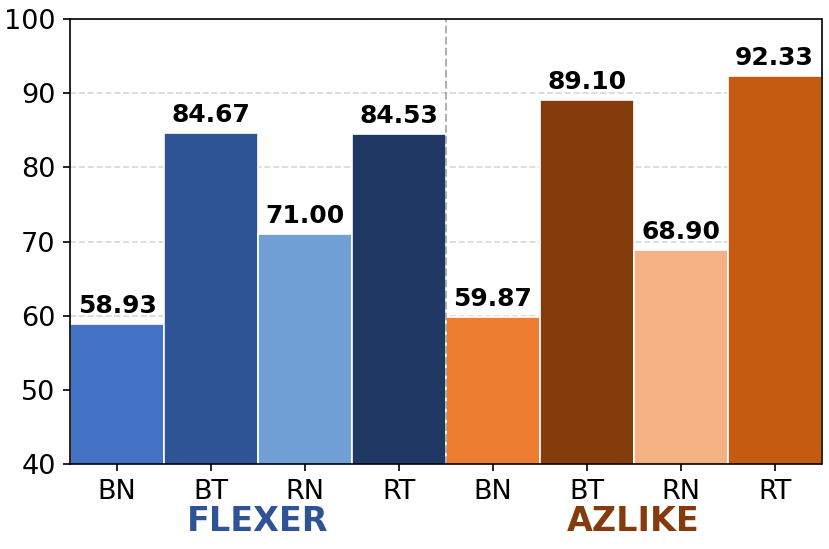}
        \caption{Goals completed}
        \label{fig:simplegrid-easy-goals}
    \end{subfigure}
    \hfill
    \begin{subfigure}{0.48\textwidth}
        \includegraphics[width=\textwidth]{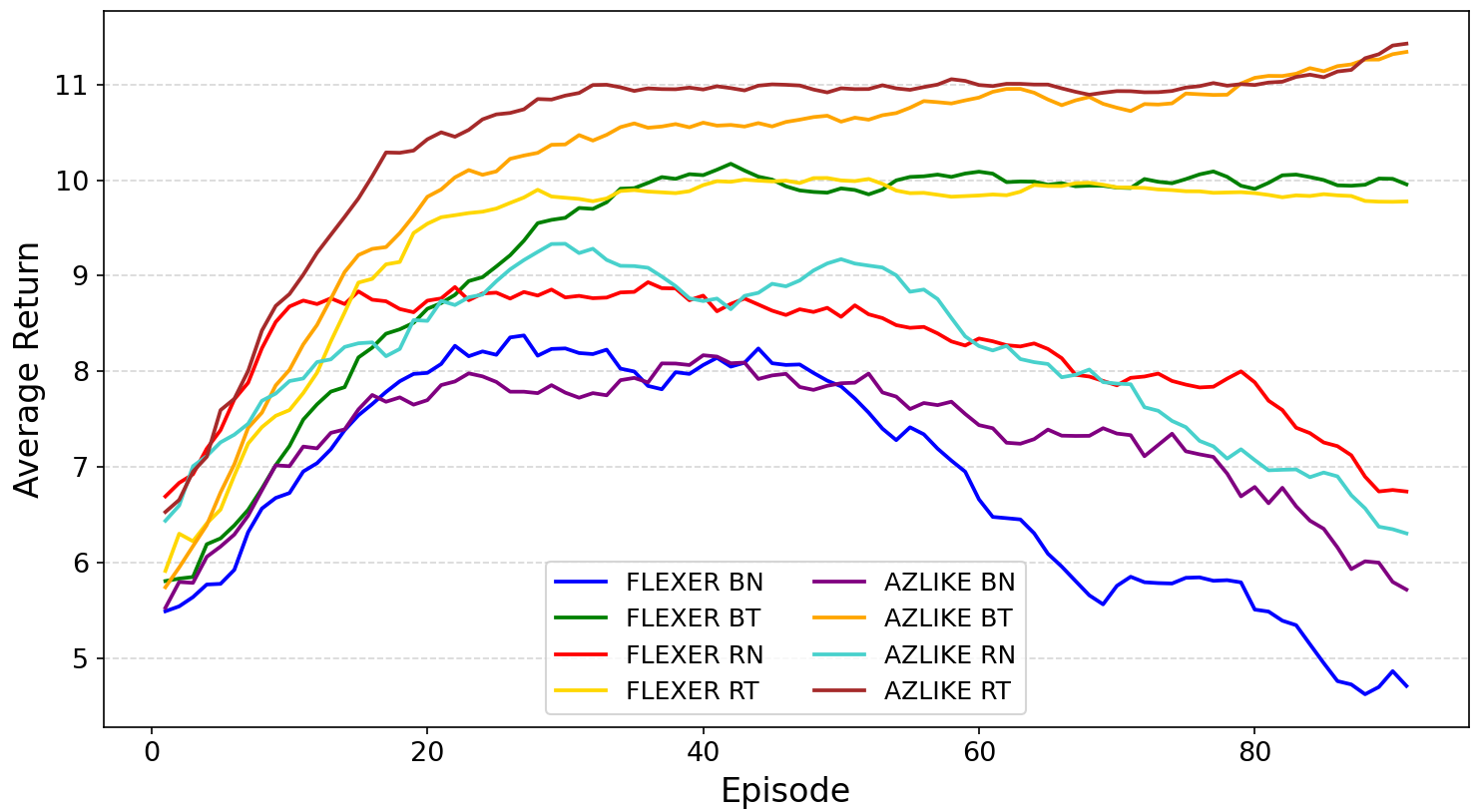}
        \caption{Return}
        \label{fig:simplegrid-easy-returns}
    \end{subfigure}
    
    \begin{subfigure}{0.48\textwidth}
    \vspace{4mm}
        \includegraphics[width=\textwidth]{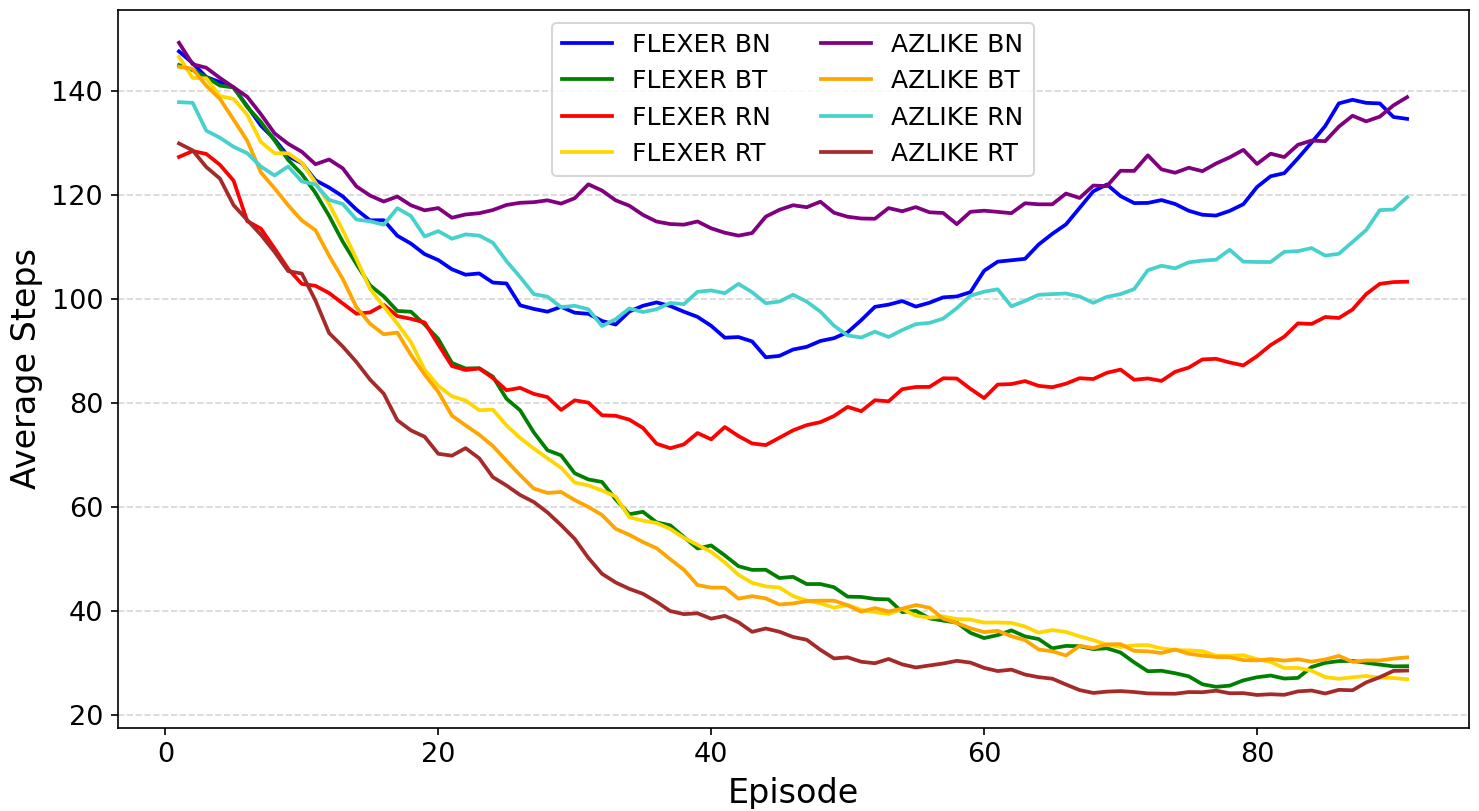}
        \caption{Steps in episode}
        \label{fig:simplegrid-easy-steps}
    \end{subfigure}
    \hfill
    \begin{subfigure}{0.48\textwidth}
        \includegraphics[width=\textwidth]{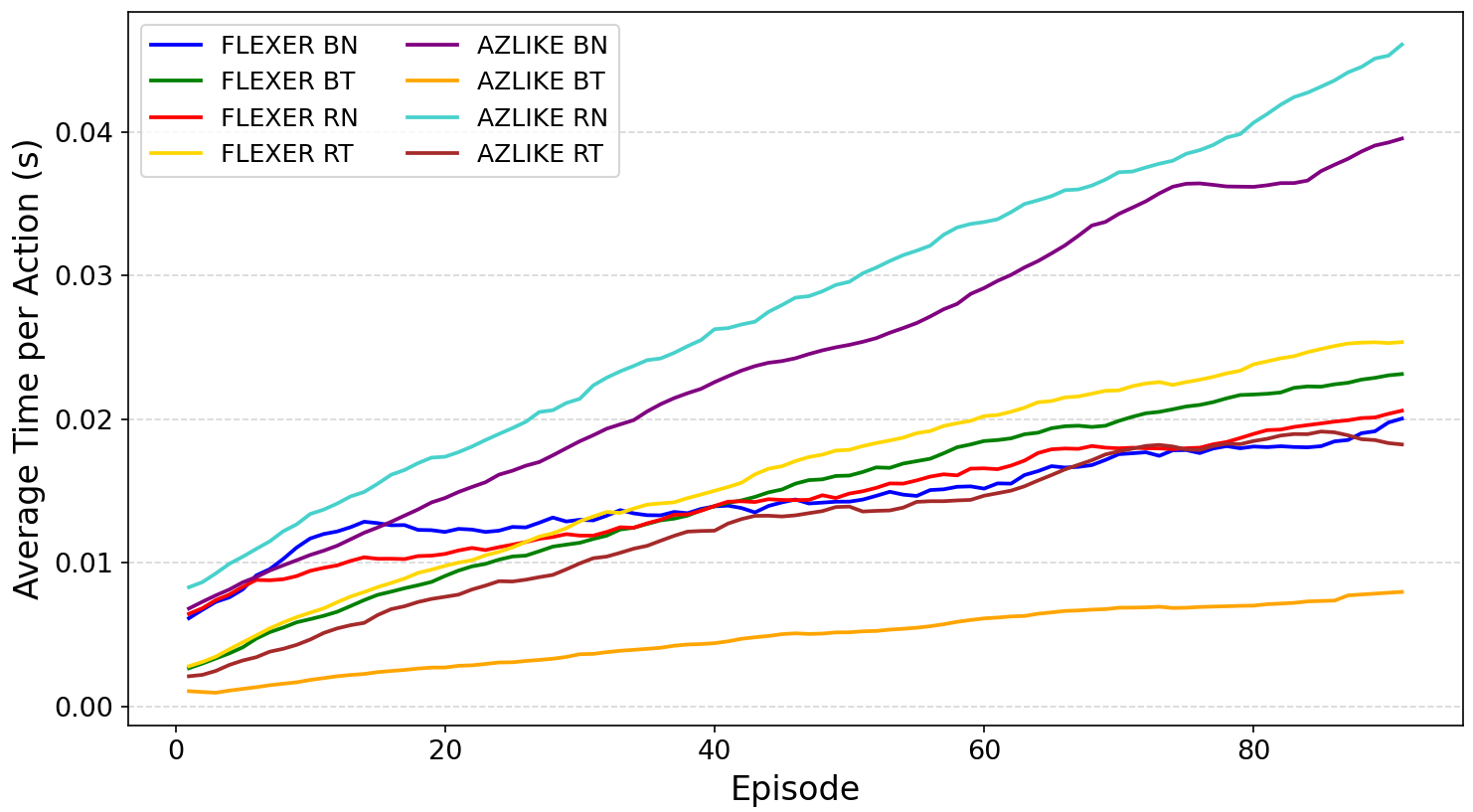}
        \caption{Time per step}
        \label{fig:simplegrid-easy-times}
    \end{subfigure}
    \caption{Results of experiments running Flexer and AZLike on the SimpleGrid[10,0.15] environment.}
    \label{fig:simplegrid-easy}
\end{figure*}

For SimpleGrid[10, 0.15], each environment instance is run 100 times. 
Figure \ref{fig:simplegrid-easy-goals} shows the number of goals achieved.
Figure \ref{fig:simplegrid-easy-returns} shows the returns.
Figure \ref{fig:simplegrid-easy-steps} shows the number of steps till goal achievement.
Figure \ref{fig:simplegrid-easy-times} shows the total time per action/step.

\begin{figure*}[htb!]
    \centering
    \begin{subfigure}{0.4\textwidth}
        \includegraphics[width=\textwidth]{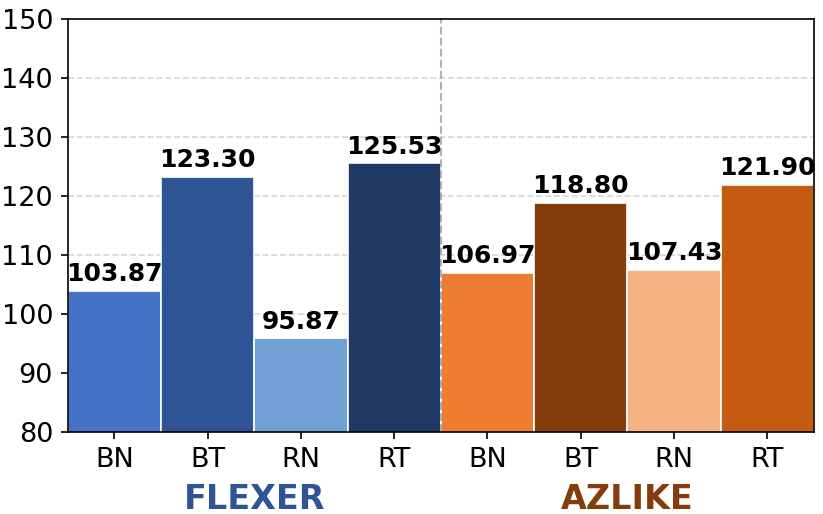}
        \caption{Goals completed}
        \label{fig:blocksworld-easy-goals}
    \end{subfigure}
    \hfill
    \begin{subfigure}{0.48\textwidth}
        \includegraphics[width=\textwidth]{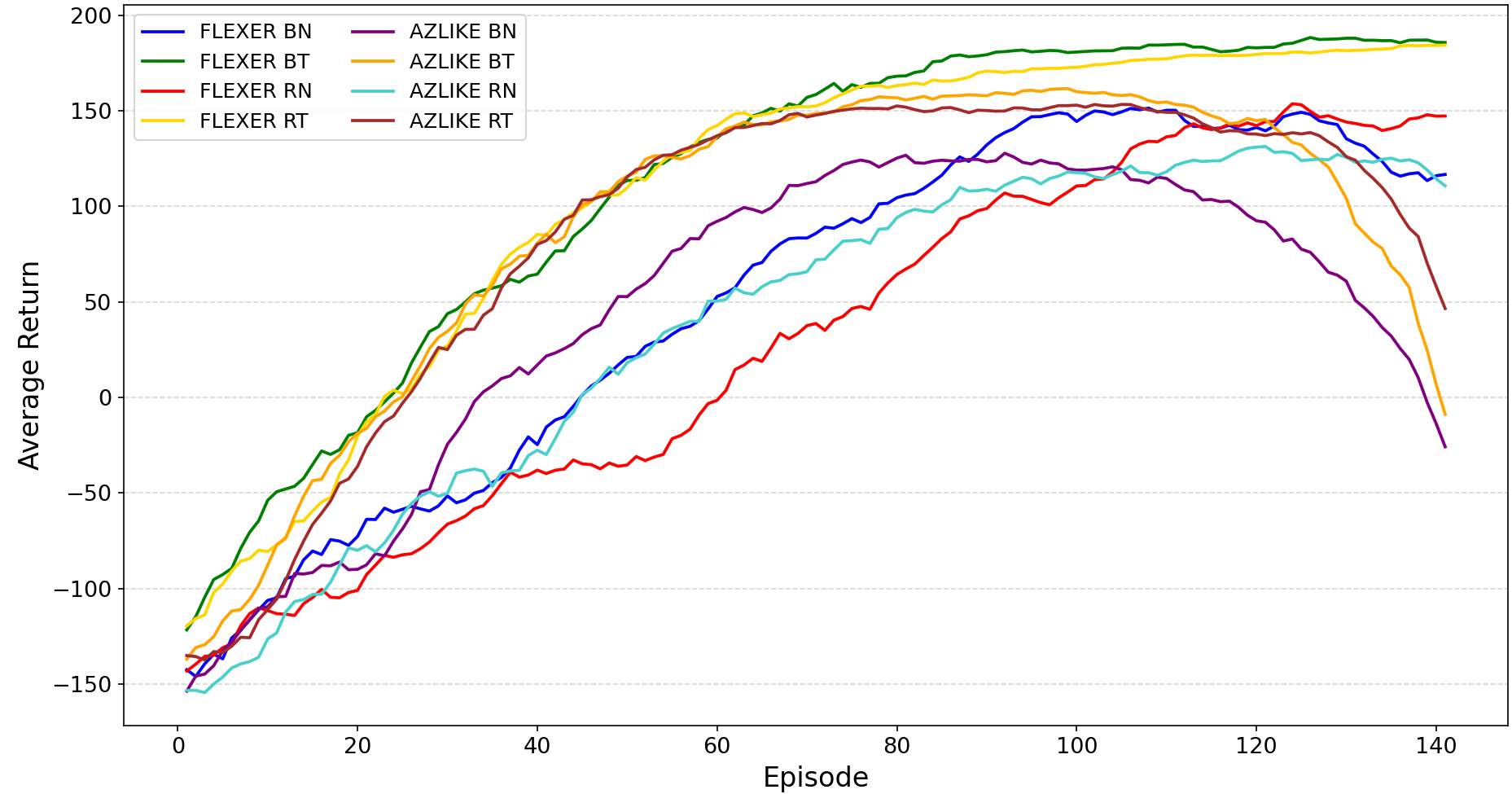}
        \caption{Return}
        \label{fig:blocksworld-easy-returns}
    \end{subfigure}
    
    \begin{subfigure}{0.48\textwidth}
    \vspace{4mm}
        \includegraphics[width=\textwidth]{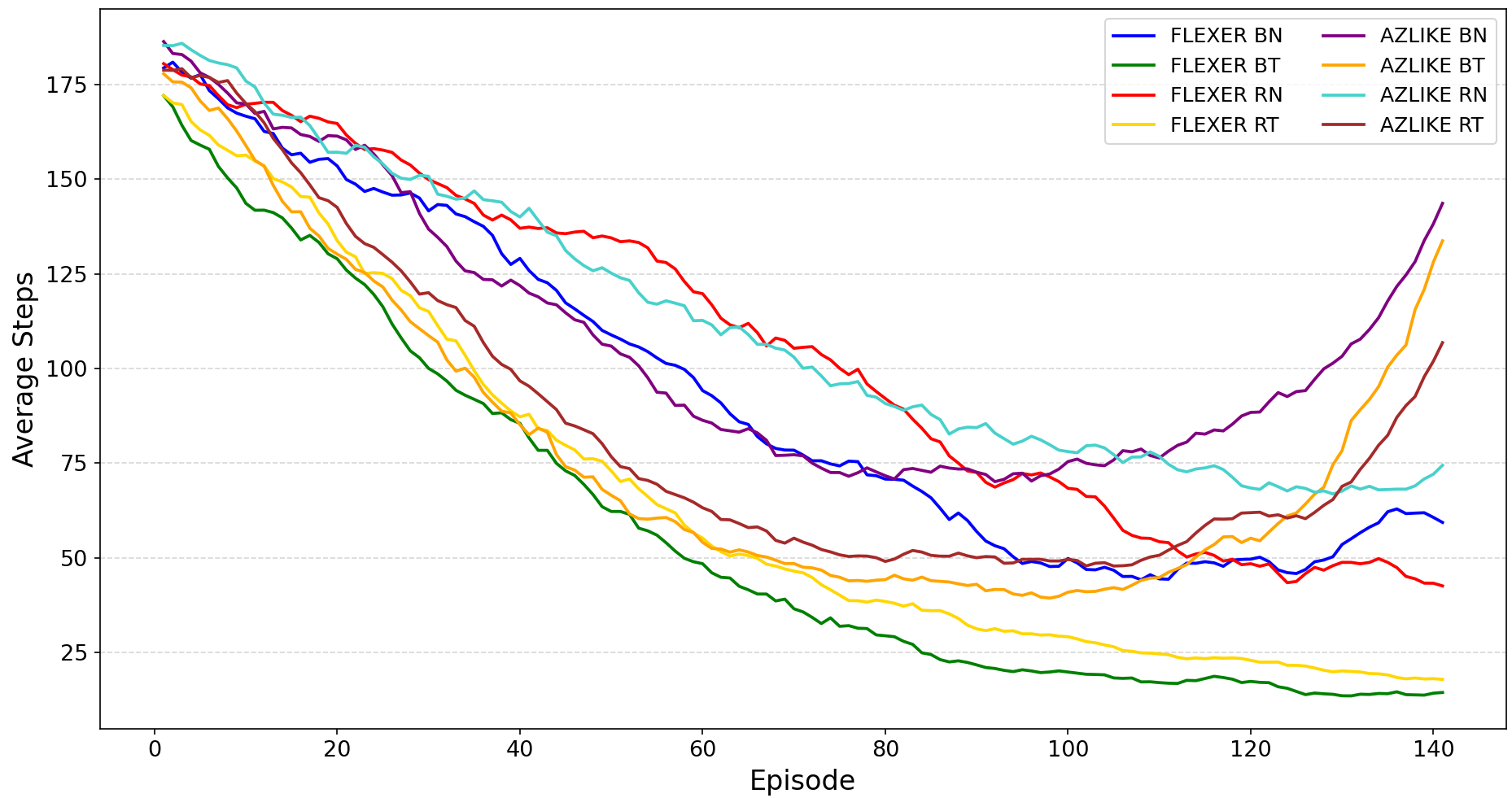}
        \caption{Steps in episode}
        \label{fig:blocksworld-easy-steps}
    \end{subfigure}
    \hfill
    \begin{subfigure}{0.48\textwidth}
        \includegraphics[width=\textwidth]{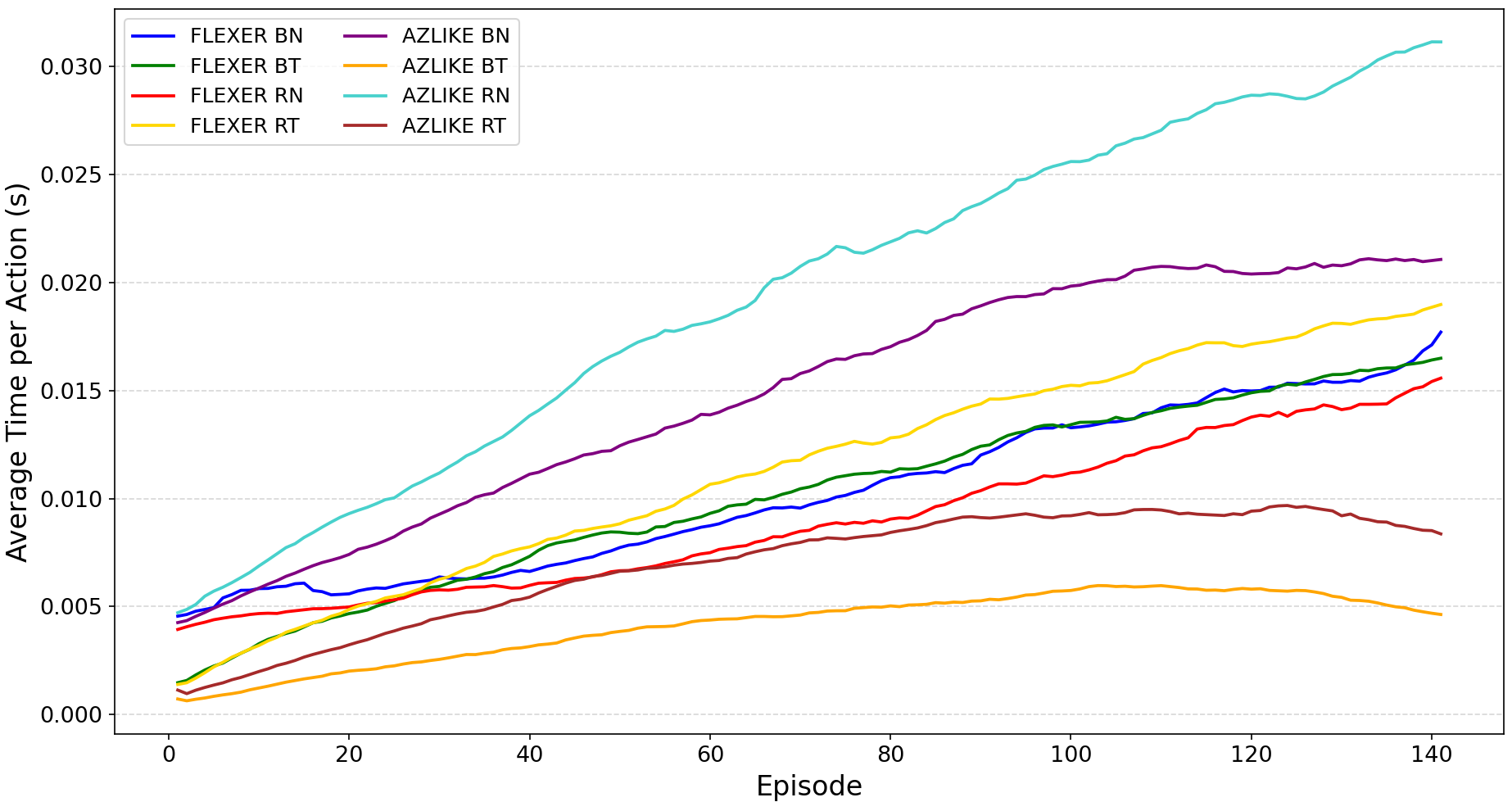}
        \caption{Time per step}
        \label{fig:blocksworld-easy-times}
    \end{subfigure}
    \caption{Results of experiments running Flexer and AZLike on the BlocksWorld[3,3] environment.}
    \label{fig:blocksworld-easy}
\end{figure*}

For BlocksWorld[3, 3], each environment instance is run 150 times.
Figure \ref{fig:blocksworld-easy-goals} shows the number of goals achieved.
Figure \ref{fig:blocksworld-easy-returns} shows the returns.
Figure \ref{fig:blocksworld-easy-steps} shows the number of steps till goal achievement.
Figure \ref{fig:blocksworld-easy-times} shows the total time per action/step.

\begin{figure*}[htb!]
    \centering
    \begin{subfigure}{0.4\textwidth}
        \includegraphics[width=\textwidth]{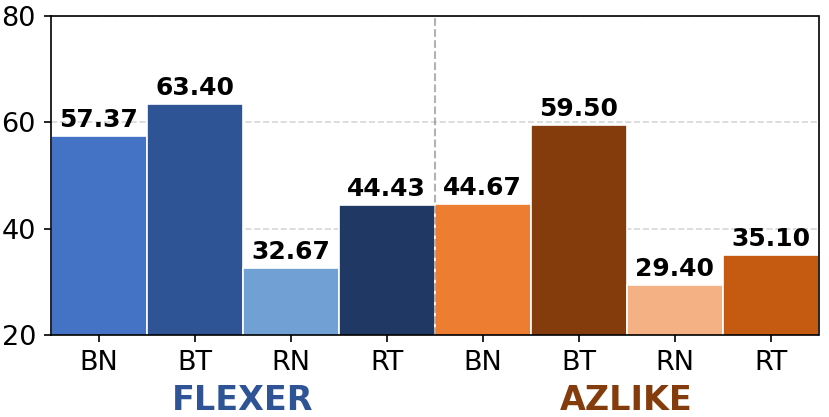}
        \caption{Goals completed}
        \label{fig:movingnumbers-easy-goals}
    \end{subfigure}
    \hfill
    \begin{subfigure}{0.48\textwidth}
        \includegraphics[width=\textwidth]{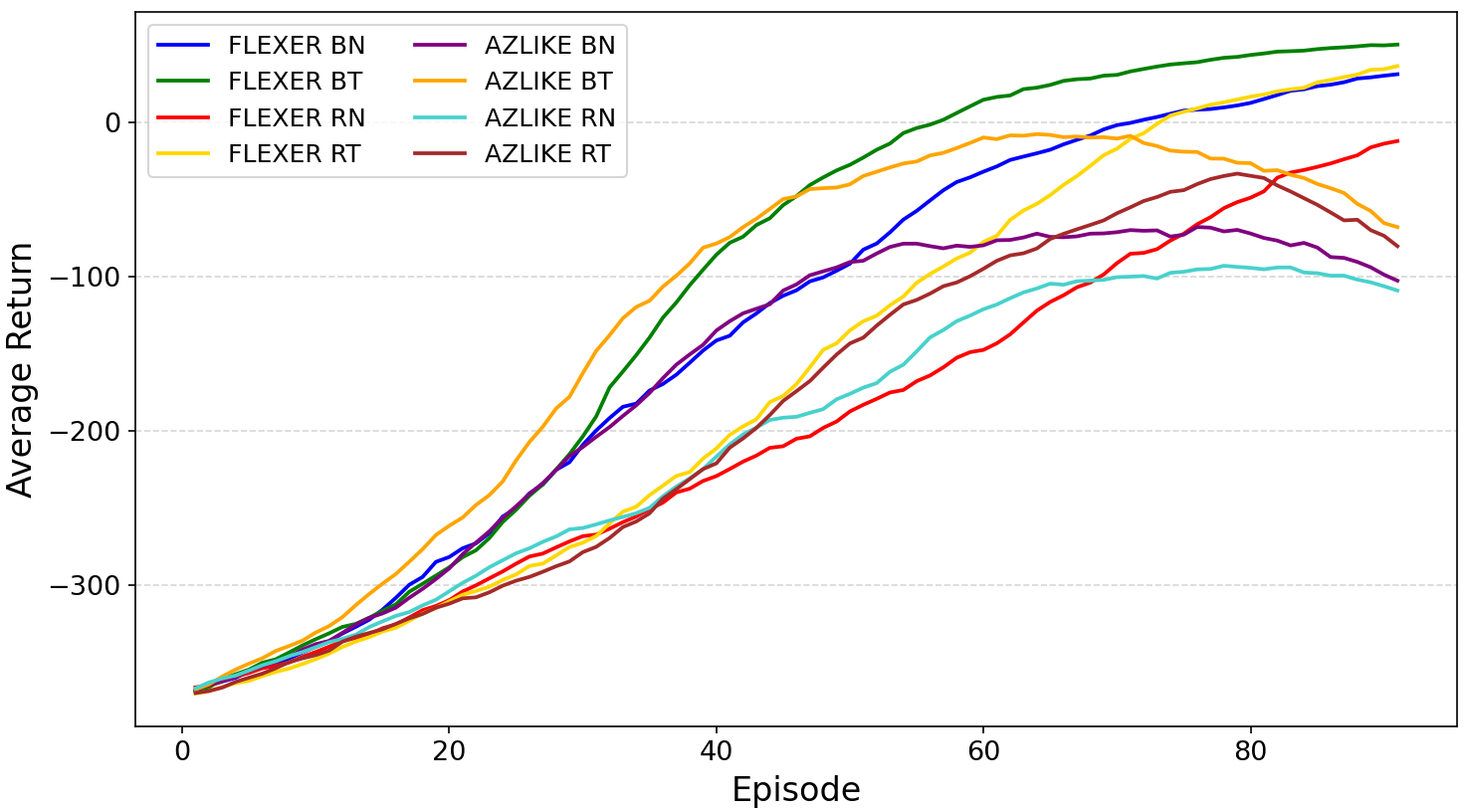}
        \caption{Return}
        \label{fig:movingnumbers-easy-returns}
    \end{subfigure}
    
    \begin{subfigure}{0.48\textwidth}
    \vspace{4mm}
        \includegraphics[width=\textwidth]{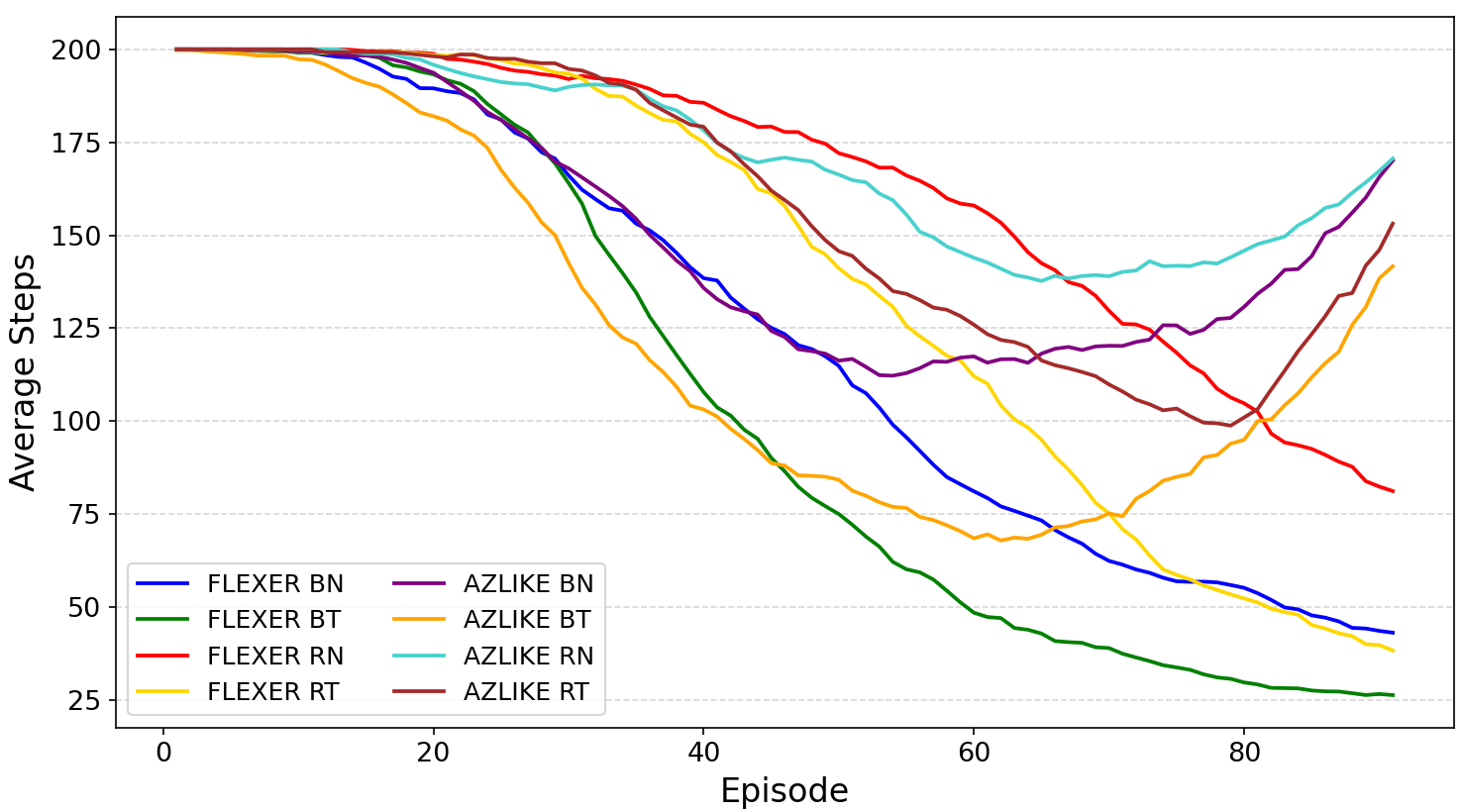}
        \caption{Steps in episode}
        \label{fig:movingnumbers-easy-steps}
    \end{subfigure}
    \hfill
    \begin{subfigure}{0.48\textwidth}
        \includegraphics[width=\textwidth]{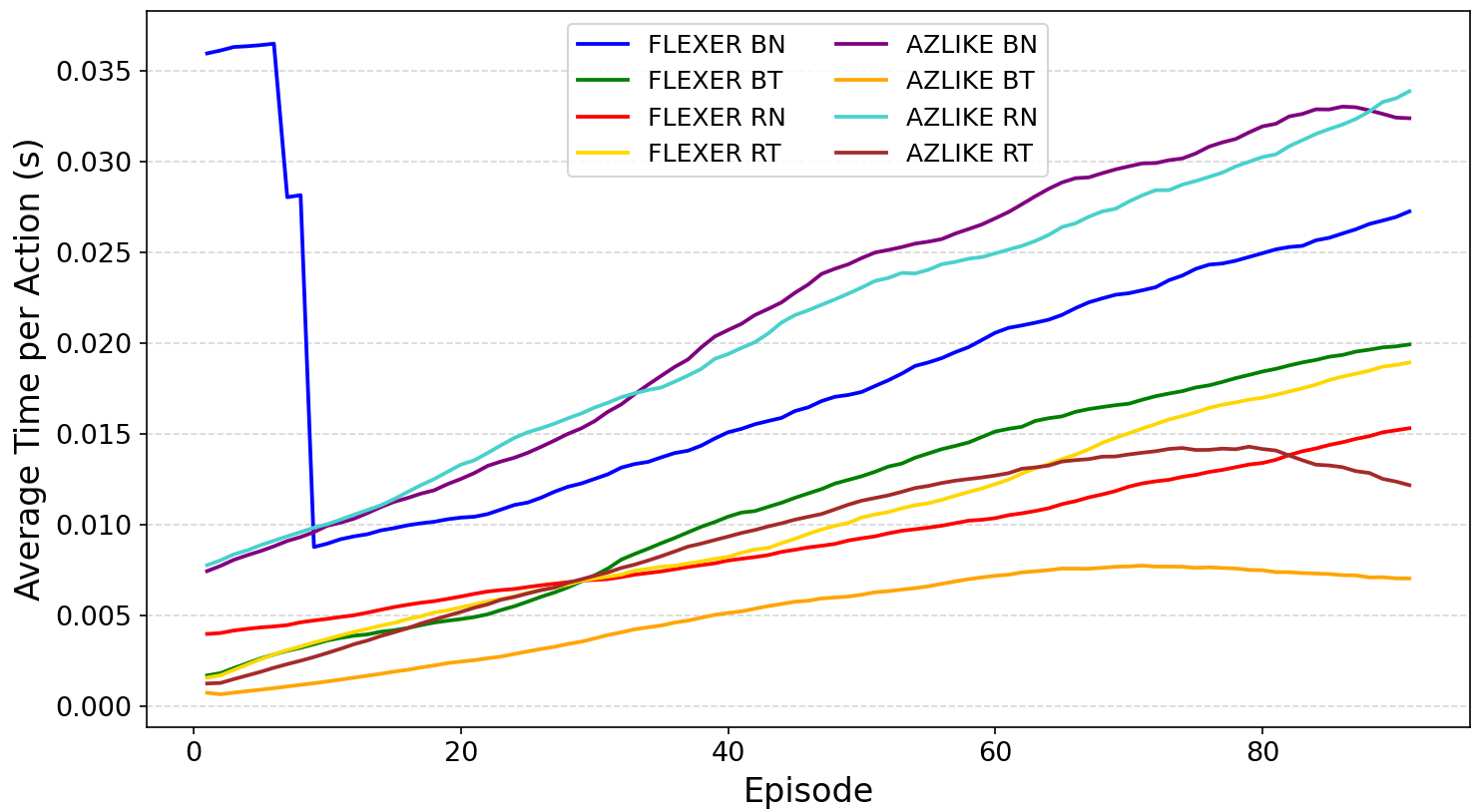}
        \caption{Time per step}
        \label{fig:movingnumbers-easy-times}
    \end{subfigure}
    \caption{Results of experiments running Flexer and AZLike on the MovingNumbers[5,2] environment.}
    \label{fig:movingnumbers-easy}
\end{figure*}

For MovingNumbers[5, 2], each environment instance is run 100 times.
Figure \ref{fig:movingnumbers-easy-goals} shows the number of goals achieved.
Figure \ref{fig:movingnumbers-easy-returns} shows the returns.
Figure \ref{fig:movingnumbers-easy-steps} shows the number of steps till goal achievement.
Figure \ref{fig:movingnumbers-easy-times} shows the total time per action/step.

\subsection{Flexer and AZLike on Harder Environs}

In variants using rollout, rollout depth is maximally 10. Hence, in variants using bootstrapping, MCTS tree depth is 10 more than rollout variants. For all experiments in this subsection, maximum MCTS tree depth is 15 for algorithms using rollout and 25 for bootstrapping. Maximum MCTS iterations is 50 for Flexer and 25 for AZLike.

\begin{figure*}[htb!]
    \centering
    \begin{subfigure}{0.4\textwidth}   
        \centering
\includegraphics[scale=0.6]{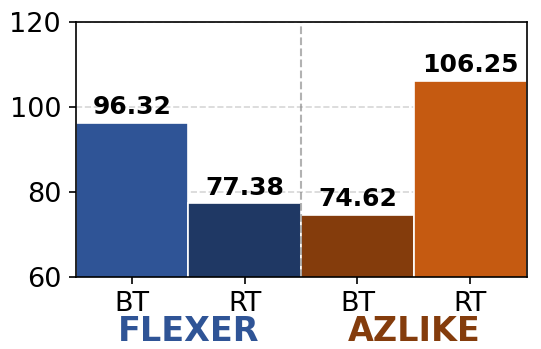}
        \caption{Goals completed}
        \label{fig:simplegrid-hard-goals}
    \end{subfigure}
    \hfill
    \begin{subfigure}{0.48\textwidth}
        \includegraphics[width=\textwidth]{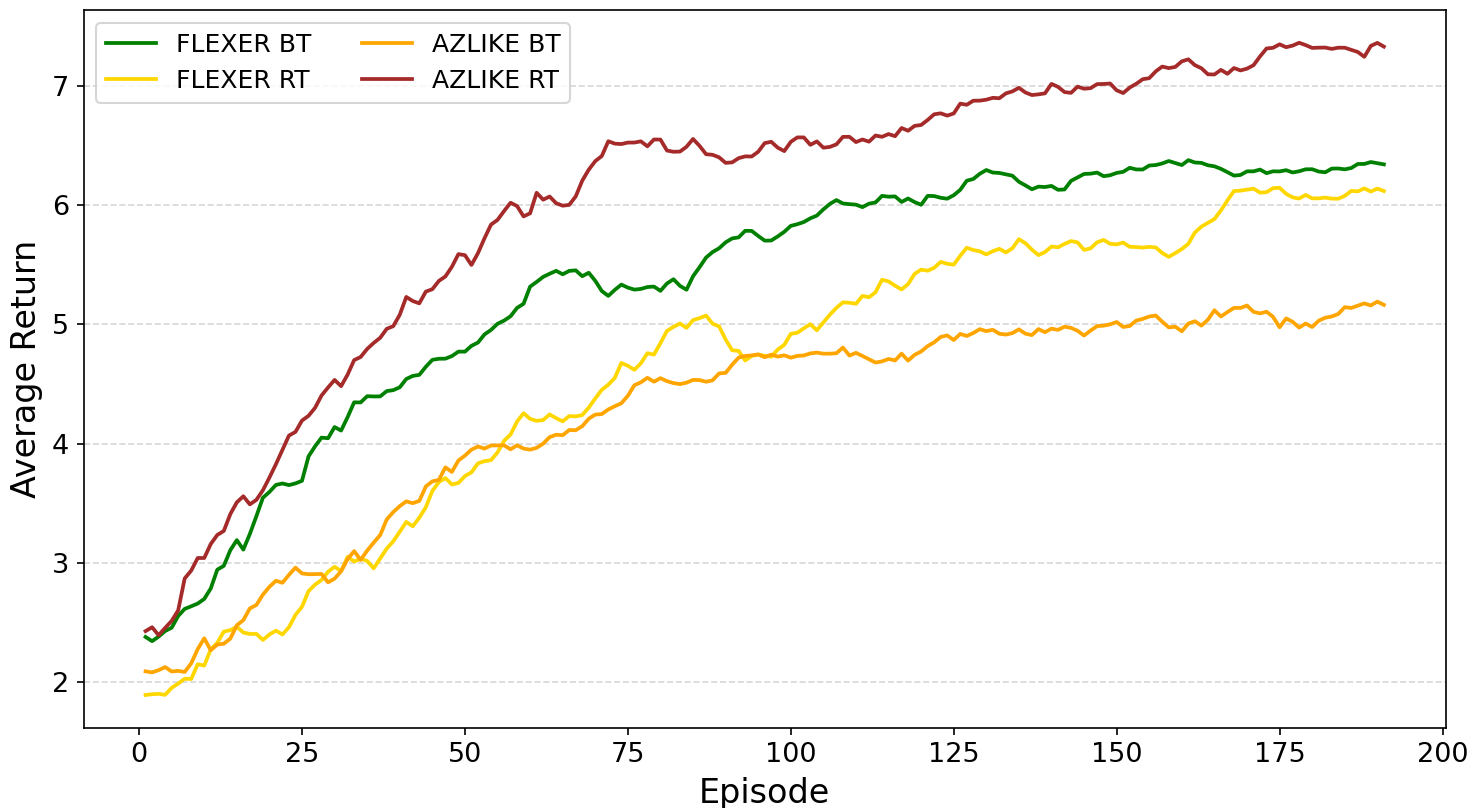}
        \caption{Return}
        \label{fig:simplegrid-hard-returns}
    \end{subfigure}
    
    \begin{subfigure}{0.48\textwidth}
    \vspace{4mm}
        \includegraphics[width=\textwidth]{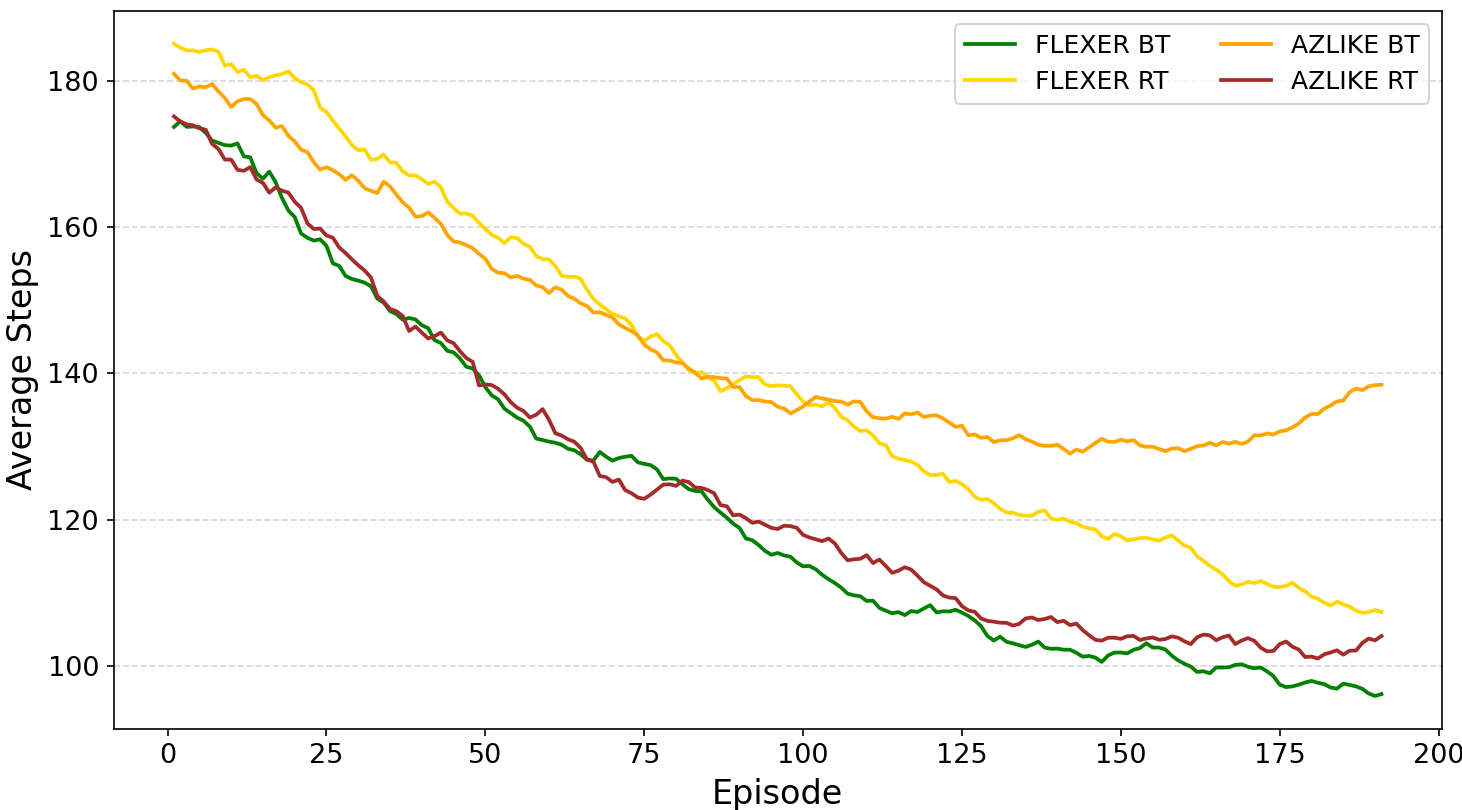}
        \caption{Steps in episode}
        \label{fig:simplegrid-hard-steps}
    \end{subfigure}
    \hfill
    \begin{subfigure}{0.48\textwidth}
        \includegraphics[width=\textwidth]{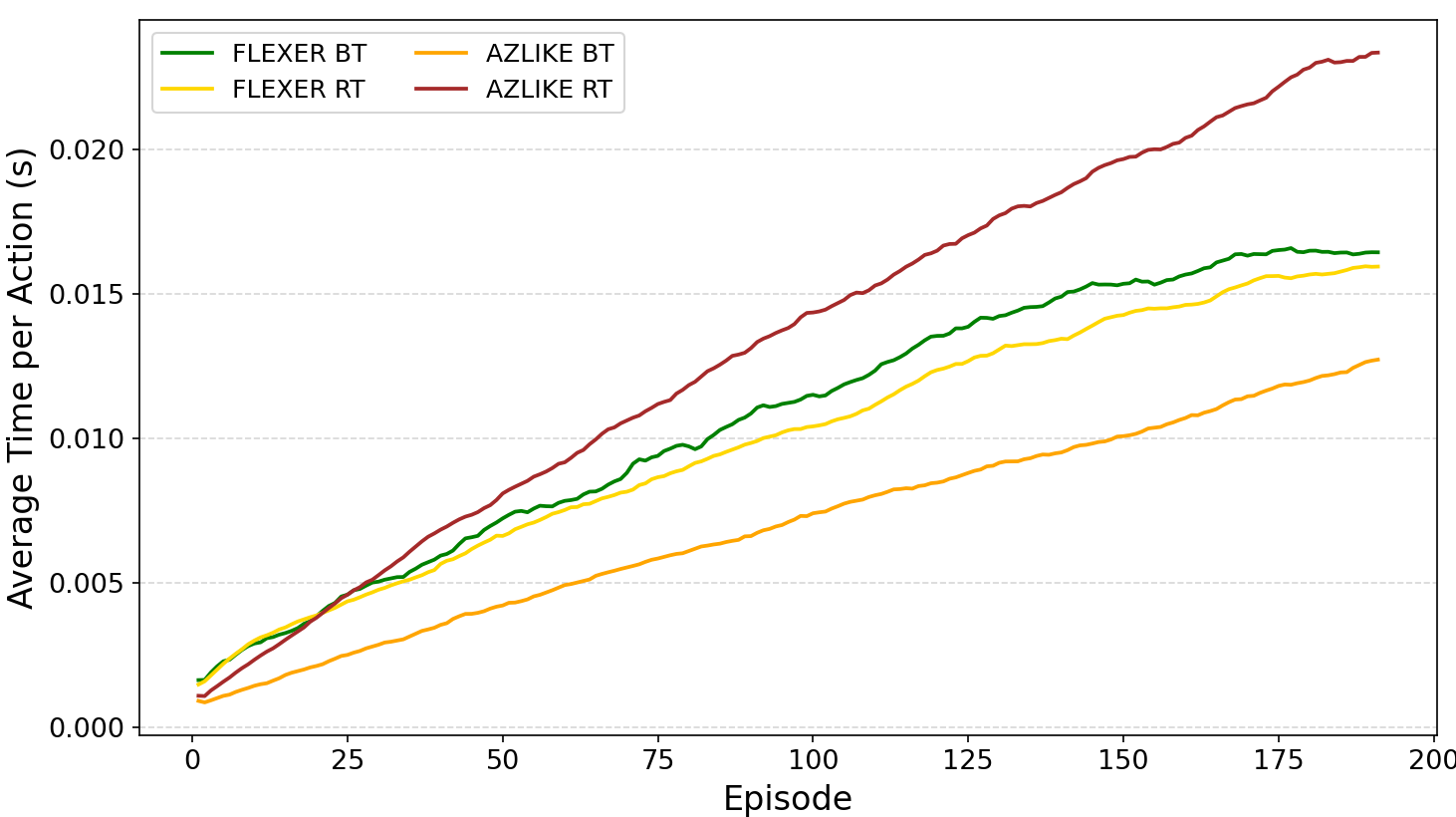}
        \caption{Time per step}
        \label{fig:simplegrid-hard-times}
    \end{subfigure}
    \caption{Results of experiments running Flexer and AZLike on the SimpleGrid[15,0.3] environment.}
    \label{fig:simplegrid-hard}
\end{figure*}

For SimpleGrid[15, 0.3], there are 60 environment instances, and learning occurs over 200 instances.
Figure \ref{fig:simplegrid-hard-goals} shows the number of goals achieved.
Figure \ref{fig:simplegrid-hard-returns} shows the returns.
Figure \ref{fig:simplegrid-hard-steps} shows the number of steps till goal achievement.
Figure \ref{fig:simplegrid-hard-times} shows the total time per action/step.

\begin{figure*}[htb!]
    \centering
    \begin{subfigure}{0.4\textwidth}
        \centering
    \includegraphics[scale=0.6]{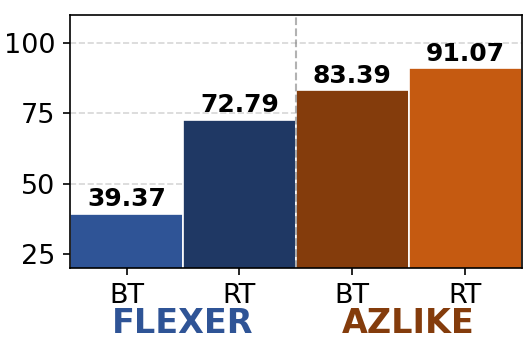}
        \caption{Goals completed}
        \label{fig:blocksworld-hard-goals}
    \end{subfigure}
    \hfill
    \begin{subfigure}{0.48\textwidth}
        \includegraphics[width=\textwidth]{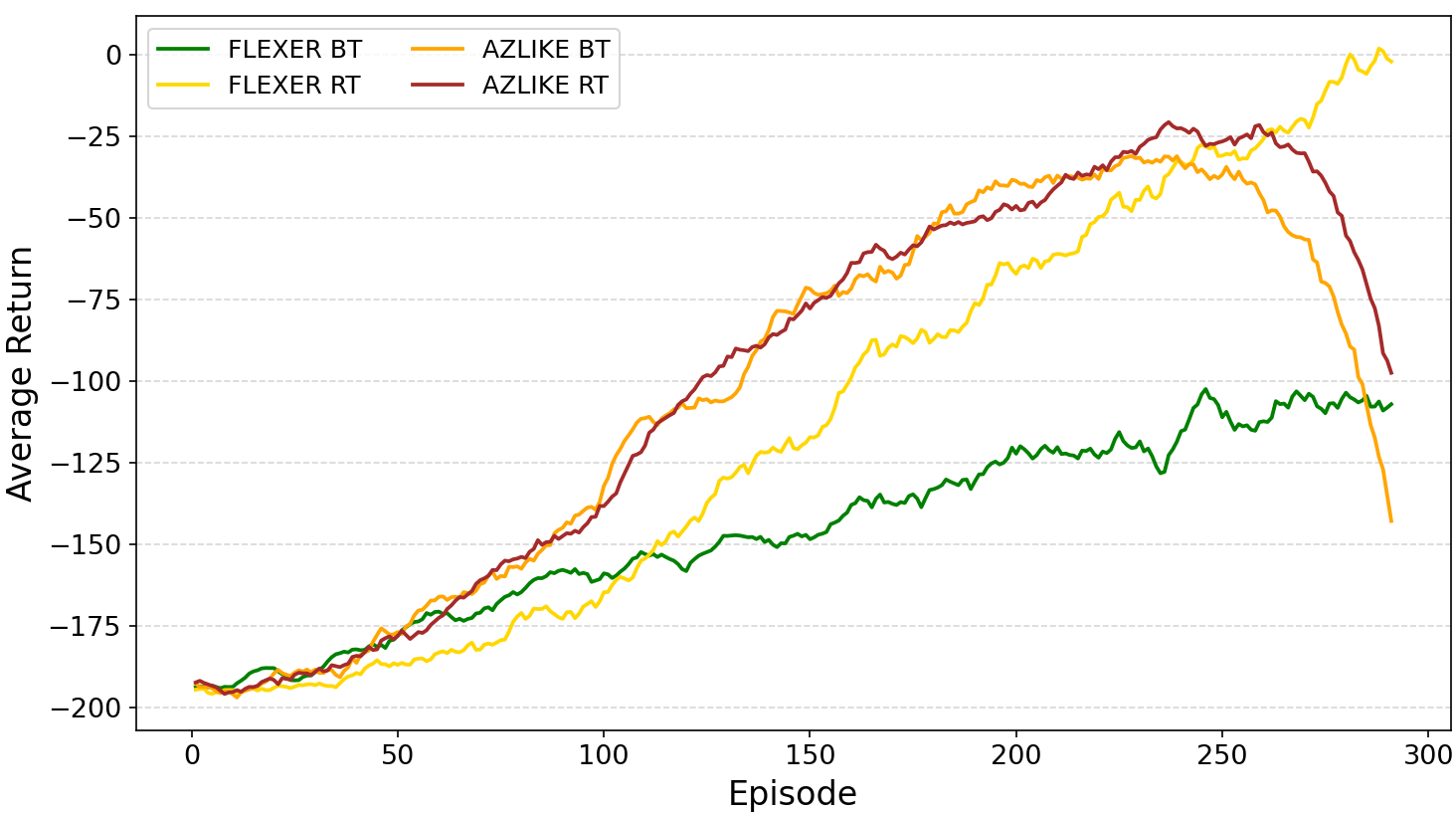}
        \caption{Return}
        \label{fig:blocksworld-hard-returns}
    \end{subfigure}
    
    \begin{subfigure}{0.48\textwidth}
    \vspace{4mm}
\includegraphics[width=\textwidth]{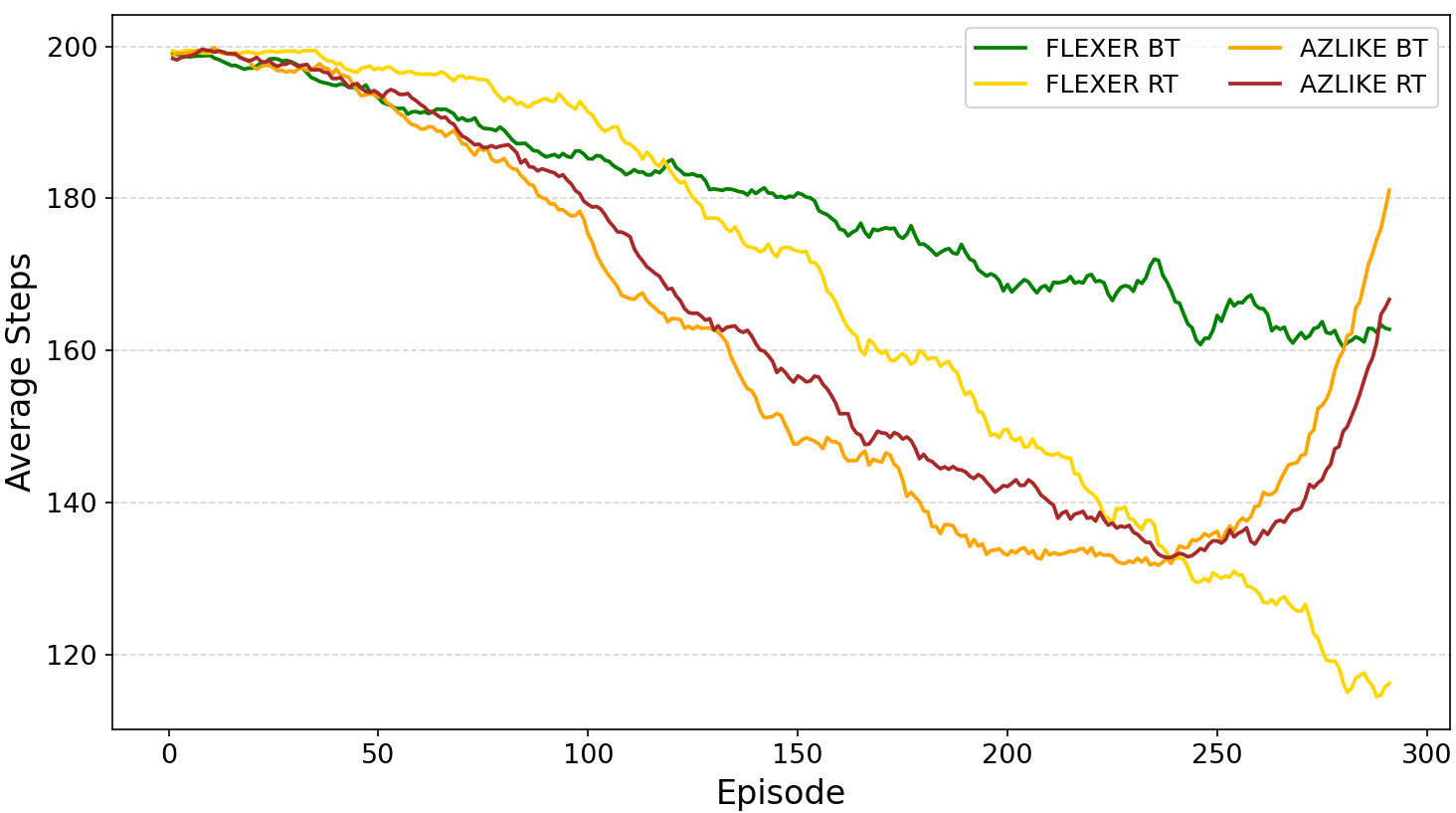}
        \caption{Steps in episode}
        \label{fig:blocksworld-hard-steps}
    \end{subfigure}
    \hfill
    \begin{subfigure}{0.48\textwidth}
        \includegraphics[width=\textwidth]{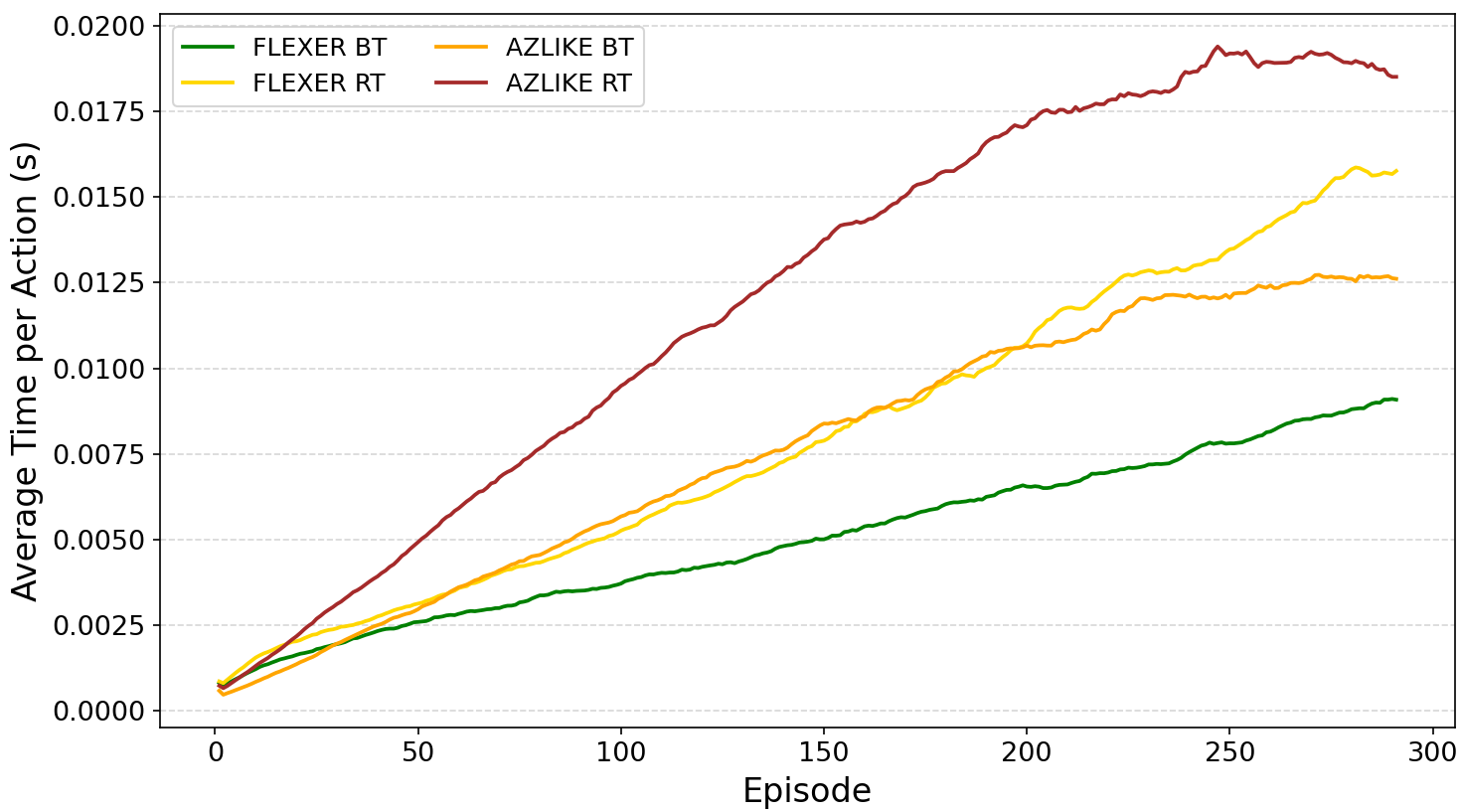}
        \caption{Time per step}
        \label{fig:blocksworld-hard-times}
    \end{subfigure}
    \caption{Results of experiments running Flexer and AZLike on the BlocksWorld[4,4] environment.}
    \label{fig:blocksworld-hard}
\end{figure*}

For BlocksWorld[4,4], there are 60 environment instances, and learning occurs over 200 instances.
Figure \ref{fig:blocksworld-hard-goals} shows the number of goals achieved.
Figure \ref{fig:blocksworld-hard-returns} shows the returns.
Figure \ref{fig:blocksworld-hard-steps} shows the number of steps till goal achievement.
Figure \ref{fig:blocksworld-hard-times} shows the total time per action/step.

\begin{figure*}[htb!]
    \centering
    \begin{subfigure}{0.4\textwidth}
        \centering
    \includegraphics[scale=0.6]{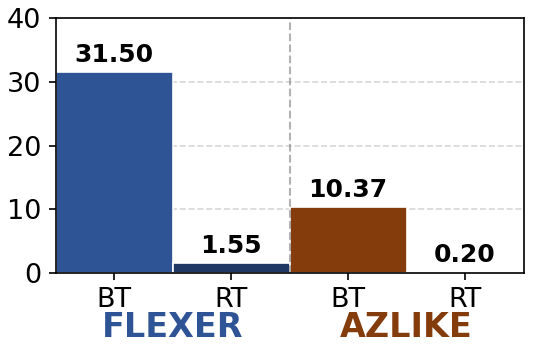}
        \caption{Goals completed}
        \label{fig:movingnumbers-hard-goals}
    \end{subfigure}
    \hfill
    \begin{subfigure}{0.48\textwidth}
        \includegraphics[width=\textwidth]{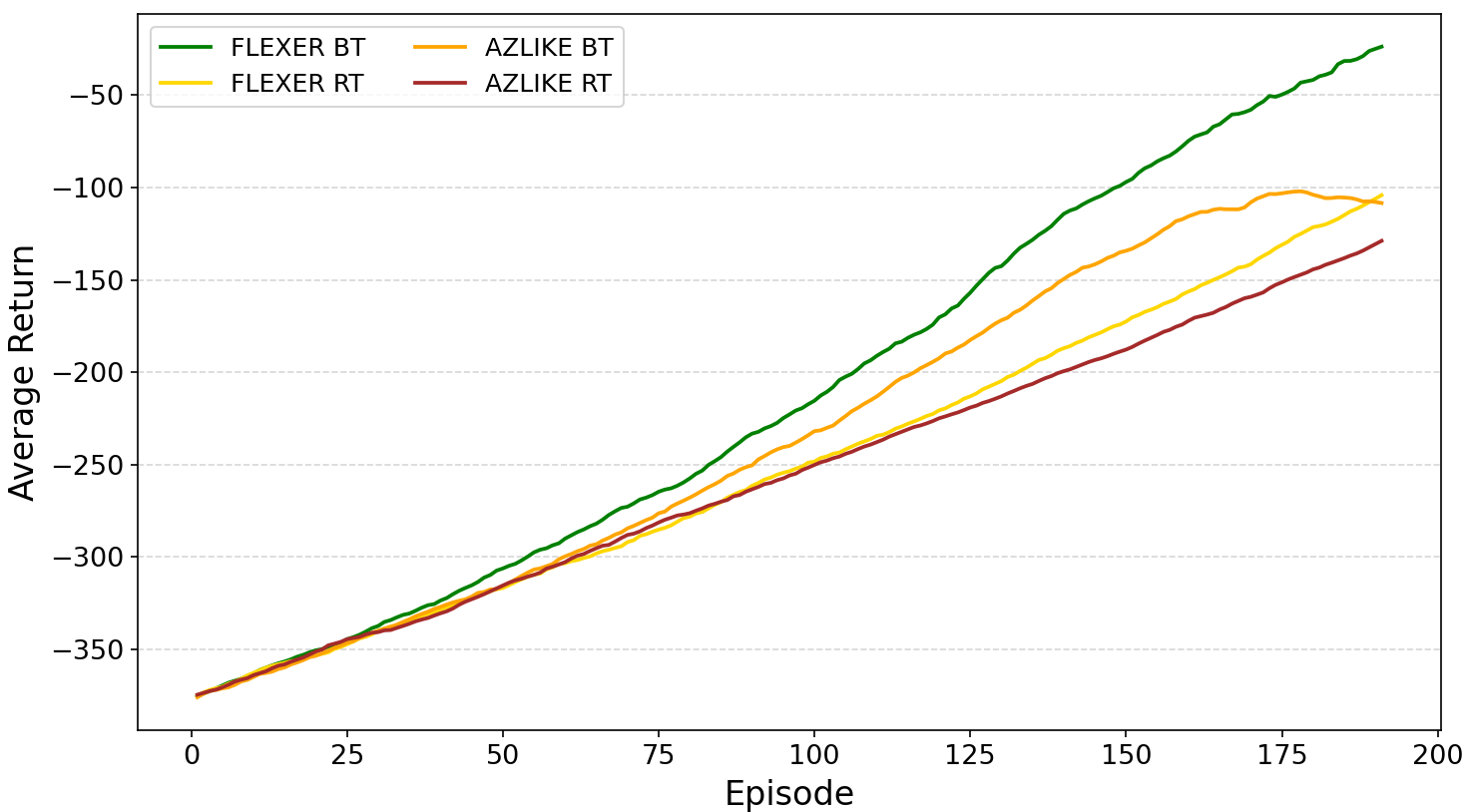}
        \caption{Return}
        \label{fig:movingnumbers-hard-returns}
    \end{subfigure}
    
    \begin{subfigure}{0.48\textwidth}
    \vspace{4mm}
        \includegraphics[width=\textwidth]{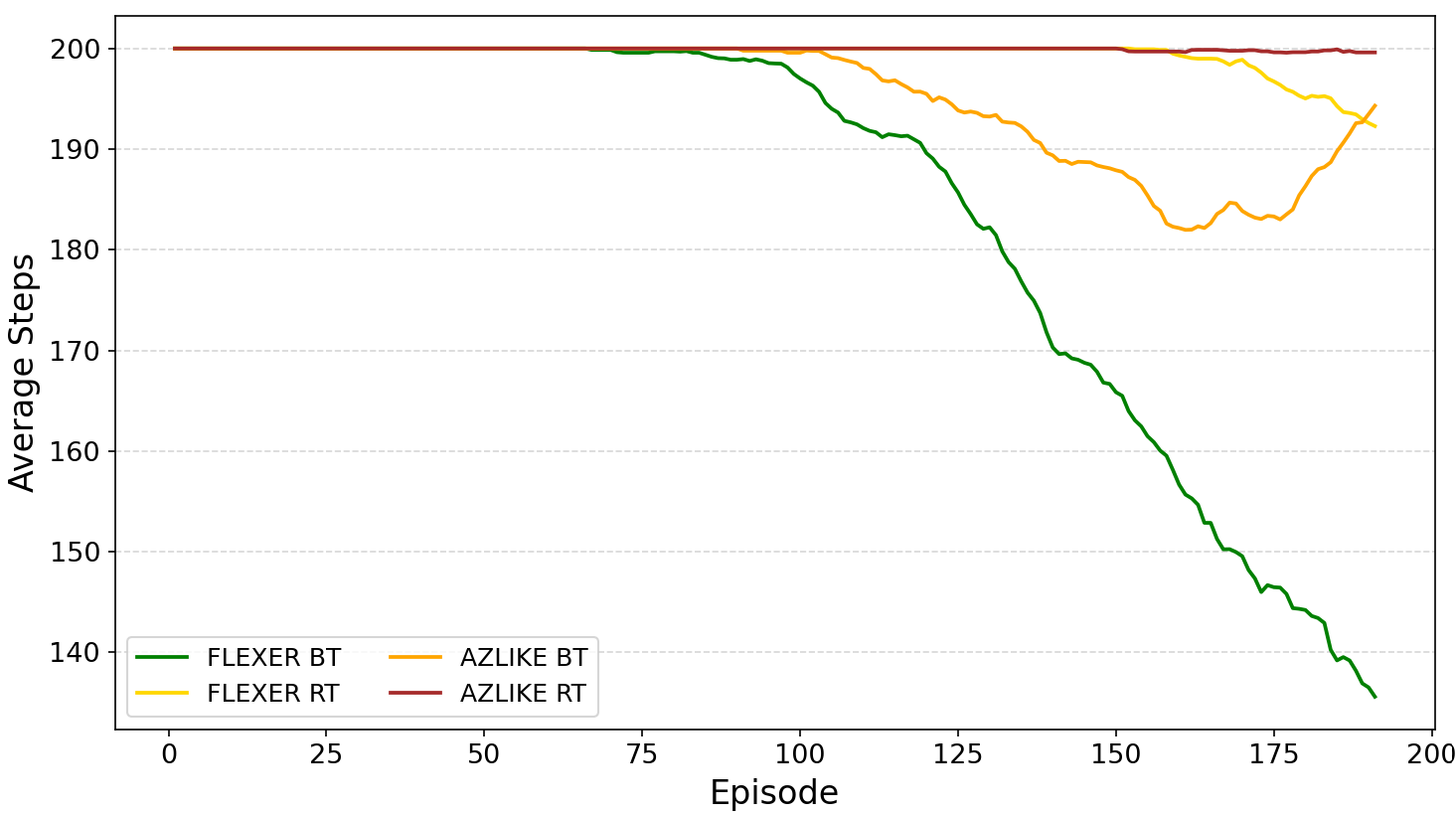}
        \caption{Steps in episode}
        \label{fig:movingnumbers-hard-steps}
    \end{subfigure}
    \hfill
    \begin{subfigure}{0.48\textwidth}
        \includegraphics[width=\textwidth]{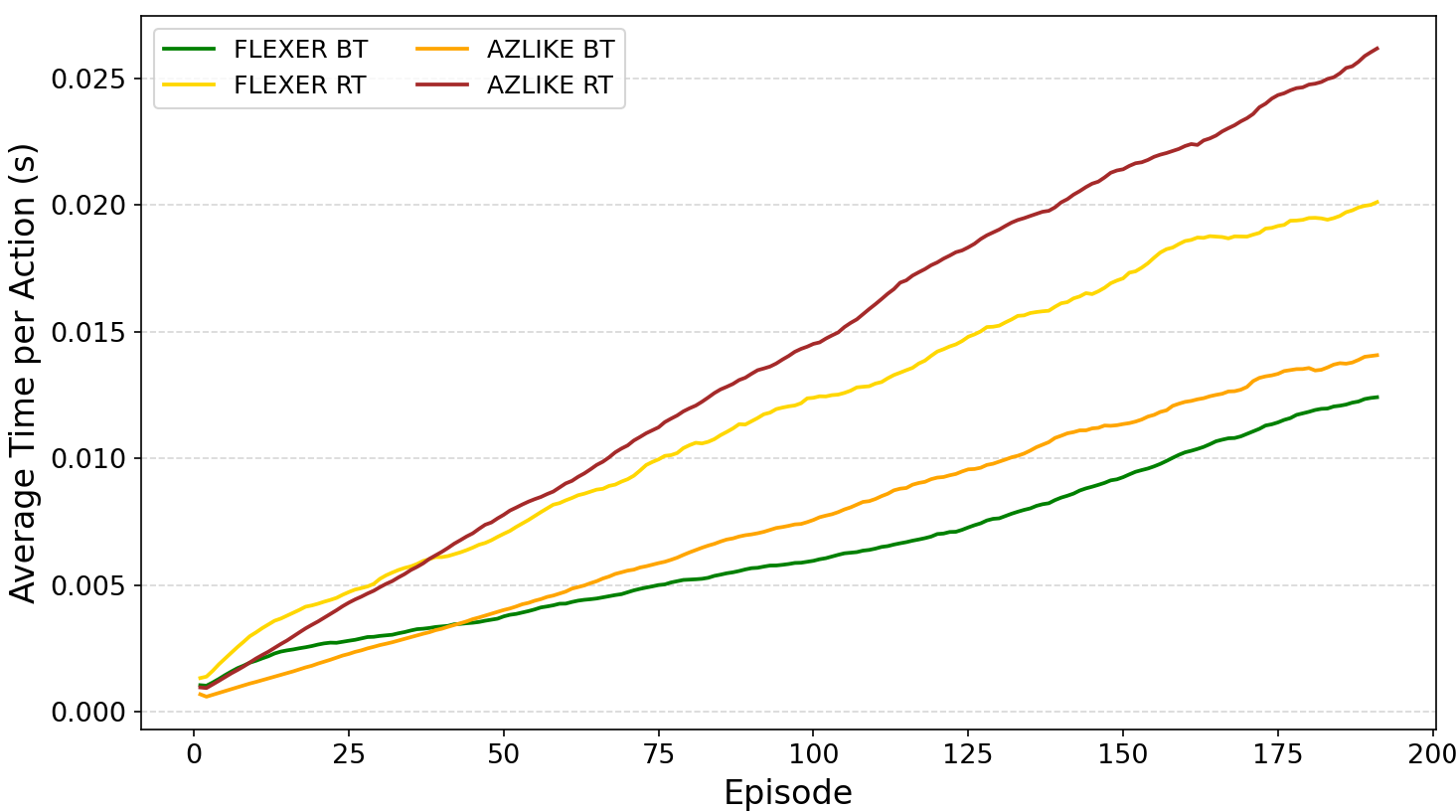}
        \caption{Time per step}
        \label{fig:movingnumbers-hard-times}
    \end{subfigure}
    \caption{Results of experiments running Flexer and AZLike on the MovingNumbers[10,3] environment.}
    \label{fig:movingnumbers-hard}
\end{figure*}

For MovingNumbers[10, 3], there are 60 environment instances, and learning occurs over 200 instances.
Figure \ref{fig:movingnumbers-hard-goals} shows the number of goals achieved.
Figure \ref{fig:movingnumbers-hard-returns} shows the returns.
Figure \ref{fig:movingnumbers-hard-steps} shows the number of steps till goal achievement.
Figure \ref{fig:movingnumbers-hard-times} shows the total time per action/step.

\subsection{Flexer with Boosted Value Function}

In the following experiments, we investigate whether boosting the value function from values found at internal MCTS tree nodes actually improves performance and whether it is worth the cost in compute. Our approach is explained in \S~\ref{sec:boosting}.
Boosting with the maximum q-value is abbreviated as MQ and boosing with weighted q-values is abbreviated as WQ.

\begin{figure*}[htb!]
    \centering
    \begin{subfigure}{0.48\textwidth}
        \includegraphics[width=\textwidth]{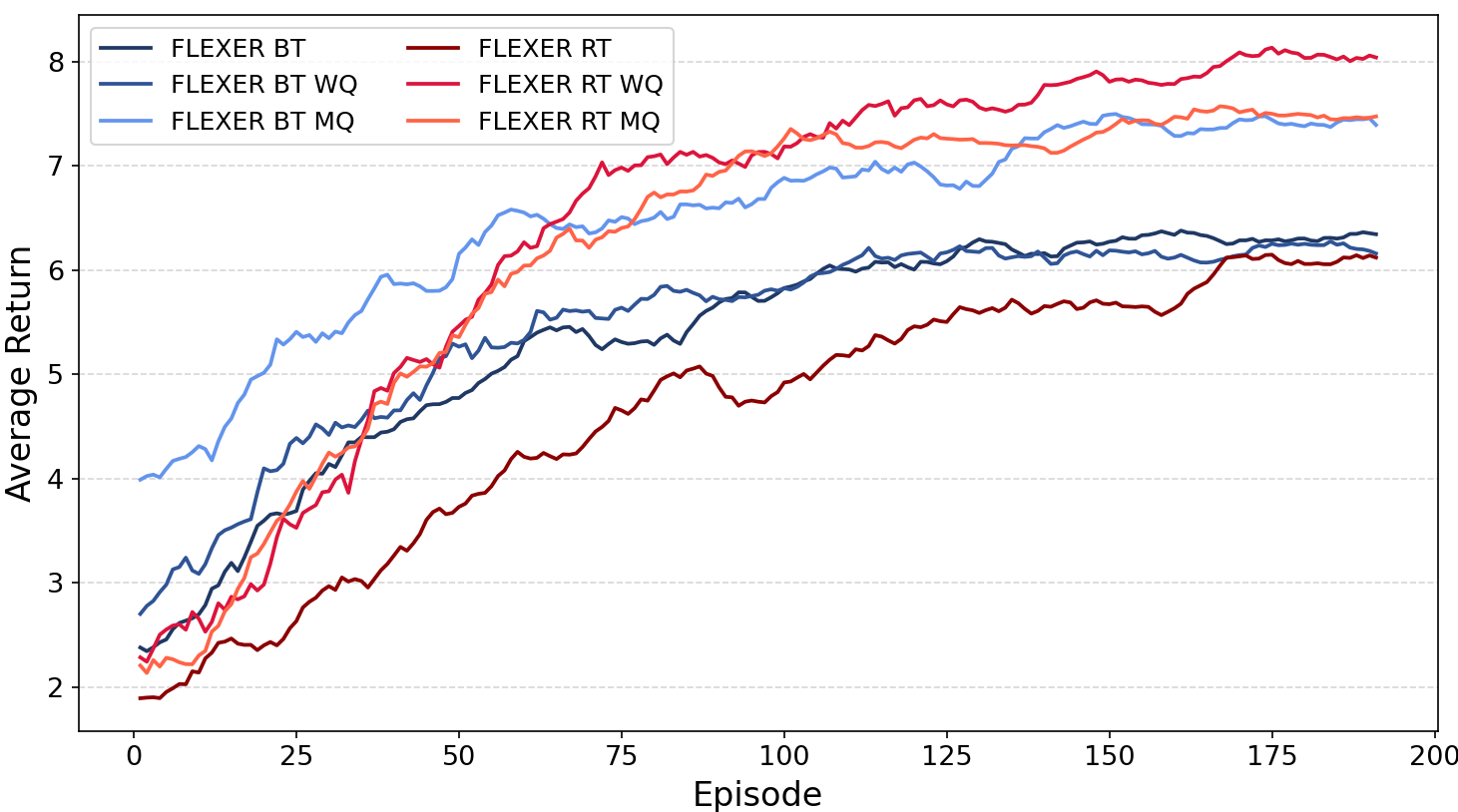}
        \caption{Return}
        \label{fig:simplegrid-boost-returns}
    \end{subfigure}
    \begin{subfigure}{0.48\textwidth}
        \includegraphics[width=\textwidth]{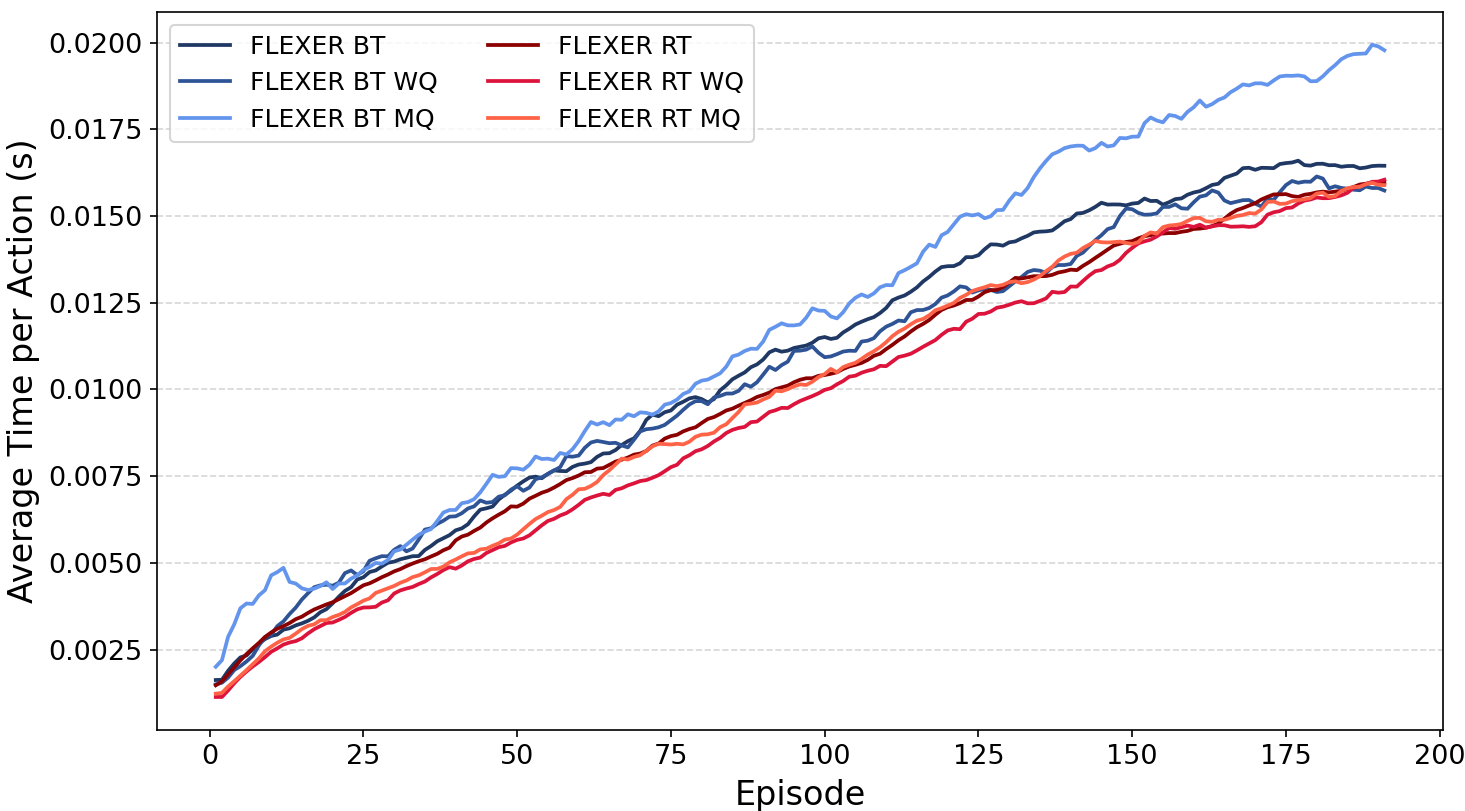}
        \caption{Time per step}
        \label{fig:simplegrid-boost-times}
    \end{subfigure}
    \caption{Results of experiments running best Flexers, with and without value boosting on the SimpleGrid[15,0.3] environment.}
    \label{fig:simplegrid-boosting}
\end{figure*}

\begin{figure*}[htb!]
    \centering
    \begin{subfigure}{0.48\textwidth}
        \includegraphics[width=\textwidth]{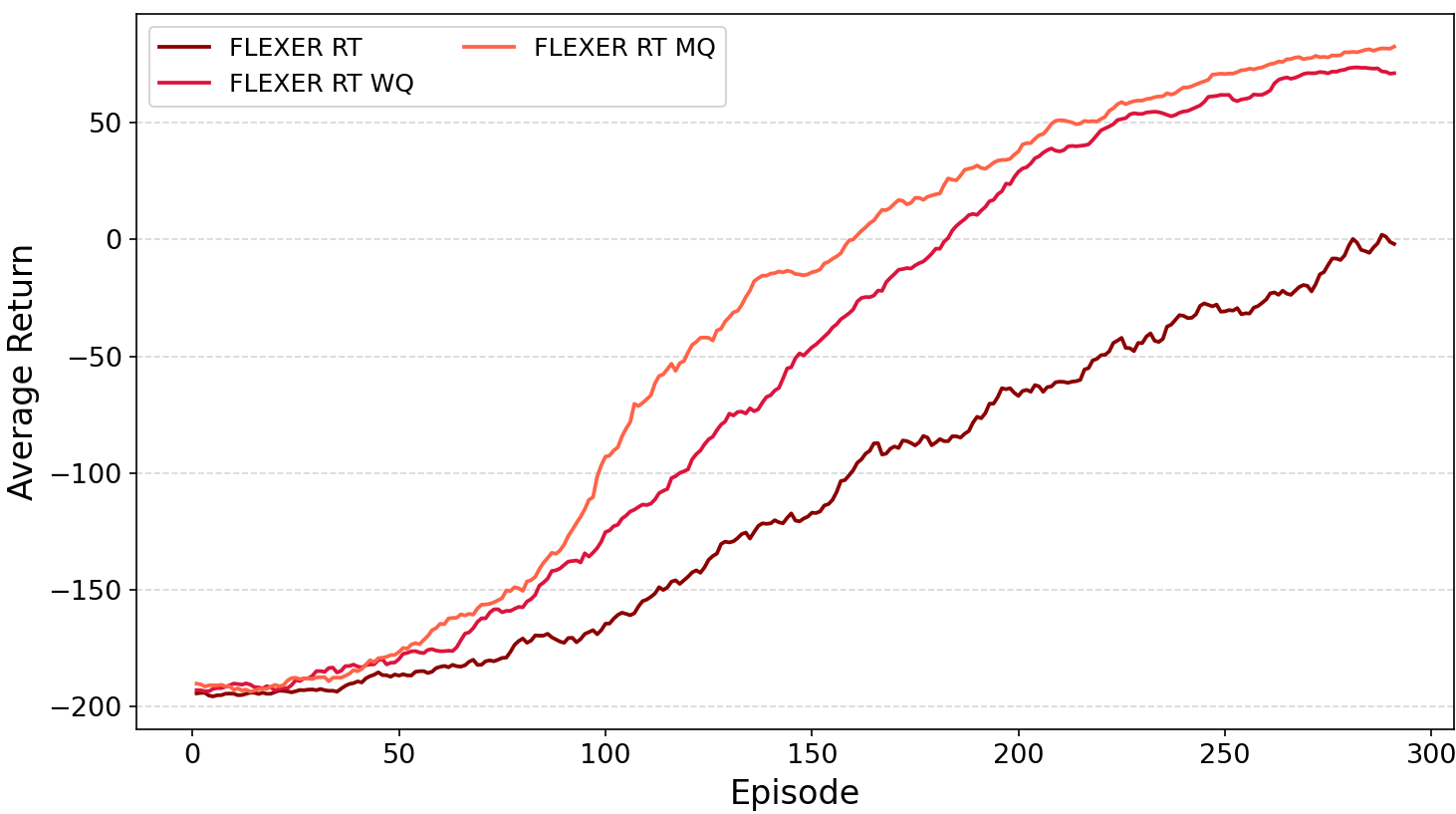}
        \caption{Return}
        \label{fig:blocksworld-boost-returns}
    \end{subfigure}
    \begin{subfigure}{0.48\textwidth}
        \includegraphics[width=\textwidth]{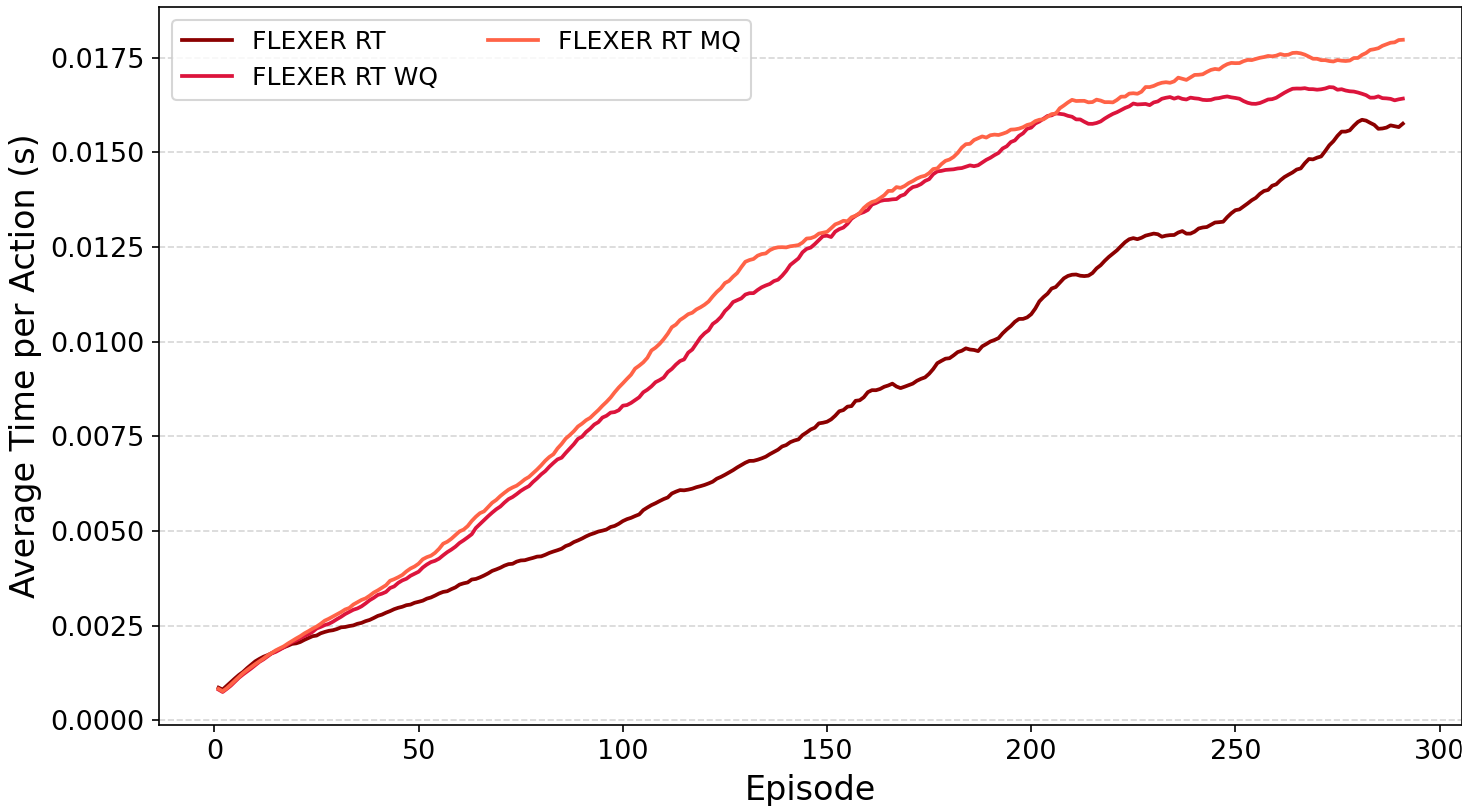}
        \caption{Time per step}
        \label{fig:blocksworld-boost-times}
    \end{subfigure}
    \caption{Results of experiments running best Flexers, with and without value boosting on the BlocksWorld[4,4] environment.}
    \label{fig:blocksworld-boosting}
\end{figure*}

\begin{figure*}[htb!]
    \centering
    \begin{subfigure}{0.48\textwidth}
        \includegraphics[width=\textwidth]{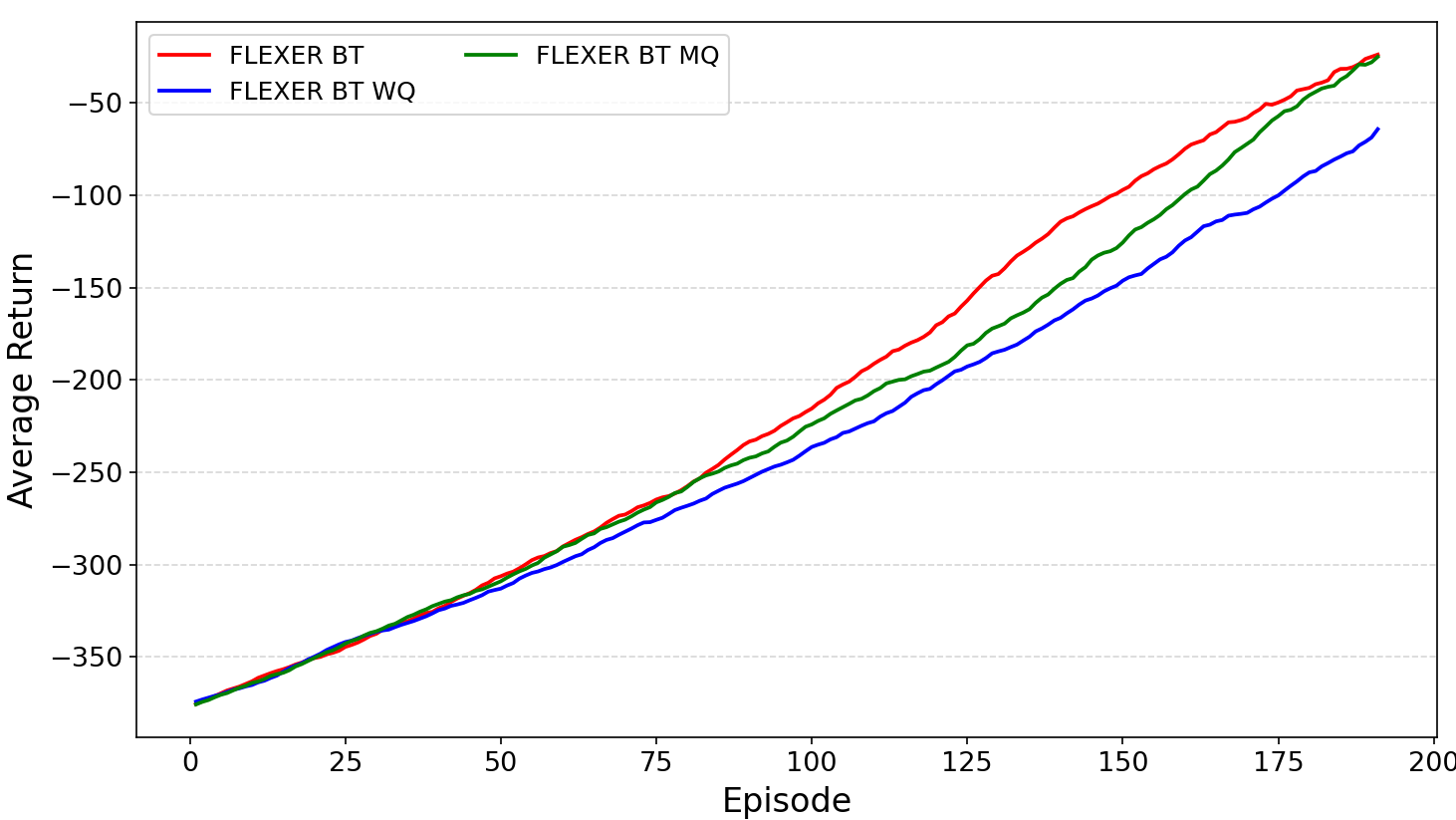}
        \caption{Return}
        \label{fig:movingnumbers-boost-returns}
    \end{subfigure}
    \begin{subfigure}{0.48\textwidth}
        \includegraphics[width=\textwidth]{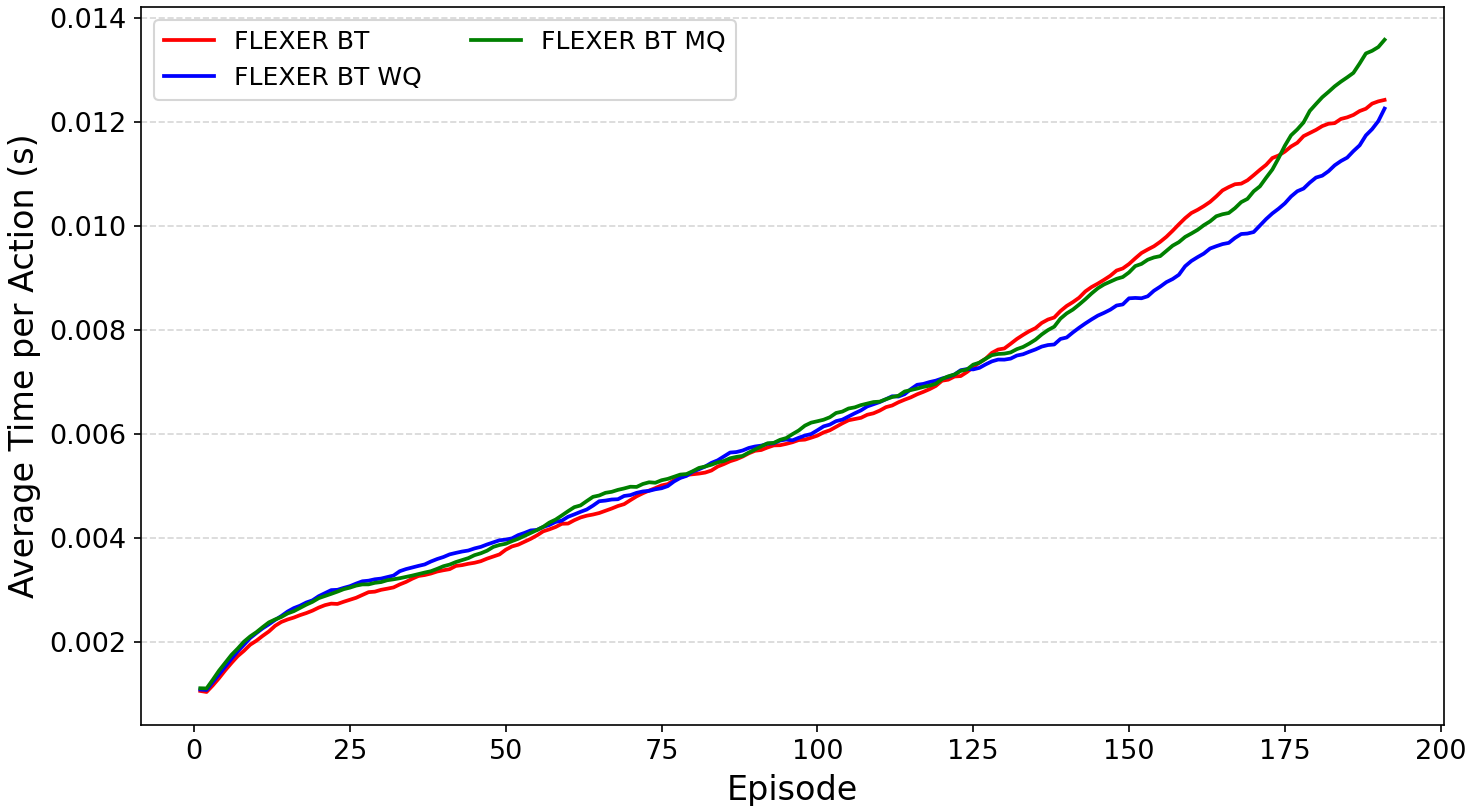}
        \caption{Time per step}
        \label{fig:movingnumbers-boost-times}
    \end{subfigure}
    \caption{Results of experiments running best Flexers, with and without value boosting on the MovingNumbers[10,3] environment.}
    \label{fig:movingnumbers-boosting}
\end{figure*}

Figure \ref{fig:simplegrid-boost-returns} shows the returns and
Figure \ref{fig:simplegrid-boost-times} shows the total time per action/step for the hard SimpleGrid problem.
Figure \ref{fig:blocksworld-boost-returns} shows the returns and
Figure \ref{fig:blocksworld-boost-times} shows the total time per action/step for the hard BlocksWorld problem.
Figure \ref{fig:movingnumbers-boost-returns} shows the returns and
Figure \ref{fig:movingnumbers-boost-times} shows the total time per action/step for the hard MovingNumbers problem.

\subsection{Comparison with DQN and ADP}

\begin{figure*}[htb!]
    \centering
    \begin{subfigure}{0.4\textwidth}  
        \centering
\includegraphics[scale=0.5]{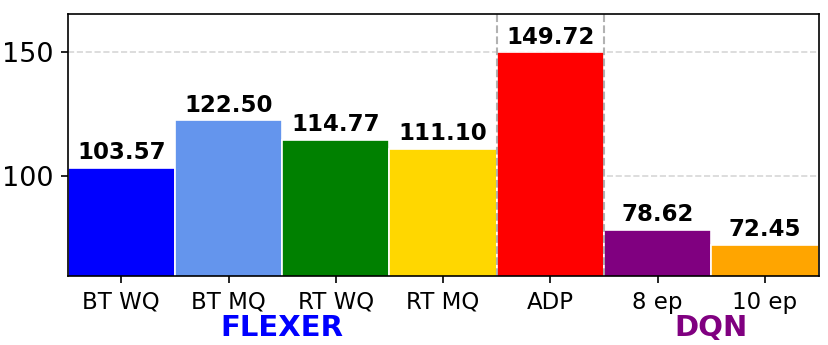}
        \caption{Goals completed}
        \label{fig:simplegrid-comp-goals}
    \end{subfigure}
    \hfill
    \begin{subfigure}{0.48\textwidth}
        \includegraphics[width=\textwidth]{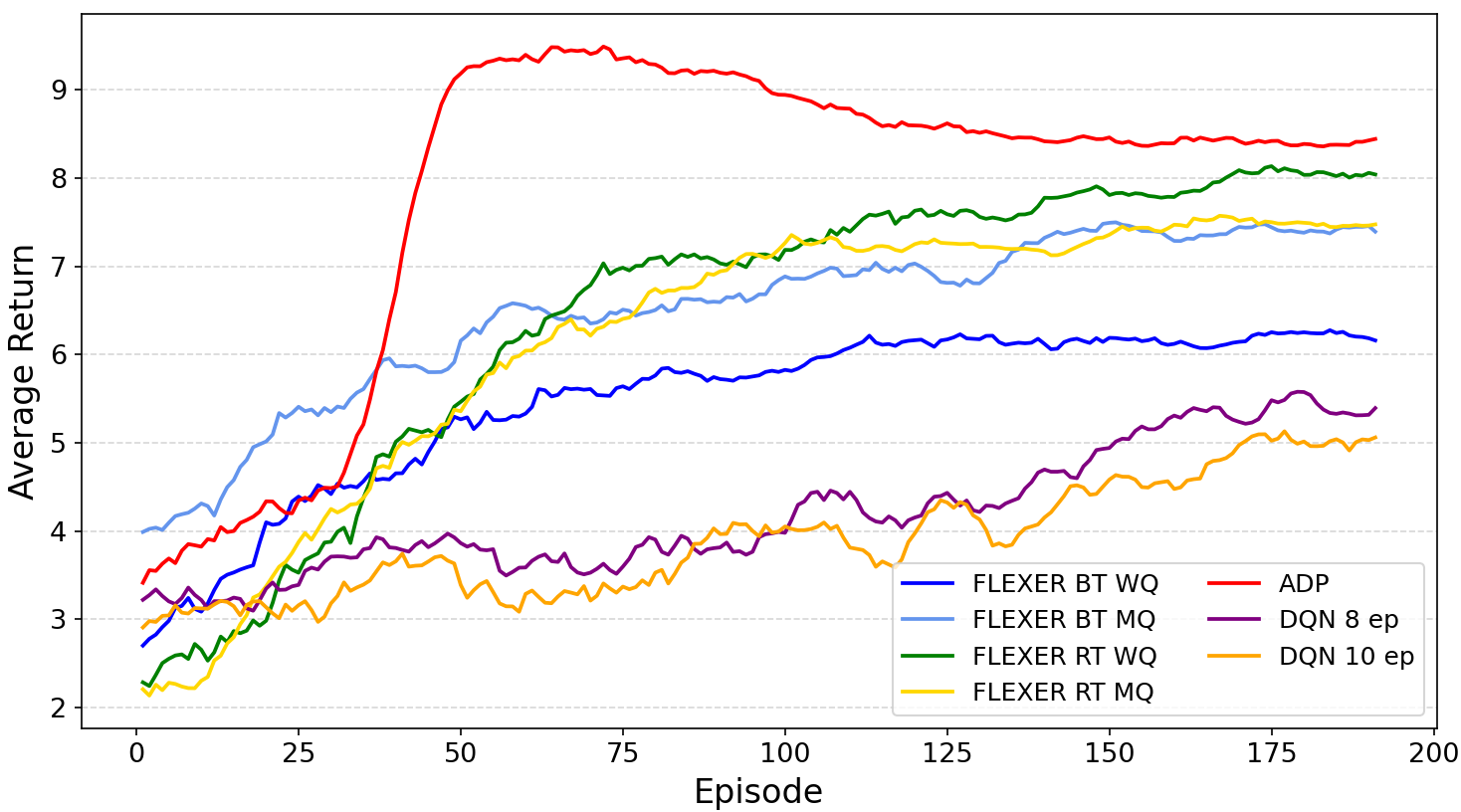}
        \caption{Return}
        \label{fig:simplegrid-comp-returns}
    \end{subfigure}
    
    \begin{subfigure}{0.48\textwidth}
    \vspace{4mm}
        \includegraphics[width=\textwidth]{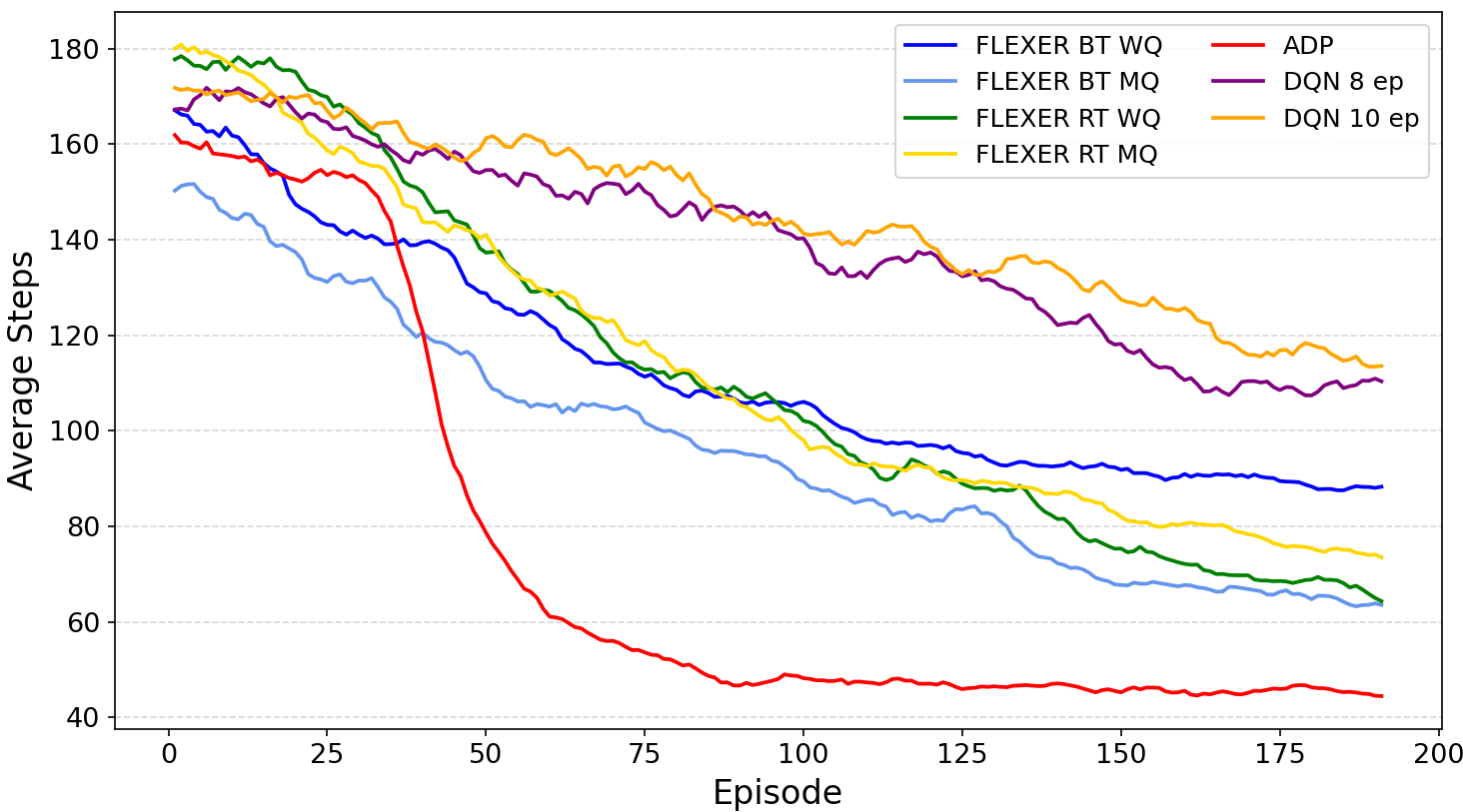}
        \caption{Steps in episode}
        \label{fig:simplegrid-comp-steps}
    \end{subfigure}
    \hfill
    \begin{subfigure}{0.48\textwidth}
        \includegraphics[width=\textwidth]{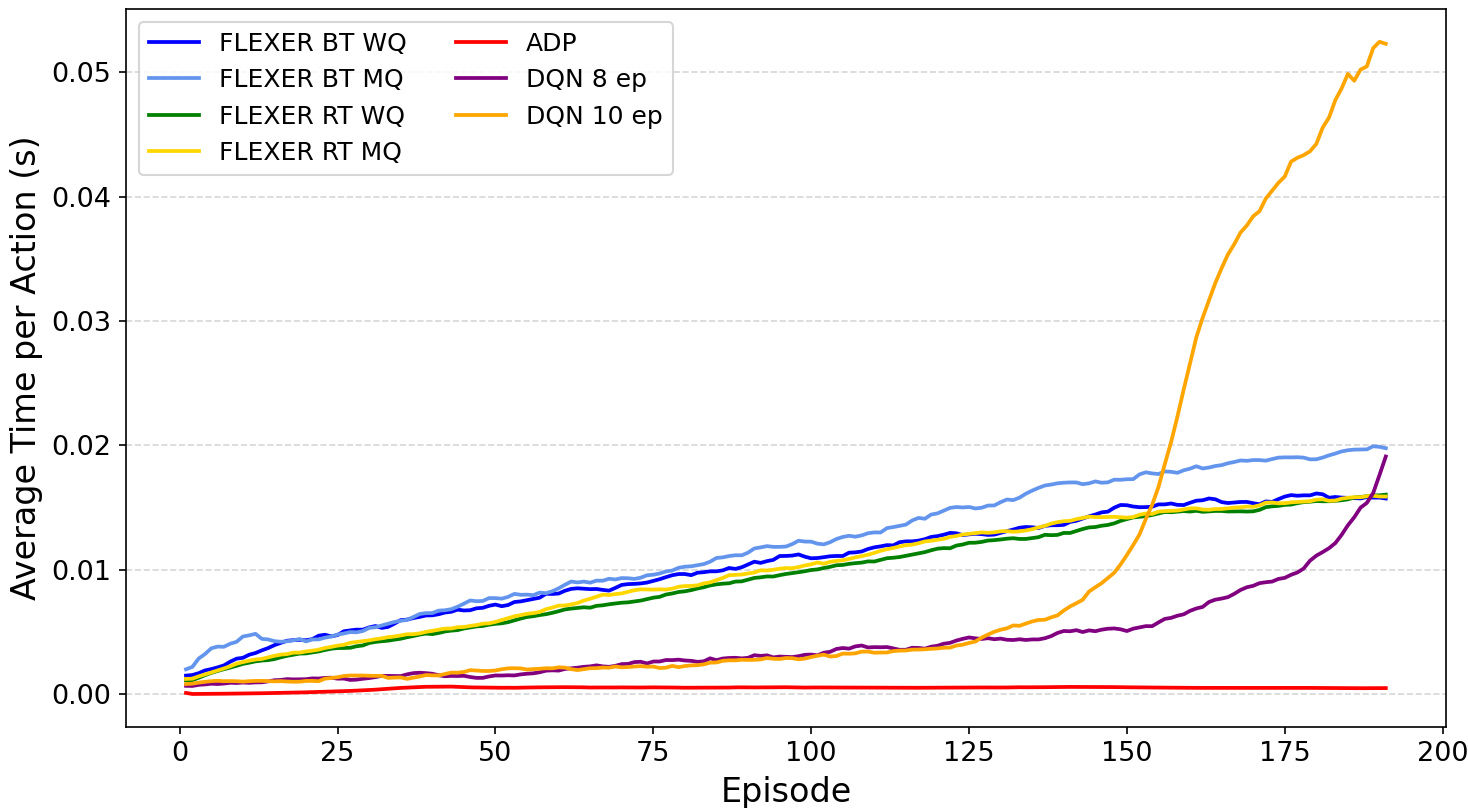}
        \caption{Time per step}
        \label{fig:simplegrid-comp-times}
    \end{subfigure}
    \caption{Results of experiments running best Flexers, DQN and ADP on the SimpleGrid[15,0.3] environment.}
    \label{fig:simplegrid-comp}
\end{figure*}

\begin{figure*}[htb!]
    \centering
    \begin{subfigure}{0.4\textwidth}  
        \centering
\includegraphics[scale=0.45]{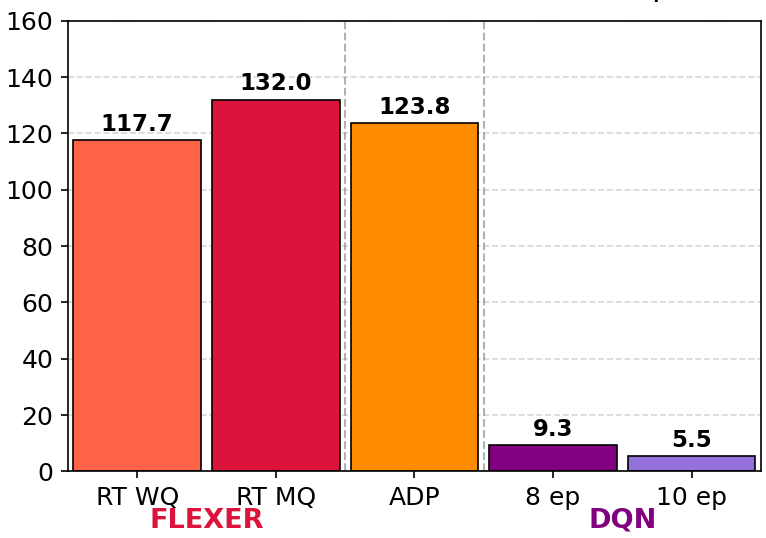}
        \caption{Goals completed}
        \label{fig:blocksworld-comp-goals}
    \end{subfigure}
    \hfill
    \begin{subfigure}{0.48\textwidth}
        \includegraphics[width=\textwidth]{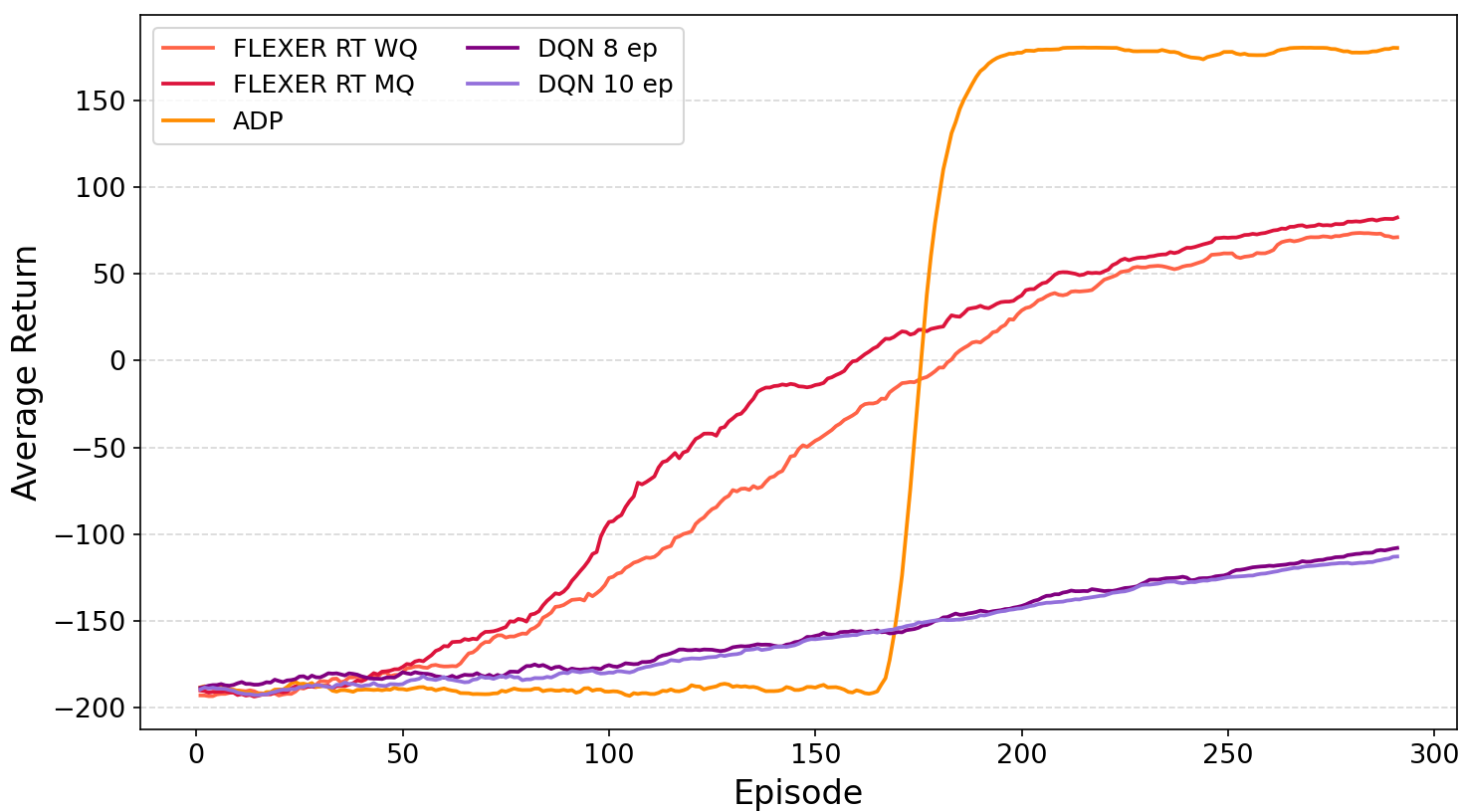}
        \caption{Return}
        \label{fig:blocksworld-comp-returns}
    \end{subfigure}
    
    \begin{subfigure}{0.48\textwidth}
    \vspace{4mm}
        \includegraphics[width=\textwidth]{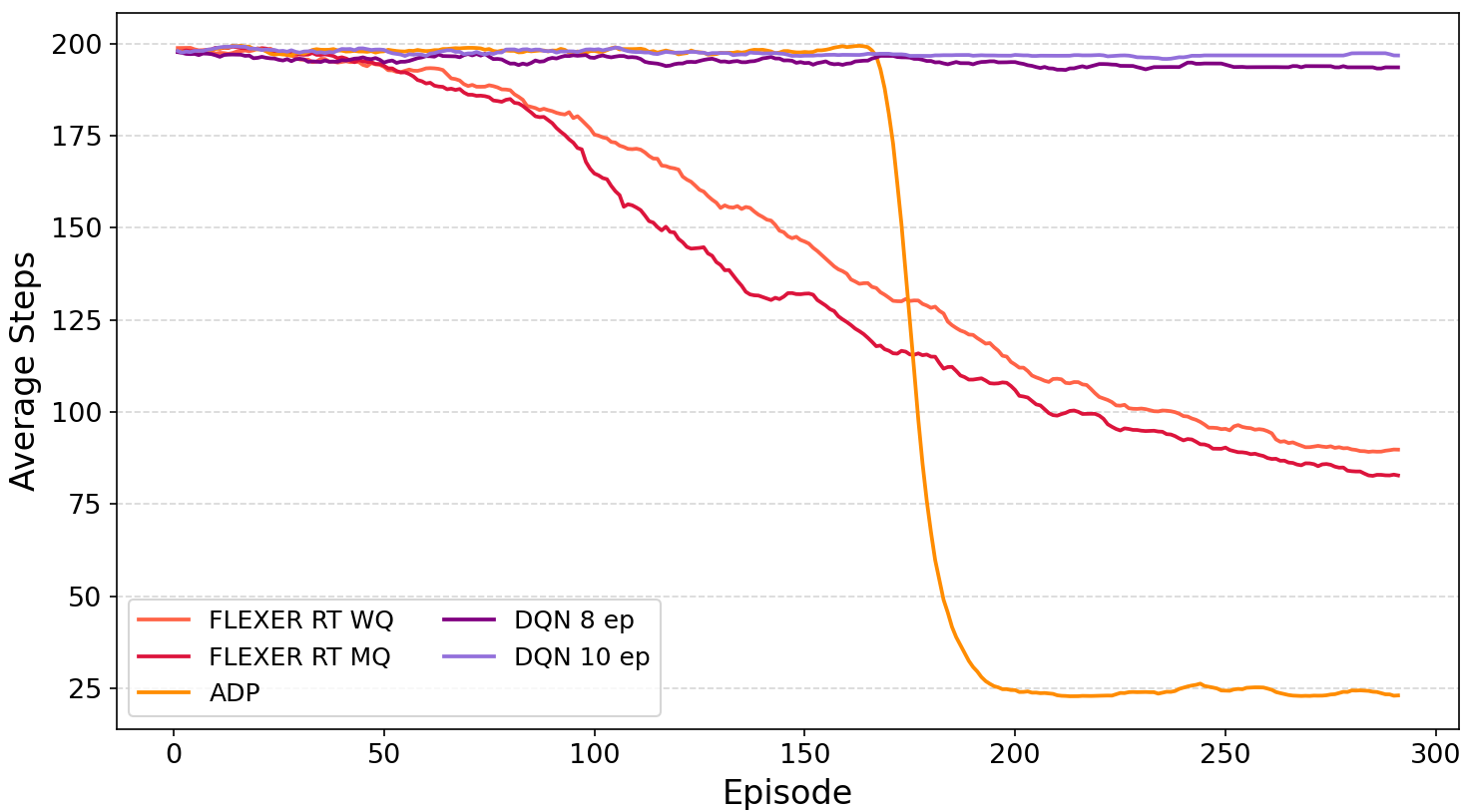}
        \caption{Steps in episode}
        \label{fig:blocksworld-comp-steps}
    \end{subfigure}
    \hfill
    \begin{subfigure}{0.48\textwidth}
        \includegraphics[width=\textwidth]{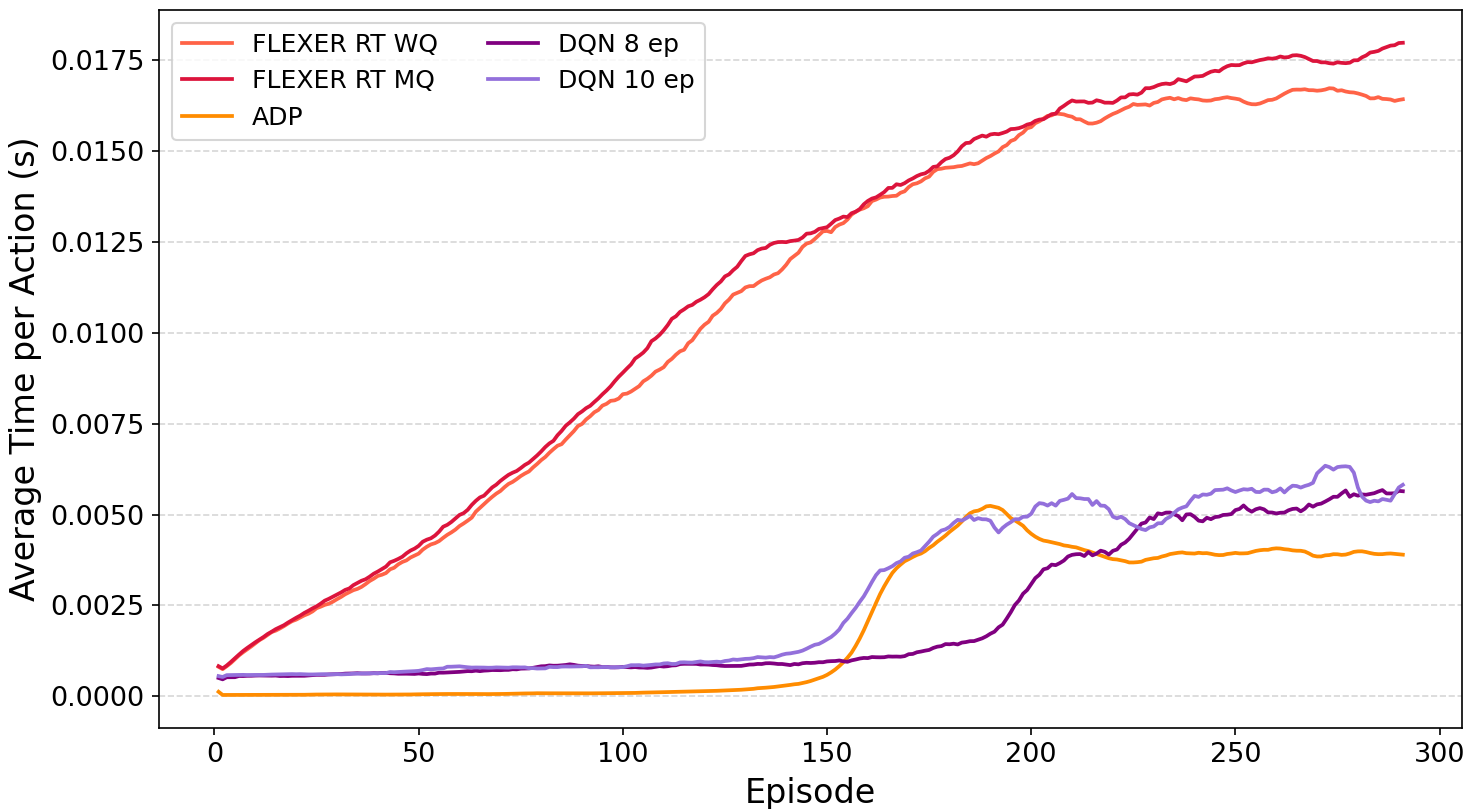}
        \caption{Time per step}
        \label{fig:blocksworld-comp-times}
    \end{subfigure}
    \caption{Results of experiments running best Flexers, DQN and ADP on the BlocksWorld[4,4] environment.}
    \label{fig:blocksworld-comp}
\end{figure*}

\begin{figure*}[htb!]
    \centering
    \begin{subfigure}{0.4\textwidth}  
        \centering
\includegraphics[scale=0.45]{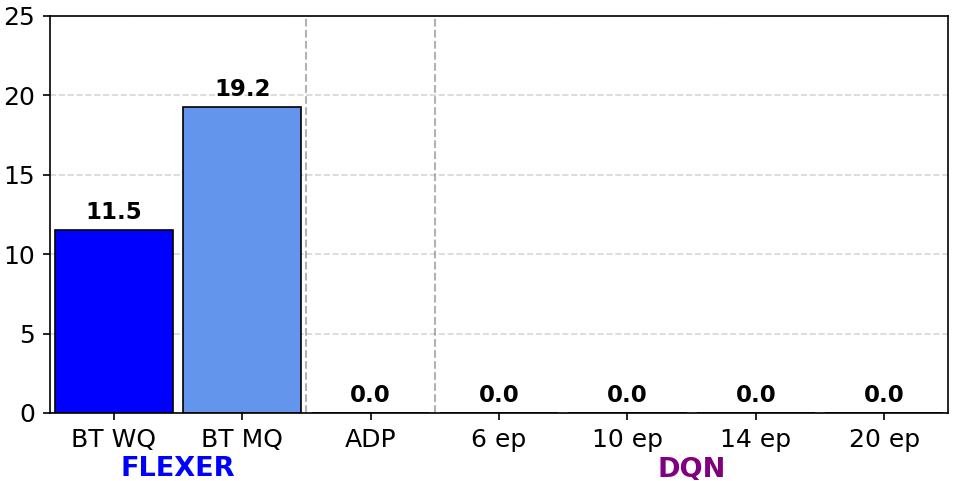}
        \caption{Goals completed}
        \label{fig:movingnumbers-comp-goals}
    \end{subfigure}
    \hfill
    \begin{subfigure}{0.48\textwidth}
        \includegraphics[width=\textwidth]{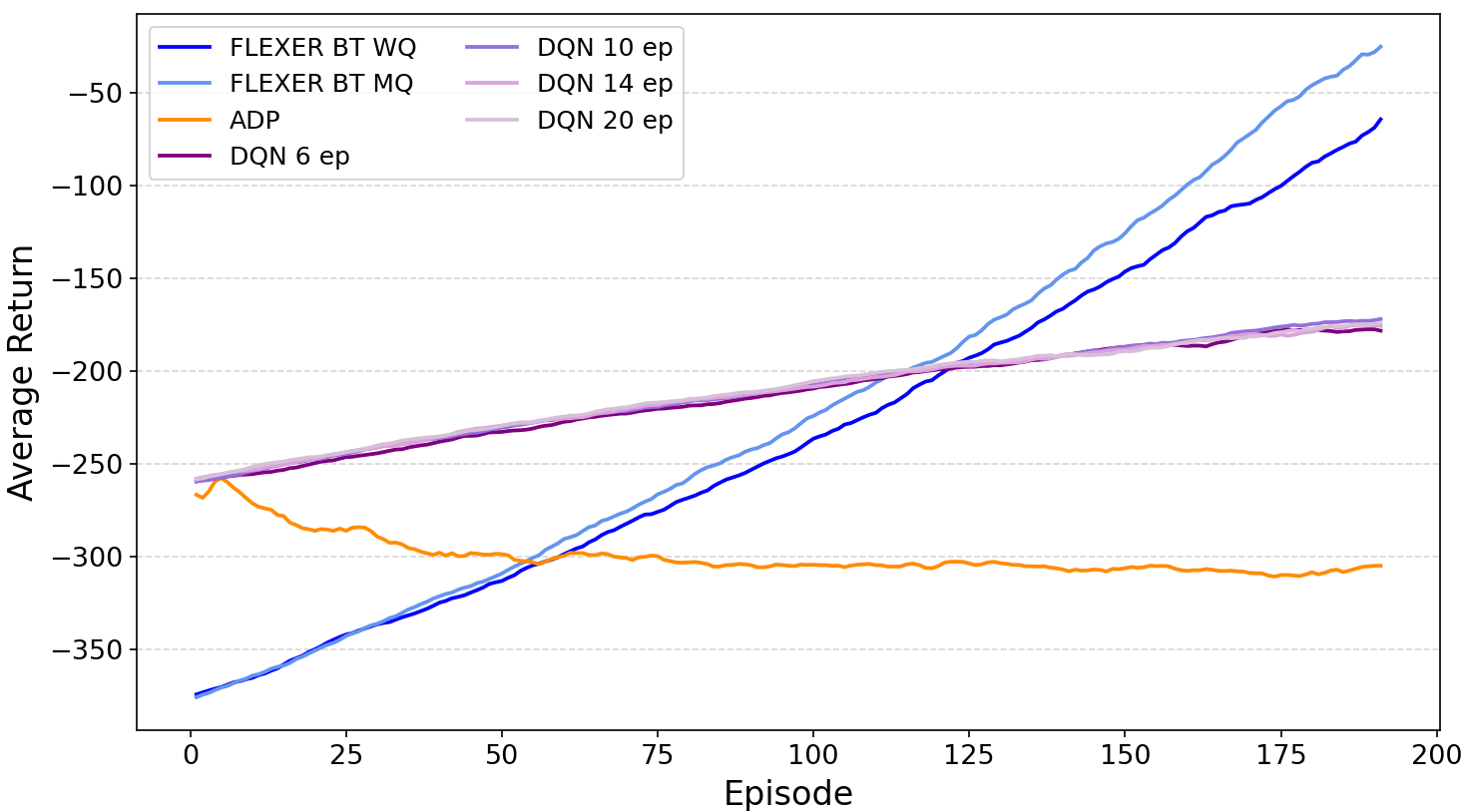}
        \caption{Return}
        \label{fig:movingnumbers-comp-returns}
    \end{subfigure}
    
    \begin{subfigure}{0.48\textwidth}
    \vspace{4mm}
        \includegraphics[width=\textwidth]{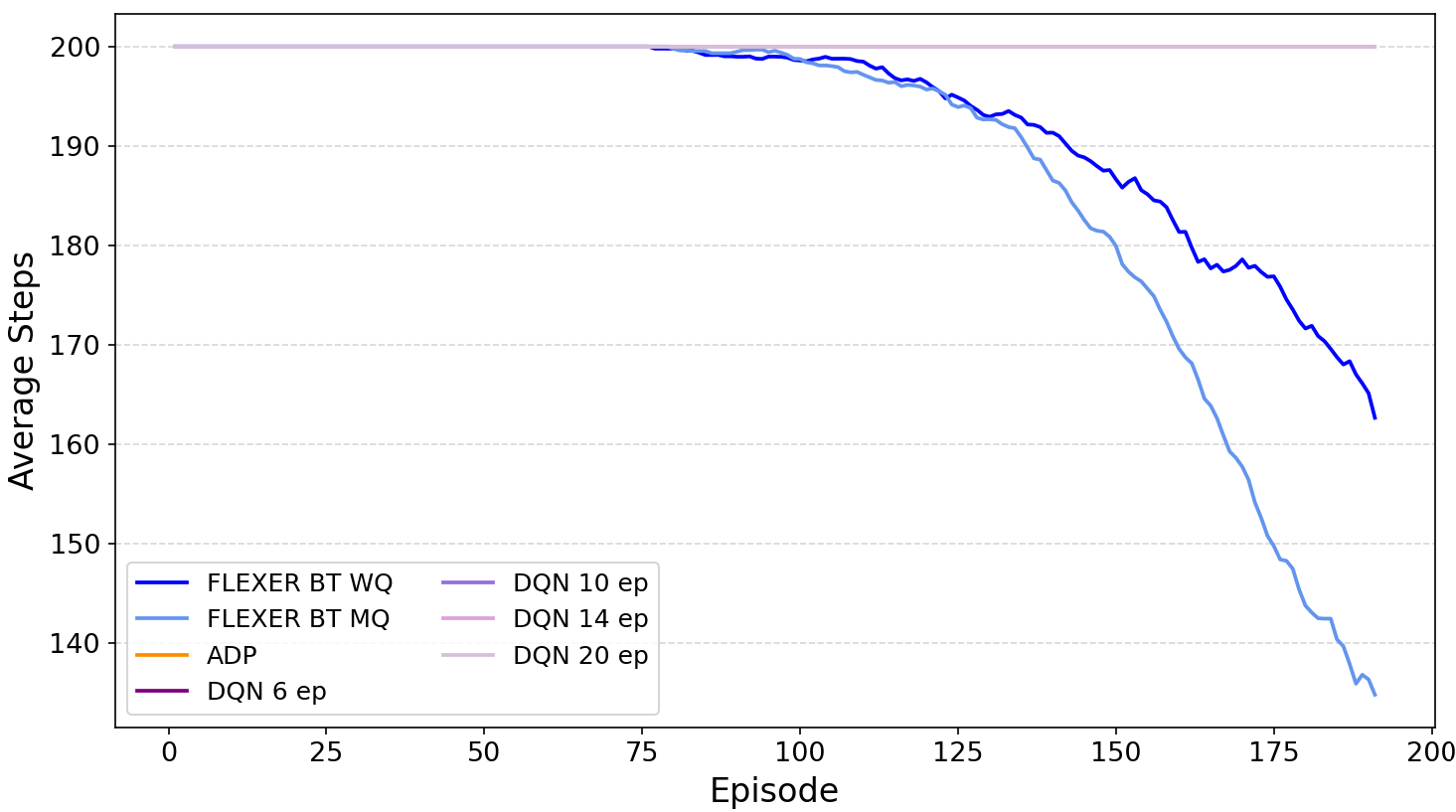}
        \caption{Steps in episode}
        \label{fig:movingnumbers-comp-steps}
    \end{subfigure}
    \hfill
    \begin{subfigure}{0.48\textwidth}
        \includegraphics[width=\textwidth]{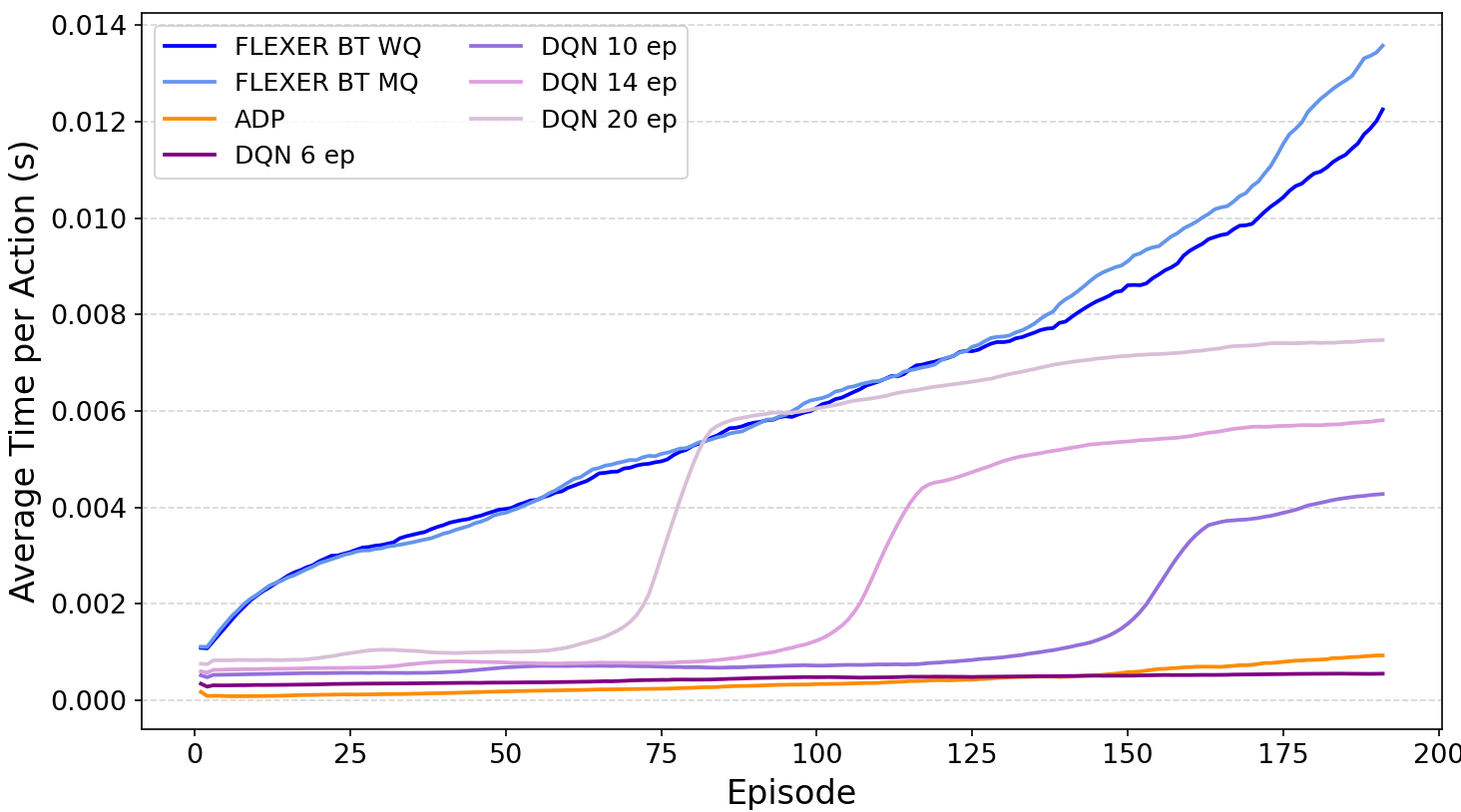}
        \caption{Time per step}
        \label{fig:movingnumbers-comp-times}
    \end{subfigure}
    \caption{Results of experiments running best Flexers, DQN and ADP on the MovingNumbers[10,3] environment.}
    \label{fig:movingnumbers-comp}
\end{figure*}

We compare the best performing (in terms of goals achieved) variant of Flexer (per environment) with the DQN and ADP algorithms. 
The number of training epochs for DQN are chosen such that the maximum time per action for DQN is comparable with the maximum time per action of any algorithm it is compared to. This turns out to be 8 and 10.
Figures \ref{fig:simplegrid-comp},  \ref{fig:blocksworld-comp} and \ref{fig:movingnumbers-comp} show the results for the hard SimpleGrid, BlocksWorld and MovingNumbers problems, respectively.

\section{Discussion}\label{sec:Discussion}

\subsection{Flexer and AZLike on Easy Environs}

\subsubsection{SimpleGrid[10,0.15]}

AZLike achieves more goals, 92 compared to 85 out of 100. All tabular variants have non-decreasing return non-increasing steps. Flexer runs twice as fast as AZLike-RN/RT but AZLike-BT is twice as fast as Flexer.

\subsubsection{BlocksWorld[3,3]}

Flexer does slightly better here: Flexer-RT gets 126 vs.\ AZLike's 122 out of 150. Flexer-BT/RT/RN have non-decreasing returns and non-increasing steps. All tabular variants do better with respect to number of goals achieved.
Runtimes per action are spread out between 0.005 and 0.03, with AZLike-RN/RT at the high end and AZLike-BT/BN at the low end. All Flexers are clustered around 0.015 to 0.02.

\subsubsection{MovingNumbers[5,2]}

Here, Flexer-BN, Flexer-BN and AZLike-BT achieve comparable number of goals ($\sim 60$ out of 100). All Flexers have non-decreasing returns and all AZLikes have decreasing returns. Flexer shows consistent decrease in number of steps whereas AZLike steps consistently start decreasing at $\sim 65$ episodes.
Running times are quite spread out from $\sim0.08$ to $\sim0.034$ per action. AZLike is fastest, with AZLike-RT and Flexer-RN second fastest at half the speed of AZLike-BT. AZLike-BN/RN take the longest.

\subsection{Flexer and AZLike on Harder Environs}

\subsubsection{SimpleGrid[15,0.3]}

AZLike-RT gests more goals than Flexer-BT (106 vs.\ 96 out of 200). AZLike-BT and Flexer-RT achieve in the mid 70s.
It is interesting that, in this case, neither variants with rollout nor with bootstrapping dominate.
All returns are non-decreasing, but for AZLike-BT, steps start decreasing and AZLike-RT also decreases a small amount at the end.
 AZLike-RT takes almost 0.025 sec/act whereas Flexer takes 0.017 sec/act.
 So, if returns and running times are considered as equally important , then AZLike-RT and Flexer-BT perform equally.

 \subsubsection{BlocksWorld[4,4]}

Both AZLike-BT and AZLike-RT achieve more goals than Flexer-BT and Flexer-RT, with Flexer-BT performing particularly poorly compared to the other three variants. ($\sim40$ vs.\ $\sim80$ out of 300). However, both AZLikes have extremely mis-shaped curves, concave for returns and convex for steps.
Flexer-RT has comparable running time to AZLike-BT and approximately twice as fast as as AZLike-RT.
There is no clear winner for this environ (although Flexer-BT clearly loses).

\subsubsection{MovingNumbers[10,3]}

On this environ, Flexer-BT significantly outperforms all three other variants. MovingNumbers[10,3] is arguably the hardest of all problems tested on in this study.

\subsection{Flexer with Boosted Value Function}

Recall that boosting with the maximum q-value is abbreviated as MQ and boosing with weighted q-values is abbreviated as WQ.

\subsubsection{SimpleGrid}
MQ clearly boosts Flexer-BT. Both MQ and WQ boost Flexer-RT. There is no significant running-time penalty for boosing in this case.

\subsubsection{BlocksWorld}
Both MQ and WQ boost Flexer-RT from 0to 50 (with -122 being the minimum) at the end of training. There is a slight running-time penalty for boosting: at $\sim160$ episodes, boosted Flexer takes $\sim27\%$ longer, but by the end of training (300 episodes), running-times are close to equal.

\subsubsection{MovingNumbers}
Boosting has no significant effect in this case.

\subsection{Comparison to Benchmarks}

\subsubsection{SimpleGrid}
With respect to goals, ADP significantly outperforms the next best algorithm (150 vs.\ 123 of Flexer-BT-MQ) out of 200.
DQN does poorly and takes long towards the end of triaing. ADP is extremely fast compared to Flexer.

\subsubsection{BlocksWorld}
The goal-achieving performance of Flexer-RT (MQ and WQ) and ADP are comparable, with Flexer-RT-MQ at 132 and ADP slightly lower at 124.
DQN performs very poorly. In terms of number of steps, ADP is significantly better than Flexer, achieving what seems to be the optimal number of steps to stack four blocks using only $\sim25$ actions. ADP takes $\sim0.004$ sec/action, while Flexer takes $\sim0.0175$ at the end of training

 \subsubsection{MovingNumbers}
 ADP and DQN cannot solve the problem, whereas Flexer can fins three numbers and place them in their locations (in the correct order) 12 times (BT-WQ) and 19 times (BT-MQ) out of 200 episodes. Both versins of boosting consume the same amount of time, but boosting with MQ achieves $\sim7$ more goals (on average) and ends up finding the goal in less than 140 steps, compared to over 160 steps in the case of Flexer-BT-WQ.

\section{Conclusions}\label{sec:conclusions}

We introduced a reinforcement learning architecture called Flexer. It maintains a measure based on the quality of the policy network (PN) and environment models (EM). This measure is mainly used as a mixing factor ($\mu$) to mix the PN output and the planner root result at every step, favoring planning in proportion to low PN quality and high EM quality. The $\mu$ and PN quality measures are also used to adaptively adjust some parameters to improve performance and save compute budget.
When both the PN and EM are poor, Flexer falls back to random actions rather than producing misleading plans, avoiding compounding errors early in training.

Another novel feature of Flexer is that it maintains a (tabular) value function from which training targets are derived. This is different from architectures like AlphaZero and MuZero, which derive training targets from the planner's root node. Flexer also has a unique 'value-boosting' mechanism that uses MCTS tree-node values to improve the value function.

One could regard Flexer's greater complexity, compared to AlphaZero/AZLike, as a drawback: It introduces several interacting components — a global value function, two quality signals ($\psi_{PN}$, $\kappa_{EM}$), variance trackers for $T$ and $R$, adaptive budget/temperature/epoch mechanisms, and optional value boosting. Each adds hyperparameters and potential failure modes, which makes analysis more challenging.

On easy environments, Flexer and AZLike perform comparably with no consistent winner. On harder problems, AZLike has an edge on BlocksWorld[4,4], but Flexer-BT decisively outperforms all variants on MovingNumbers[10,3], the hardest problem tested. 
Flexer-BT succeeds on hard MovingNumbers because the bootstrapped value function provides meaningful gradient signal even without goal completion, the adaptive budget prevents the agent from being misled by poor early models, and the cumulative value updates create a cross-episode memory that neither AZLike, DQN, nor ADP possess.

Throughout, Flexer produces more stable learning curves (non-decreasing returns, non-increasing steps), while AZLike's curves degrade under difficulty. Bootstrapping (BT) is the most impactful design choice in both families. Value-function boosting (MQ/WQ) helps on SimpleGrid and BlocksWorld, with MQ outperforming WQ on hard MovingNumbers at no extra runtime cost. Among benchmarks, ADP is competitive on simpler environments and achieves near-optimal efficiency on BlocksWorld, while DQN is consistently the weakest. Critically, both ADP and DQN fail entirely on hard MovingNumbers, where only Flexer-BT solves the problem — pointing to a clear scalability advantage for Flexer on complex, sequential planning tasks.

In future, we could investigate other measures of model quality, based on, for instance, Shannon entropy and/or more sophisticated measures of variance or risk minimization.

Earlier in our research, the value function was represented as a neural network. Performance of those versions were generally poor. However, for Flexer to scale to larger, real-world environments, neural nets or some kind of function approximation will have to be employed. Neural net inference is relatively expensive. Future work will involve performing an in-depth literature survey on how others have tackled this inference-cost issue and investigate how to minimize the need for inference, using it sparingly, when most useful in Flexer.

In variants that learn and use tabular environment models, computing $var(T, s)$ and $var(R, s)$ requires growing a breadth-first tree of reachable states and filtering large transition buffers at every step — an overhead that grows with state space size. On the other hand, NN-based environment models have other challenges, as discussed above. These issues will have to be tackled.

Lastly, Flexer could be enhanced to leverage the latest techniques employing latent spaces to achieve true scalability, as is done in DVRL \cite{izlww18}, MuZero \cite{muzero20},  Dreamer \cite{hlbn20} and BetaZero \cite{mcck24}.

\bibliographystyle{apalike}
{\small
\bibliography{references}}

\end{document}